\documentclass[10pt,journal,compsoc]{IEEEtran}
\usepackage{times}
\usepackage{epsfig}
\usepackage{graphicx}
\usepackage{amsmath}
\usepackage{amssymb}
\usepackage{mathrsfs}
\usepackage{wasysym}
\usepackage{color}
\usepackage{booktabs}
\usepackage{multirow}

\usepackage[lined,ruled,linesnumbered]{algorithm2e}

\usepackage[pagebackref=false,breaklinks=true,citecolor=blue,linkcolor=blue,letterpaper=true,colorlinks,bookmarks=false]{hyperref}

\usepackage[normalem]{ulem}
\useunder{\uline}{\ul}{}

\DeclareMathOperator*{\argmax}{argmax}
\DeclareMathOperator*{\argmin}{argmin}

\newcommand\tb[1]{\textbf{#1}}

\ifCLASSOPTIONcompsoc
\usepackage[nocompress]{cite}
\else
\usepackage{cite}
\fi

\newcommand{\first}[1]{\textcolor{red}{#1}}
\newcommand{\second}[1]{\textcolor{black}{#1}}
\newcommand{\third}[1]{\textcolor{green}{#1}}

\begin{document}
%\title{Learning Hierarchical Convolutional Features \\ for Robust Visual Tracking}
\title{Robust Visual Tracking\\via Hierarchical Convolutional Features}
\author{Chao~Ma, 
Jia-Bin~Huang, 
Xiaokang~Yang,
and~Ming-Hsuan~Yang% <-this % stops a space
\\ 
\IEEEcompsocitemizethanks{\IEEEcompsocthanksitem C. Ma and X. Yang are with the MoE Key Lab of Artificial Intelligence, AI Institute, Shanghai Jiao Tong University,  Shanghai, 200240, P. R. China. E-mail: {\{chaoma, xkyang\}@sjtu.edu.cn}. C. Ma is also with the Australian Centre for Robotic Vision, The University of Adelaide, Adelaide, 5005, Australia.
%E-mail: {c.ma@adelaide.edu.au}.
\IEEEcompsocthanksitem J.-B. Huang is with the Department of Electrical and Computer Engineering, Virginia Tech, VA. E-mail: {jbhuang@vt.edu}
%\IEEEcompsocthanksitem X. Yang 
%is with Institute of Image Communication and Network Engineering, Shanghai Jiao Tong University, Shanghai, 200240, P. R. China.
%E-mail: {xkyang@sjtu.edu.cn}. 
\IEEEcompsocthanksitem M.-H. Yang is with School of Engineering, 
University of California, Merced, CA, 95344.
E-mail: {mhyang@ucmerced.edu}.
}% <-this % stops a space
%\thanks{Manuscript received April 19, 2005; revised September 17, 2014.}
}

\IEEEtitleabstractindextext{
\begin{abstract}
Visual tracking is challenging as target objects often undergo significant appearance changes caused by deformation, abrupt motion, background clutter and occlusion.
In this paper, we propose to exploit the rich hierarchical features of deep convolutional neural networks to improve the accuracy and robustness of visual tracking.
Deep neural networks trained on object recognition datasets consist of multiple convolutional layers.
These layers encode target appearance with different levels of abstraction.
For example, the outputs of the last convolutional layers encode the semantic information of targets and such representations are invariant to significant appearance variations.
However, their spatial resolutions are too coarse to precisely localize the target.
In contrast, features from earlier convolutional layers provide more precise localization but are less invariant to appearance changes.
We interpret the hierarchical features of convolutional layers as a nonlinear counterpart of an image pyramid representation and explicitly exploit these multiple levels of abstraction to represent target objects.
Specifically, we learn adaptive correlation filters on the outputs from each convolutional layer to encode the target appearance. 
We infer the maximum response of each layer to locate targets in a coarse-to-fine manner. 
\textcolor{black}{To further handle the issues with scale estimation and re-detecting target objects from tracking failures caused by heavy occlusion or out-of-the-view movement, we conservatively learn another correlation filter, that maintains a long-term memory of target appearance, as a discriminative classifier.}
We apply the classifier to two types of object proposals: 
(1) proposals with a small step size and tightly around the estimated location for scale estimation; and 
(2) proposals with large step size and across the whole image for target re-detection. 
Extensive experimental results on large-scale benchmark datasets show that the proposed algorithm performs favorably against the state-of-the-art tracking methods.
\end{abstract}

% Note that keywords are not normally used for peer review papers.
}

% make the title area
\maketitle

\IEEEpeerreviewmaketitle

\IEEEraisesectionheading{\section{Introduction}
\label{sec:introduction}}

\IEEEPARstart{V}{isual} object tracking is one of the fundamental problems in computer vision with numerous applications such as intelligent video surveillance, human computer interaction and autonomous driving~\cite{DBLP:journals/csur/YilmazJS06,DBLP:journals/tist/LiHSZDH13,DBLP:journals/pami/SmeuldersCCCDS14,DBLP:journals/pami/KristanMLVPFNPC16}, to name a few.
Given the initial state (e.g., position and scale) of a target object in the first frame, the goal of visual tracking is to estimate the unknown states of the target in the subsequent frames. 
Despite the significant progress made in the recent years~\cite{DBLP:journals/pami/ComaniciuRM03, DBLP:journals/pami/Avidan04,DBLP:journals/ijcv/RossLLY08,DBLP:conf/iccv/MeiL09,DBLP:journals/pami/BabenkoYB11,DBLP:journals/pami/KalalMM12,DBLP:journals/pami/HenriquesC0B15}, visual object tracking remains challenging due to large appearance variations caused by numerous negative factors including illumination variation, occlusion, background clutters, abrupt motion, and target moving out of views. 
To account for the appearance changes over time, considerable efforts have been made to design invariant feature descriptors to represent target objects, 
such as color histograms~\cite{DBLP:journals/pami/ZhaoYT10}, HOG~\cite{DBLP:conf/cvpr/DalalT05}, Haar-like features~\cite{DBLP:journals/pami/BabenkoYB11}, SURF~\cite{DBLP:conf/cvpr/TaCGP09}, ORB~\cite{DBLP:conf/iccv/RubleeRKB11}, subspace representations~\cite{DBLP:journals/ijcv/RossLLY08}, and superpixels~\cite{DBLP:conf/iccv/WangLYY11}.
Recently, features learned from convolutional neural networks (CNNs) have been used in a wide range of vision tasks, e.g., image classification~\cite{DBLP:conf/nips/KrizhevskySH12}, object recognition~\cite{DBLP:conf/cvpr/GirshickDDM14}, and image segmentation~\cite{DBLP:conf/cvpr/LongSD15}, with the state-of-the-art performance. 
%
%CM: I revised this sentence below
%
%\textcolor{black}{Recently, convolutional neural networks (CNNs) ~\cite{DBLP:conf/nips/KrizhevskySH12,DBLP:journals/corr/SimonyanZ14a,DBLP:conf/cvpr/HeZRS16} pre-trained for the image classification task~\cite{DBLP:conf/cvpr/DengDSLL009} have been widely used as backbone feature extractor to advance numerous vision tasks, e.g., object recognition~\cite{DBLP:conf/cvpr/GirshickDDM14}, and image segmentation~\cite{DBLP:conf/cvpr/LongSD15}, with the state-of-the-art performance.} 
%
It is thus of great interest to understand how to best exploit the hierarchical features in CNNs for visual tracking.

Existing tracking methods based on deep learning~\cite{DBLP:conf/nips/WangY13,DBLP:conf/bmvc/LiLP14,DBLP:journals/tip/WangLWCY15,DBLP:conf/icml/HongYKH15,DBLP:journals/corr/NamH15} 
typically formulate the tracking task as a detection problem in each frame. 
These trackers first draw positive and negative training samples around the estimated target location to incrementally learn a classifier over the features extracted from a CNN.
The learned classifier is then used to detect the target in the subsequent frames.
%
%Two issues ensue with such approaches. 
%There are two ensuring problems that need to be addressed. 
%%
%The first issue lies in the use of a CNN as an online classifier following recent image classification approaches, where only the outputs of the last layer are used to represent targets.
%%
%For high-level visual recognition problems, it is effective to use features from the last layer as they are
%closely related to category-level semantics and invariant to nuisance variables such as intra-class variations and precise localizations.
%%
%However, the objective of visual tracking is to locate targets precisely rather than to infer their semantic categories.
%%
%Using only the features from the last layer is thus not ideal for representing target objects in visual tracking.
%%
%The second issue is concerned with extracting training samples.
%%
%Training a robust classifier requires a considerably large number of positive and negative samples, 
%which are not available in the context of visual tracking.
%%
%In addition, there lies ambiguity in determining a decision boundary since positive and negative samples are highly correlated due to sampling near a target.
%
\textcolor{black}{
There are two issues with such approaches: (i) how to best exploit CNN features and (ii) how to effectively extract training samples. 
The first issue lies in the use of a CNN as an online classifier similar to recent image classification approaches, where the outputs of the last layer are mainly used to represent targets.}
For high-level visual recognition problems, it is effective to use features from the last layer as they are
closely related to category-level semantics and invariant to nuisance variables such as intra-class variations and precise localizations.
However, the objective of visual tracking is to locate targets precisely rather than to infer their semantic categories.
Using only the features from the last layer is thus not ideal for representing target objects in visual tracking.
%
%The second issue is concerned with extracting training samples.
%
\textcolor{black}{The second issue is concerned with that training a robust classifier requires a considerably large number of positive and negative samples, 
which are not available in the context of visual tracking. }
In addition, there lies the ambiguity in determining a decision boundary since positive and negative samples are highly correlated due to sampling near a target.

\begin{figure*}[t]
\centering
\includegraphics[width=.8\textwidth]{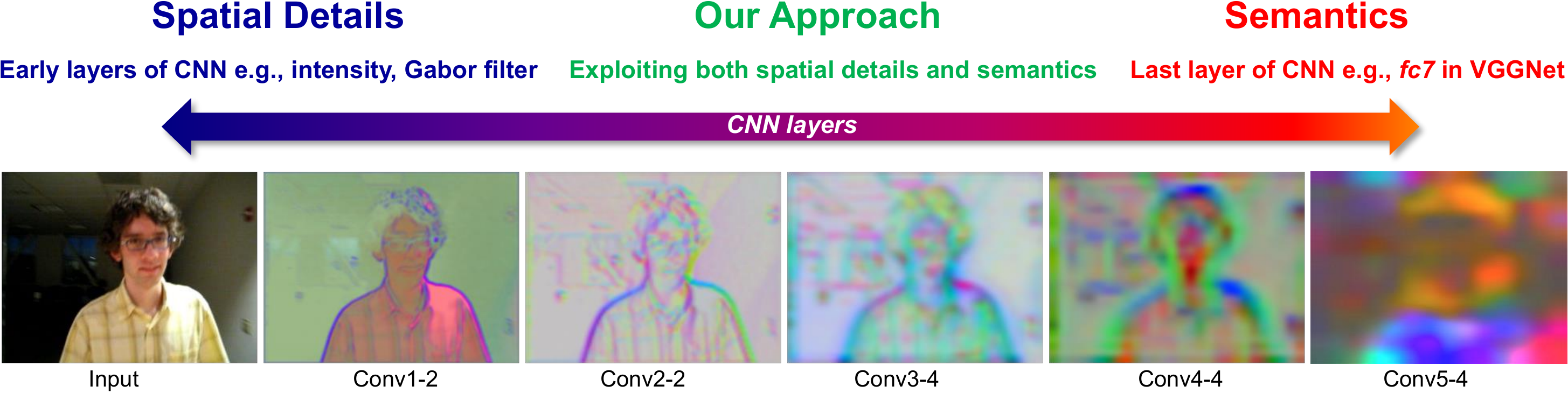} 
\vspace{-2mm}
\caption{
\tb{Visualization of hierarchical deep features.}
Convolutional layers of a typical CNN model, e.g., AlexNet~\cite{DBLP:conf/nips/KrizhevskySH12} or VGGNet~\cite{DBLP:journals/corr/SimonyanZ14a}, provide multiple levels of abstraction in the feature hierarchies.
We visualize a sample frame 
using the VGGNet-19~\cite{DBLP:journals/corr/SimonyanZ14a} to extract CNN features from the first to fifth convolutional layers.
Deep features in the earlier layers retain higher spatial resolution for precise localization with low-level visual information similar to the response map of Gabor filters~\cite{DBLP:journals/pami/BovikCG90}.
On the other hand, features in the latter layers capture more semantic information and less fine-grained spatial details.
Our approach aims to exploit the semantic information of last layers (right) to handle large appearance changes and alleviate drifting by using features of earlier layers (left) for precise localization.
}
\label{fig:cnnlayer} 
\vspace{-3mm}
\end{figure*}

In this work, we address these two issues by 
(i) using the features from the hierarchical layers of CNNs rather than only the last layer to represent target objects and 
(ii) learning adaptive correlation filters on each CNN layer without the need for target state sampling.
Our approach is built on the observation that although the last layers of \textcolor{black}{pre-trained} CNNs are more effective to capture semantics, they are insufficient to encode the fine-grained spatial details such as the target positions.
The earlier layers, on the other hand, are precise in localization but do not capture semantics as illustrated in Figure~\ref{fig:cnnlayer}.
\textcolor{black}{Further discussions on the roles of different CNN layers for object recognition can be found in \cite{zeiler2011adaptive,zeiler2014visualizing}.}
%\textcolor{black}{We refer readers to \cite{zeiler2011adaptive,zeiler2014visualizing} for further discussions of the roles of different CNN layers for object recognition.}
%
This observation suggests that reasoning with multiple layers of CNN features for visual tracking is of great importance as semantics are robust to significant appearance variations, and spatial details are effective 
for precise localization.
We exploit both hierarchical features from the recent advances in CNNs and 
an inference approach across multiple levels in classical computer vision problems.
For example, computing optical flow from the coarse levels of the image pyramid is efficient, but finer levels ensure obtaining an accurate and detailed flow field.
A coarse-to-fine search strategy is often adopted for coping with large motion~\cite{DBLP:conf/ijcai/LucasK81}.
As such, we learn one adaptive correlation filter~\cite{DBLP:conf/cvpr/BolmeBDL10, DBLP:conf/eccv/HenriquesCMB12,
DBLP:conf/cvpr/DanelljanKFW14, DBLP:conf/eccv/ZhangZLZY14,
DBLP:conf/bmvc/DanelljanKFW14, DBLP:journals/pami/HenriquesC0B15}
over features extracted from each CNN layer and use these multi-level correlation response maps to infer the target location collaboratively.
We consider \textit{all} the shifted versions of features as training samples and regress them to soft labels generated by a Gaussian function with a small spatial bandwidth and a range of zero to one, 
thereby alleviating the sampling ambiguity of training a binary discriminative classifier.

%The preliminary results of this work have been published in~\cite{DBLP:conf/iccv/MaHYY15}.
%%
%Compared to the previous work~\cite{DBLP:conf/iccv/MaHYY15}, we further address two additional challenges: (i) scale estimation and (ii) target re-detection from tracking failures caused by heavy occlusion or targets moving out of the view. 
%%
%We propose to use another correlation filter to maintain a long-term memory of target appearance as a classifier.
%%
%By adjusting the parameters, we tailor the off-the-shelf EdgeBox~\cite{DBLP:conf/eccv/ZitnickD14} to generate two types of region proposals: 
%(i) proposals tightly around the estimated location as candidates for scale estimation; 
%(ii) proposals sampled across the whole image as candidates for target re-detection.
%%
%We apply the classifier to these two types of proposals and respectively select the proposals with highest response scores for scale estimation and target re-detection.
%%

%The main contribution of this work is an effective approach that best exploits rich hierarchies of CNN features for robust visual tracking.
%
%Specifically, we make the following four contributions.
%We propose an an effective approach that best exploits rich hierarchies of CNN features for robust visual tracking with the following contributions:
The main contributions of this work are summarized as below:
\begin{itemize}
%\item We exploit the hierarchical features of CNNs as target representations for visual tracking, where both semantics and fine-grained details are exploited to handle large appearance variations and avoid drifting.
%\item We learn adaptive linear correlation filters on each CNN layer to alleviate the sampling ambiguity.
%
%We infer the target location using the multi-level correlation response maps in a coarse-to-fine fashion.
%
\item \textcolor{black}{We exploit the hierarchical convolutional layers of pre-trained CNNs as features to capture multi-level semantic abstraction of target objects. At each layer, we adaptively learn linear correlation filter to alleviate the sampling ambiguity. We infer target positions by reasoning multi-level correlation response maps in a coarse-to-fine manner.}
\item We propose to learn another correlation filter with a conservative learning rate to maintain a long-term memory of target appearance as a discriminative classifier. We apply the classifier to two types of region proposals 
for scale estimation and target re-detection from tracking failures. 
\item We carry out experiments on the large-scale benchmark datasets: OTB2013~\cite{DBLP:conf/cvpr/WuLY13}, OTB2015~\cite{DBLP:journals/pami/WuLY15}, VOT2014~\cite{Hadfield14d}, and VOT2015~\cite{DBLP:conf/iccvw/KristanMLFCFVHN15}. Extensive experimental results demonstrate that the proposed tracking algorithm performs favorably against the state-of-the-art methods in terms of accuracy and robustness.
\end{itemize}

\vspace{-2mm}
\section{Related Work}

%Visual object tracking has been an active research topic in computer vision.
%
We discuss tracking methods closely related to this work in this section.
A comprehensive review on visual tracking can be found in~\cite{DBLP:journals/csur/YilmazJS06,DBLP:journals/tist/LiHSZDH13,DBLP:journals/pami/SmeuldersCCCDS14,DBLP:journals/pami/KristanMLVPFNPC16}.

%\subsection{Tracking-by-Detection}
{\flushleft {\bf Tracking by Detection.}}
Visual tracking can be posed as a sequential detection problem in a local window, 
where classifiers are often updated online with positive and negative examples. 
%
%For each frame, a set of positive and negative training samples are collected for incrementally updating a discriminative classifier to separate a target from its surrounding background. 
%
%However, the sampling ambiguity problem arises with such approach that draws samples for learning online classifiers.
%
Since the classifiers are updated to adapt to appearance variations over time, slight inaccuracies in the labeled samples negatively affect the classifier and gradually cause the trackers to drift.
Considerable efforts have been made to alleviate these model update problems caused by the sampling ambiguity.
The core idea of these algorithms lies in how to properly update a discriminative classifier to reduce tracking drifts.
Examples include ensemble learning~\cite{DBLP:journals/pami/Avidan07,DBLP:conf/iccv/BaiWSBM13}, semi-supervised learning~\cite{DBLP:conf/eccv/GrabnerLB08}, multiple instance learning (MIL)~\cite{DBLP:journals/pami/BabenkoYB11}, and transfer learning~\cite{DBLP:conf/eccv/GaoLHX14}. 
Instead of learning only one single classifier, Kalal et al.~\cite{DBLP:journals/pami/KalalMM12} decompose the tracking task into tracking, learning, and detection (TLD) modules where the tracking and detection modules facilitate each other, i.e., the results from the tracker provide additional training samples to update the detector. 
The online learned detector can be used to reinitialize the tracker when tracking failure occurs. 
Similar mechanisms have also been exploited~\cite{DBLP:conf/eccv/Pernici12,DBLP:conf/cvpr/SupancicR13, DBLP:conf/eccv/HuaAS14} to recover target objects from tracking failures. 
Hare et al.~\cite{DBLP:conf/iccv/HareST11} show that the objective of label prediction using a classifier is not explicitly coupled to the objective of tracking (accurate position estimation) and pose the problem 
as a joint structured output prediction task. 
Zhang et~al.~\cite{DBLP:conf/eccv/ZhangMS14} combine multiple classifiers with different learning rates for visual tracking. 
By alleviating the sampling ambiguity problem, these methods perform well in the benchmark study~\cite{DBLP:conf/cvpr/WuLY13}. 
We address the sample ambiguity with correlation filters where training samples are regressed to soft labels generated by a Gaussian function rather than binary labels for learning discriminative classifier.

%\vspace{-2mm}
{\flushleft \bf Tracking by Correlation Filters.}
%\subsection{Tracking by Correlation Filters}
Correlation filters for visual tracking have attracted considerable attention due to its high computational efficiency with the use of fast Fourier transforms. 
Tracking methods based on correlation filters regress all the circular-shifted versions of input features to soft labels generated by a Gaussian function and ranging from zero to one. 
As a result, learning correlation filters does not require hard-thresholded samples 
corresponding to a target object. 
Bolme et al.~\cite{DBLP:conf/cvpr/BolmeBDL10} learn a minimum output sum of squared error filter over the luminance channel for fast visual tracking.
Several extensions have been proposed to considerably improve tracking accuracy including kernelized correlation filters~\cite{DBLP:conf/eccv/HenriquesCMB12}, multi-dimensional features~\cite{DBLP:journals/pami/HenriquesC0B15,DBLP:conf/cvpr/DanelljanKFW14}, context learning~\cite{DBLP:conf/eccv/ZhangZLZY14}, spatial weights~\cite{DBLP:conf/iccv/DanelljanHKF15}, scale estimation~\cite{DBLP:conf/bmvc/DanelljanKFW14}, and efficient filter mining~\cite{conf/cvpr/DanelljanBKF17}.
In this work, we propose to learn correlation filters over multi-dimensional features in a way similar to the recent tracking methods~\cite{DBLP:conf/cvpr/DanelljanKFW14,DBLP:journals/pami/HenriquesC0B15}. 
The main differences lie in the use of learned deep
features rather than hand-crafted features (e.g., HOG~\cite{DBLP:conf/cvpr/DalalT05} or color-attributes~\cite{DBLP:conf/cvpr/DanelljanKFW14}), 
and we construct multiple correlation filters on hierarchical convolutional layers 
as opposed to only one single filter by existing approaches.
We also note that the methods mentioned above often adopt a moving average scheme to update correlation filters in each frame to account for appearance variations over time.
These trackers are prone to drift due to noisy updates~\cite{DBLP:journals/pami/MatthewsIB04}, and cannot recover from tracking failures due to the lack of re-detection modules.
Recently, numerous methods address this issue using an online trained detector~\cite{DBLP:conf/cvpr/MaYZY15} or 
a bag of SIFT descriptors~\cite{DBLP:journals/ijcv/Lowe04}. 
In this work, we learn another correlation filter with a conservative learning rate 
to maintain the long-term memory of target appearance as a discriminative classifier.
We apply the classifier over selected region proposals by the EdgeBox~\cite{DBLP:conf/eccv/ZitnickD14} 
method for recovering targets 
from tracking failures as well as for scale change estimation. 

{\flushleft \bf Tracking by Deep Neural Networks.}
%\subsection{Tracking by Deep Neural Networks}
%
The recent years have witnessed significant advances in deep neural networks on a wide range of computer vision problems.
However, considerably less attention has been made to apply deep networks to visual tracking.
%
%The main reason may be that there exists a limited amount of data to train deep networks in visual tracking because only the target state (i.e., position and scale) in the first frame is available.
One potential reason is that the training data is very limited as the target state (i.e., position and scale) is only available in the first frame.
Several methods address this issue by learning a generic representation offline from auxiliary training images.
Fan et al.~\cite{DBLP:journals/tnn/FanXWG10} learn a specific feature extractor with CNNs from an offline training set (about 20000 image pairs) for human tracking.
Wang and Yeung~\cite{DBLP:conf/nips/WangY13} pre-train a multi-layer autoencoder network on the part of the 80M tiny image \cite{DBLP:journals/pami/TorralbaFF08} in an unsupervised fashion.
Using a video repository~\cite{DBLP:conf/nips/ZouNZY12}, Wang et al.~\cite{DBLP:journals/tip/WangLWCY15} learn video features by imposing temporal constraints. 
To alleviate the issues with offline training, the DeepTrack~\cite{DBLP:conf/bmvc/LiLP14} and CNT~\cite{DBLP:journals/tip/ZhangL0Y16} methods incrementally learn target-specific CNNs without pre-training.
\textcolor{black}{Note that existing tracking methods based on deep networks~\cite{DBLP:conf/bmvc/LiLP14,DBLP:journals/tip/WangLWCY15,DBLP:journals/tip/ZhangL0Y16} 
use two or fewer convolutional layers to represent target objects, 
and do not fully exploit rich hierarchical features.}

Recently, CNNs pre-trained on the large-scale ImageNet~\cite{DBLP:conf/cvpr/DengDSLL009} dataset have been used as feature extractors for numerous vision problems. 
To compute an output response map (i.e., a heat map indicating the probability of the target offset), deep features from pre-trained CNNs can be regressed to three types of labels: 
(i) \emph{binary labels}~\cite{DBLP:conf/icml/HongYKH15,DBLP:journals/corr/WangLGY15,DBLP:conf/cvpr/NamH16,DBLP:conf/cvpr/TaoGS16}, 
(ii) \emph{soft labels} generated by a Gaussian function~\cite{DBLP:conf/iccv/WangOWL15,DBLP:conf/cvpr/QiZQYHL016,DBLP:conf/eccv/DanelljanRKF16,DBLP:conf/eccv/BertinettoVHVT16}, and 
(iii) \emph{bounding box parameters}~\cite{DBLP:conf/eccv/HeldTS16}. 
The deep trackers~\cite{DBLP:conf/icml/HongYKH15,DBLP:journals/corr/WangLGY15,DBLP:conf/cvpr/NamH16,DBLP:conf/cvpr/TaoGS16} using binary labels often draw multiple samples around the estimated target position in the current frame 
and inevitably suffer from the sampling ambiguity issue. 
The MDNet~\cite{DBLP:conf/cvpr/NamH16}  tracker uses a 
negative-mining scheme to alleviate this issue and achieves favorable performance 
on recent benchmark datasets.
In contrast, the proposed approach is built on adaptive correlation filters which regress 
circularly shifted samples with soft labels to mitigate sampling ambiguity. 
Compared to the \emph{sparse} response values at sampled states in~\cite{DBLP:conf/cvpr/NamH16}, our algorithm generates a \emph{dense} response map.
%
%The recently developed FCNT tracker~\cite{DBLP:conf/iccv/WangOWL15} 
%also exploits the deep feature hierarchies for visual tracking.
%
%Our approach differs in the use of learning adaptive correlation filters over multiple convolutional layers.
%
In contrast to the FCNT~\cite{DBLP:conf/iccv/WangOWL15} tracker which uses two convolutional layers for visual tracking, we exploit the feature hierarchies of deep networks by learning adaptive correlation filters over multiple convolutional layers. 
The proposed algorithm resembles the recent HDT~\cite{DBLP:conf/cvpr/QiZQYHL016} and C-COT~\cite{DBLP:conf/eccv/DanelljanRKF16} trackers in learning correlation filters over multiple convolutional layers. 
The main differences lie in explicitly exploring the deep feature hierarchies for precisely locating target objects as well as on exploiting region proposals to address the issues with 
scale estimation and target re-detection from tracking failures. 

{\flushleft {\bf Tracking by Region Proposals.}}
%\subsection{Tracking by Region Proposals} 
Region proposal methods~\cite{cvpr14Cheng,DBLP:conf/eccv/ZitnickD14,iccv11Van,DBLP:conf/eccv/ZitnickD14,iccv15Ghodrati} provide candidate regions (in bounding boxes) for object detection and recognition.
By generating a relatively small number of candidate regions (compared to exhaustive sliding windows), region proposal methods enable the use of CNNs for classification~\cite{DBLP:conf/cvpr/GirshickDDM14}.
Several recent methods exploit region proposal algorithms for visual tracking.
Hua et al.~\cite{DBLP:conf/iccv/HuaAS15} compute the objectness scores~\cite{DBLP:conf/eccv/ZitnickD14} to select highly confident sampling proposals as tracking outputs.
Huang et al.~\cite{bmvc15huang} apply region proposals to refine the estimated position and scale changes of target objects.
In~\cite{axiv15gao}, Gao et al. improve the Struck~\cite{DBLP:conf/iccv/HareST11} tracker using region proposals.
Similar to \cite{bmvc15huang,axiv15gao}, we use region proposals to generate candidate bounding boxes.
The main difference is that we learn a correlation filter with long-term memory of target appearance to compute the confidence score of every proposal.
In addition, we tailor the EdgeBox~\cite{DBLP:conf/eccv/ZitnickD14} method to generate two types of proposals for scale estimation and target re-detection, respectively.

\begin{figure}[t]
\centering
\small
\setlength{\tabcolsep}{1pt}
\includegraphics[width=.45\textwidth]{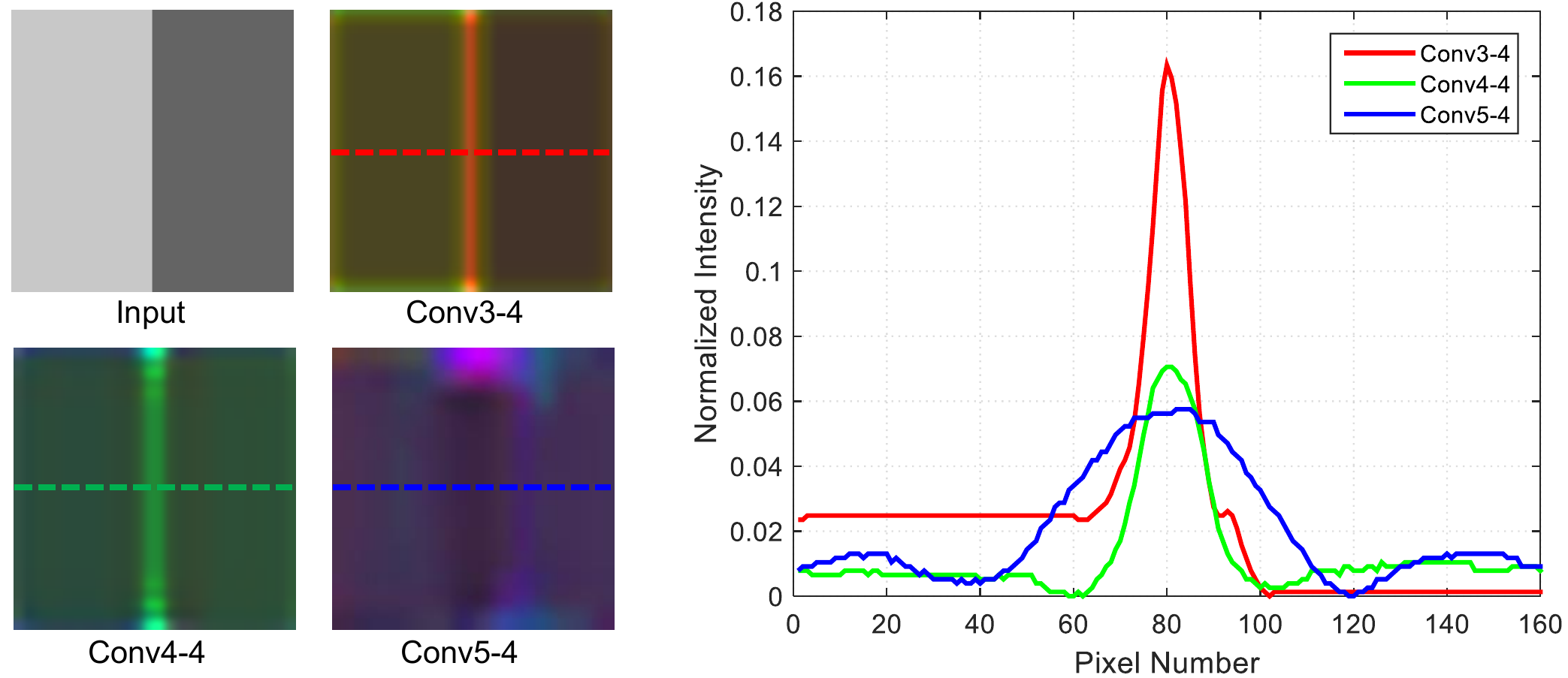}
\vspace{-2mm}
\caption{\tb{Spatial resolution of CNN features.} Visualization of the CNN features of a toy image with a horizontal step edge using VGGNet~\cite{DBLP:journals/corr/SimonyanZ14a}.
We visualize each CNN layer by converting its first three principle components of convolutional channels to RGB values. Intensities of the dash lines are visualized on the right.
Note that the \textit{conv5-4} layer is less effective in locating the step edge due to its low spatial resolution, while the \textit{conv3-4} layer is more useful for precise localization.
}
\label{fig:step}
\vspace{-4mm}
\end{figure}

\begin{figure*}[t]
\centering
\includegraphics[width=.65\textwidth]{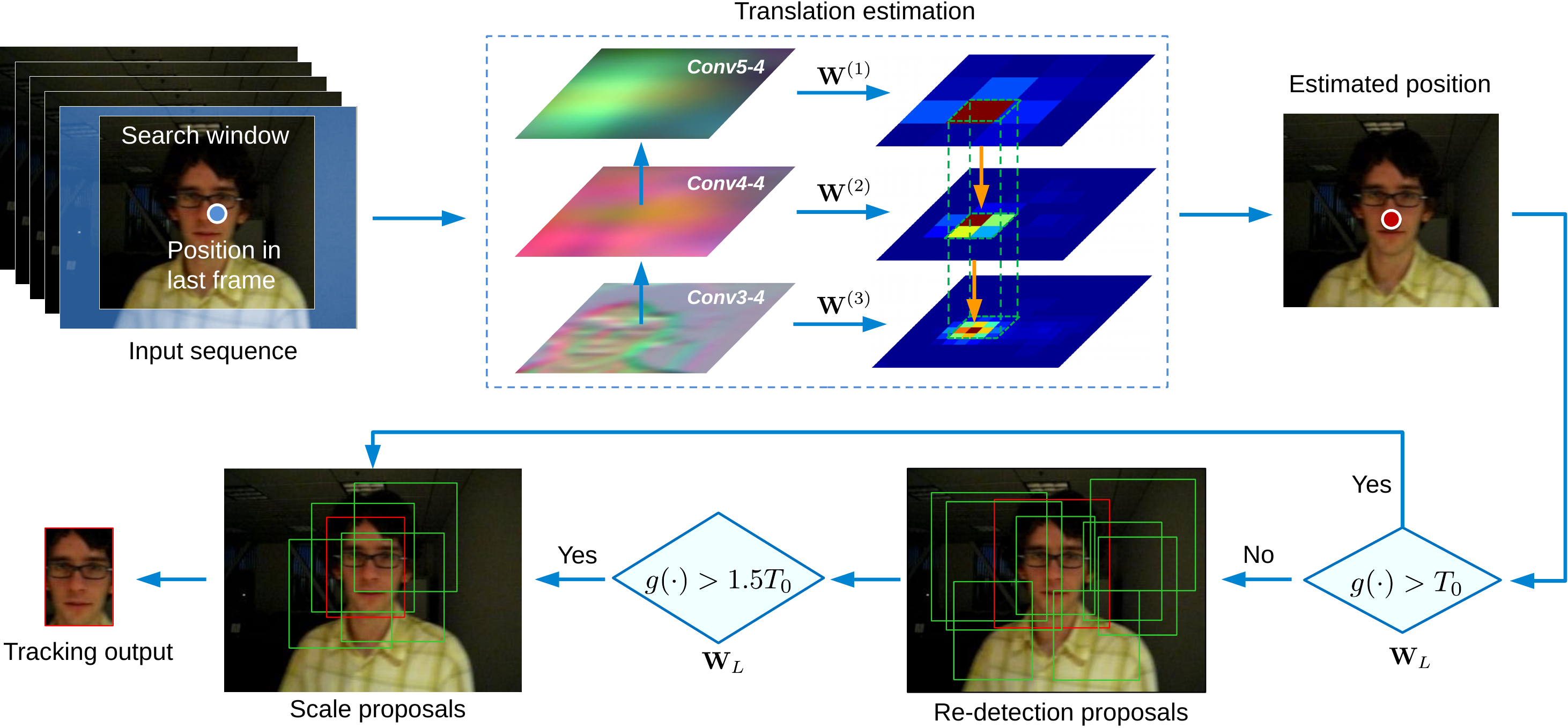}
\vspace{-2mm}
\caption{\tb{Main steps of the proposed algorithm.}
Given an image, we first crop the search window centered at the estimated position in the previous frame.
We use the third, fourth and fifth convolutional layers as our target object representations (Section~\ref{sec:feature}).
Each layer indexed by $i$ is then convolved with the learned linear correlation filter $\mathbf{w}^{(i)}$ to generate a response map, 
whose location of the maximum value indicates the estimated target position (Section~\ref{sec:cf}). 
We search the multi-level response maps to infer the target location in a coarse-to-fine fashion (Section~\ref{sec:estimation}).
Centered at the estimated target position, we crop an image patch using the same scale in the previous frame.
We apply the long-term memory filter $\mathbf{w}_L$ to compute the confidence score $g$ of this patch. 
We detect tracking failures by checking if the confidence score is below a given threshold $T_0$.
We then generate region proposals across the whole image and compute their confidence scores using $\mathbf{w}_L$. 
We search for the proposal with the highest confidence score as the re-detected result (Section~\ref{sec:proposals}). 
Note that we conservatively take each re-detected result by setting the confidence threshold to $1.5 T_0$. 
As for scale estimation, we generate region proposals with a smaller step size to make them tightly around the estimated target position (Section~\ref{sec:proposals}). 
We infer the scale change by searching for the proposal with the highest confidence score (using the long-term memory filter $\mathbf{w}_L$ to compute confidence scores). 
}
\label{fig:top2down} 
\vspace{-4mm}
\end{figure*}

\vspace{-2mm}
\section{Overview}
In this work, we propose the hierarchical correlation features-based tracker (HCFT*) 
based on the observation that deeper convolutional layers of CNNs encode semantic abstraction of the target object and the corresponding 
feature responses are robust to appearance variations.
On the other hand, features at early layers retain more fine-grained spatial details and 
are useful for precise localization. 
In Figure~\ref{fig:step}, we show an illustrative example --- an image with a horizontal step edge. 
We visualize the CNN activations from the third, fourth, and fifth convolutional layers by converting their first three main components of convolutional channels to RGB values. 
The intensity plot of three CNN layers in Figure~\ref{fig:step} shows that the features at the 
fifth convolutional layer are less effective in locating the sharp boundary due to its low spatial resolution, while the features at the third layer are more useful to locate the step edge precisely.
Our goal is to exploit the best of both semantics and fine-grained details for robust visual object tracking.
Figure~\ref{fig:top2down} illustrates the main steps of our tracking algorithm. For translation estimation, we learn adaptive linear correlation filters over the outputs of each convolutional layer and perform a coarse-to-fine search over the multi-level correlation response maps 
to infer the locations of targets.
Centered at the estimated target position, we crop an image patch with the same scale in the previous frame. 
We apply the learned long-term memory correlation filter $\mathbf{w}_L$ to compute the confidence score of this patch. 
When the confidence score is below a given threshold $T_0$ (i.e., when tracking failures may occur), we generate region proposals across the whole image and compute their confidence scores using $\mathbf{w}_L$.
We search for the proposal with the highest confidence score as the re-detected result. 
Note that we take the re-detected result for training the long-term memory filter $\mathbf{w}_L$ by setting a larger confidence threshold, e.g., $1.5 T_0$. 
For scale estimation, we generate region proposals tightly around the estimated target position.
We compute their confidence scores again using the long-term memory filter $\mathbf{w}_L$ and search for the proposal with the highest confidence score to infer the scale change. 

\begin{figure}
\centering
\small`
\setlength{\tabcolsep}{1pt}
\begin{tabular}{cccc}
\includegraphics[trim = 0mm 0mm 10mm 2mm, clip,width=.11\textwidth]{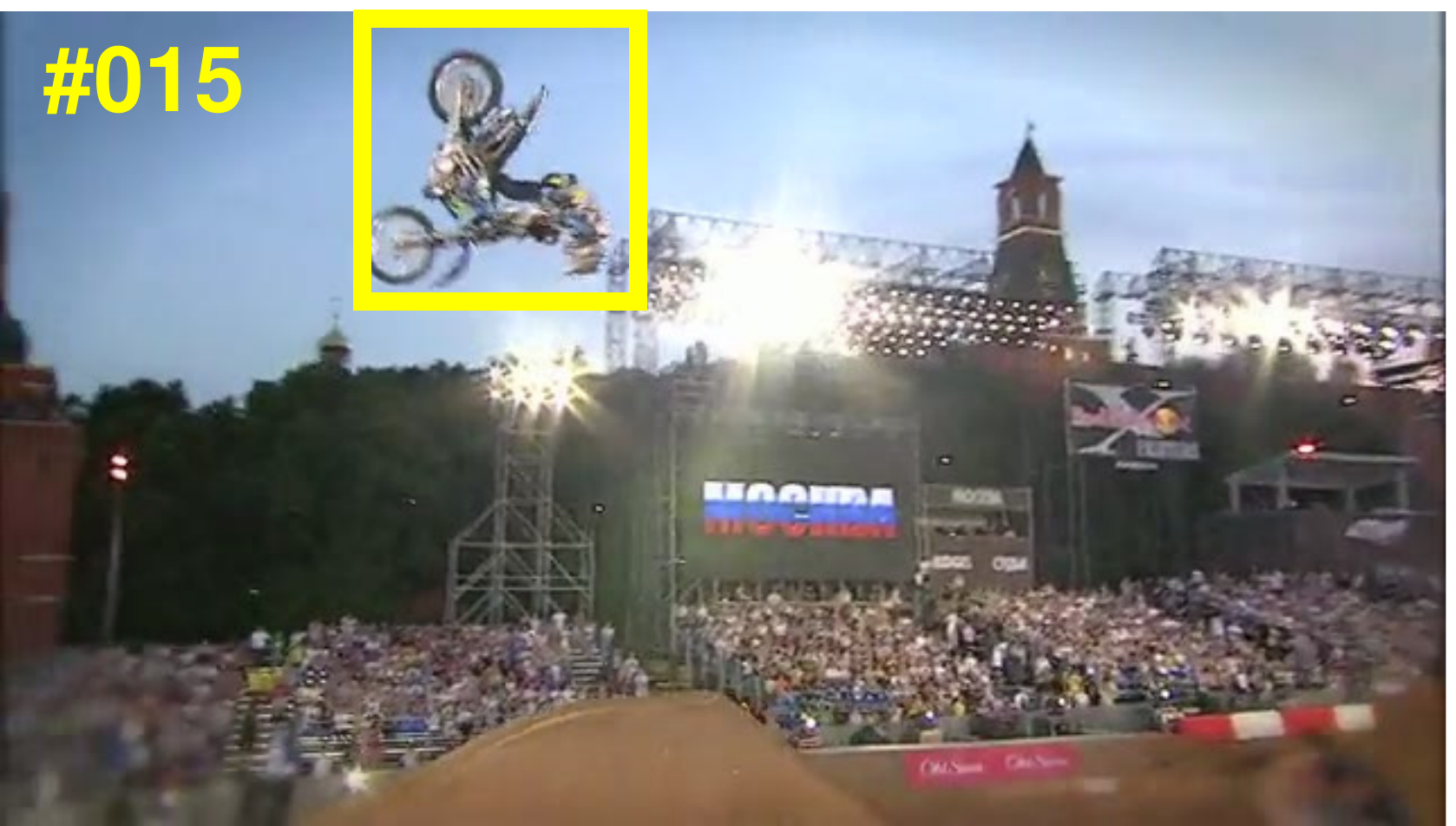}&
\includegraphics[trim = 0mm 0mm 10mm 2mm, clip,width=.11\textwidth]{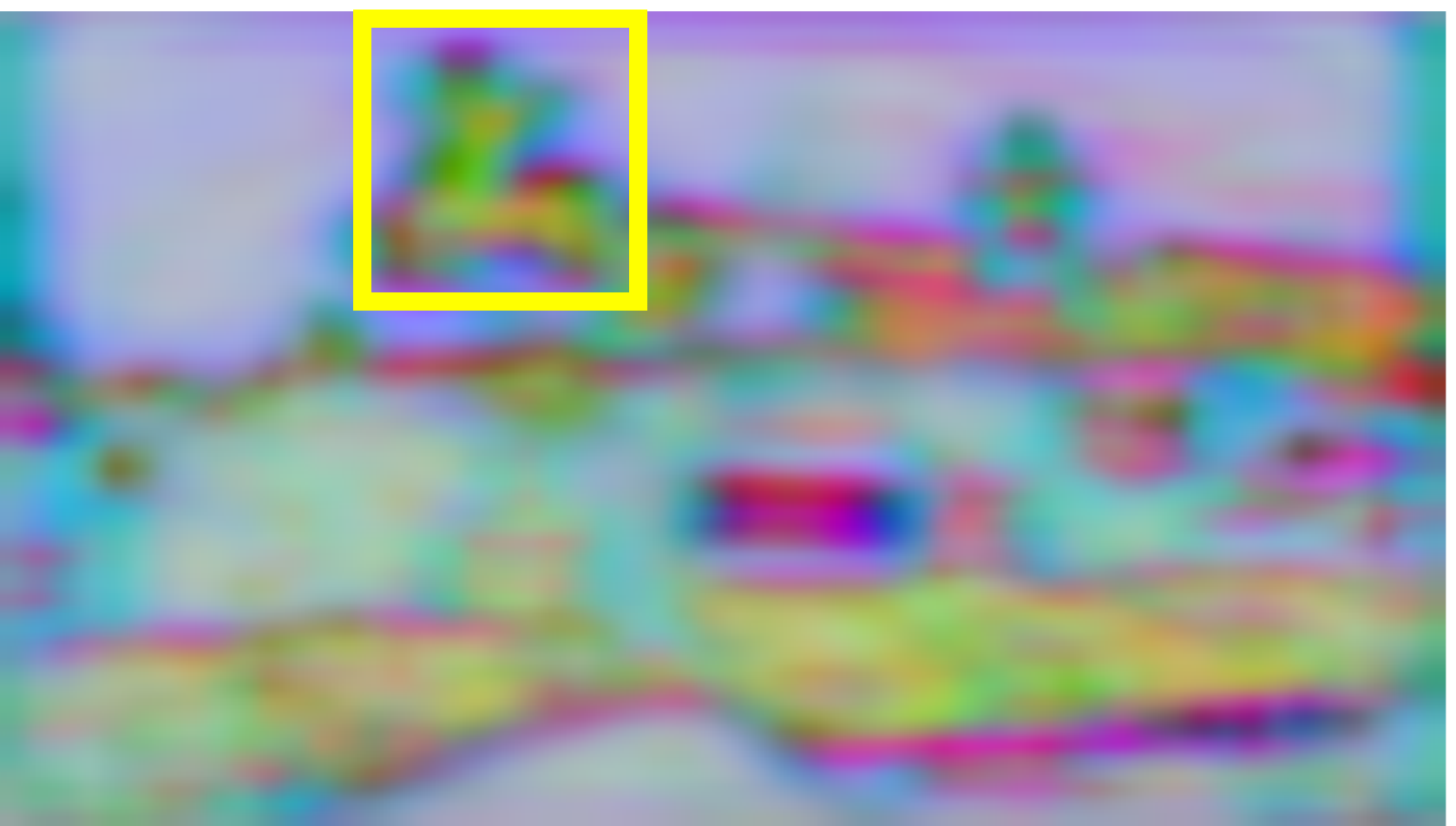} &
\includegraphics[trim = 0mm 0mm 10mm 2mm, clip,width=.11\textwidth]{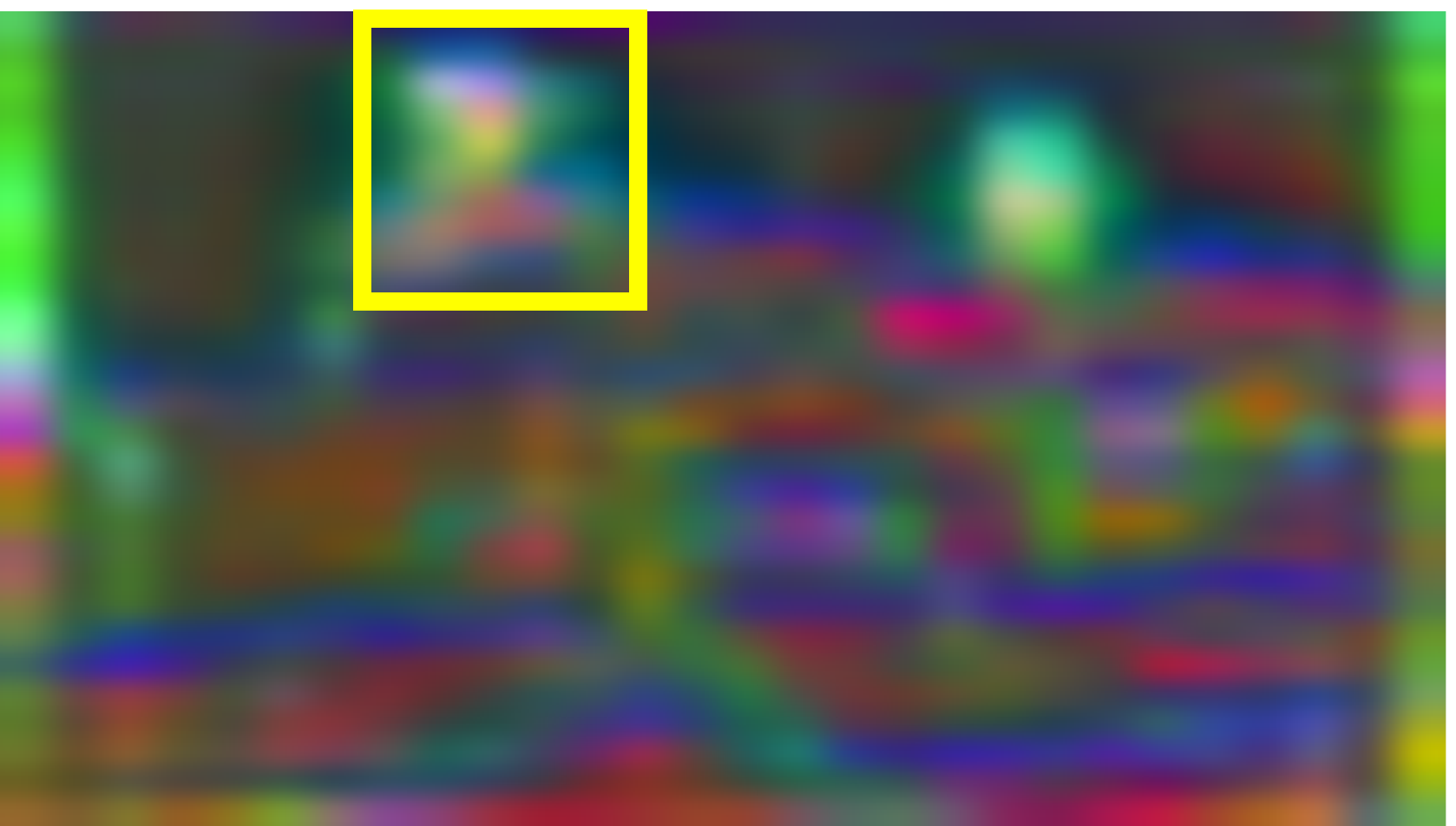}&
\includegraphics[trim = 0mm 0mm 10mm 2mm, clip,width=.11\textwidth]{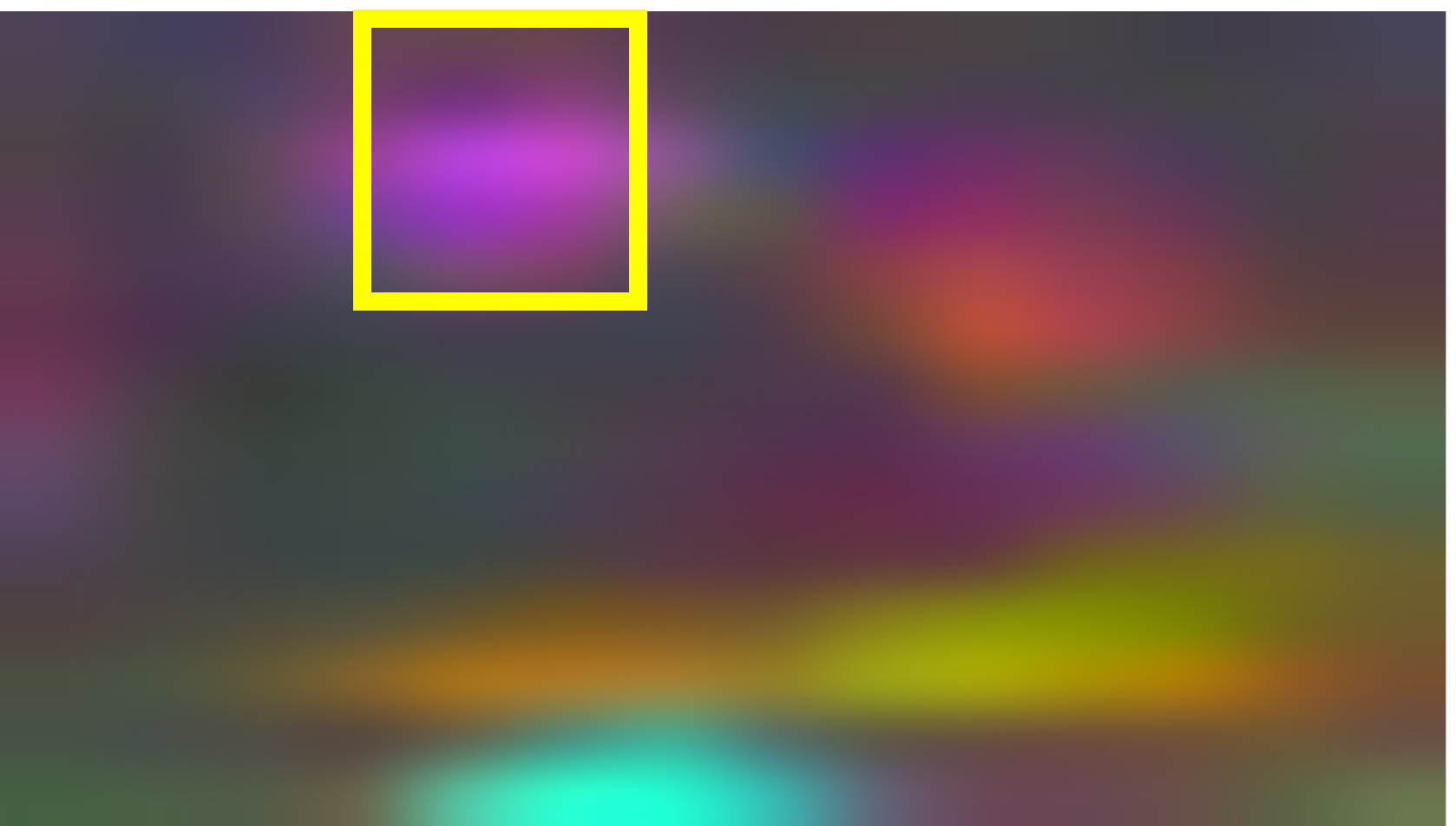}\\
\includegraphics[trim = 0mm 0mm 10mm 2mm, clip,width=.11\textwidth]{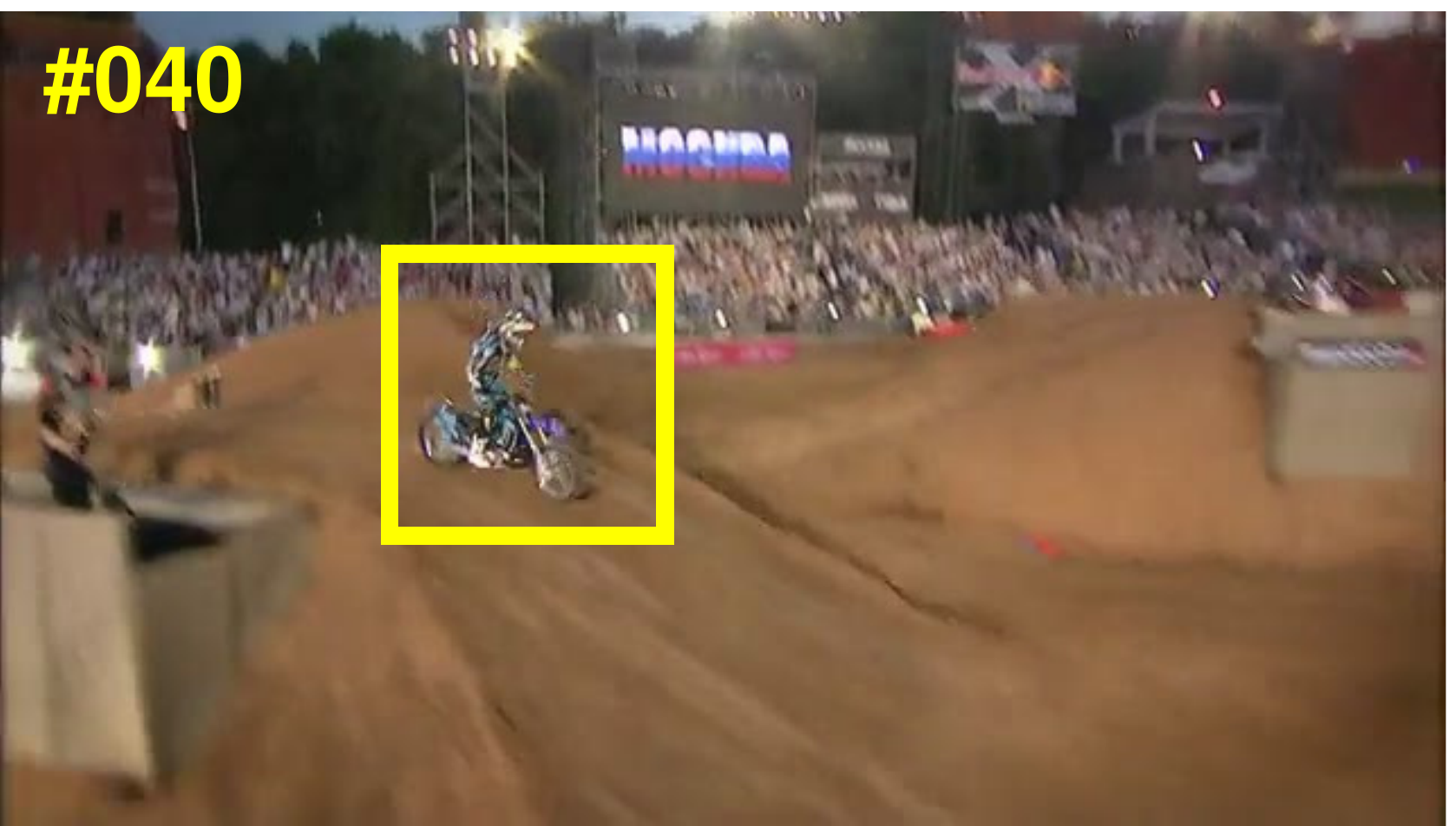}&
\includegraphics[trim = 0mm 0mm 10mm 2mm, clip,width=.11\textwidth]{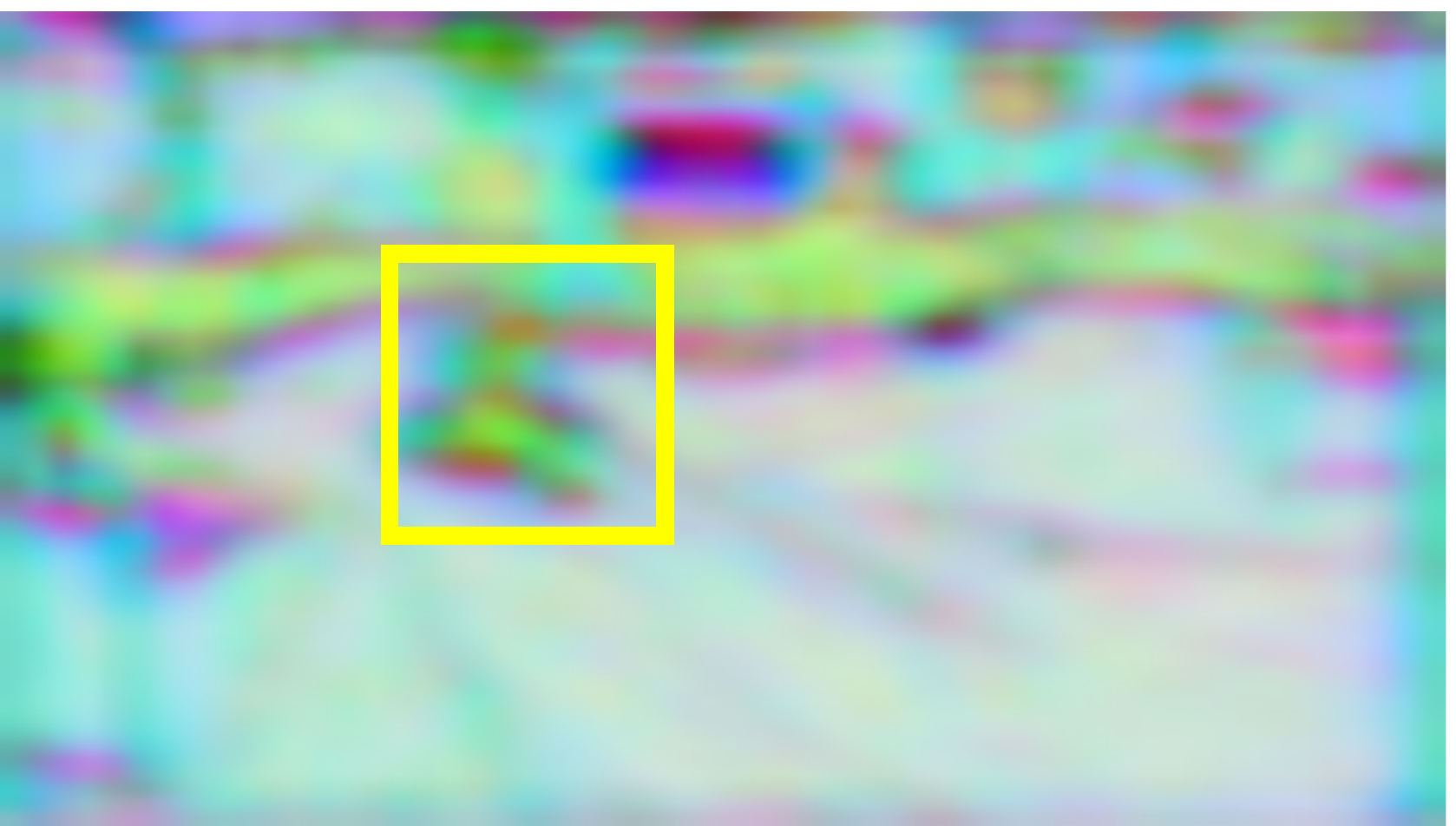}&
\includegraphics[trim = 0mm 0mm 10mm 2mm, clip,width=.11\textwidth]{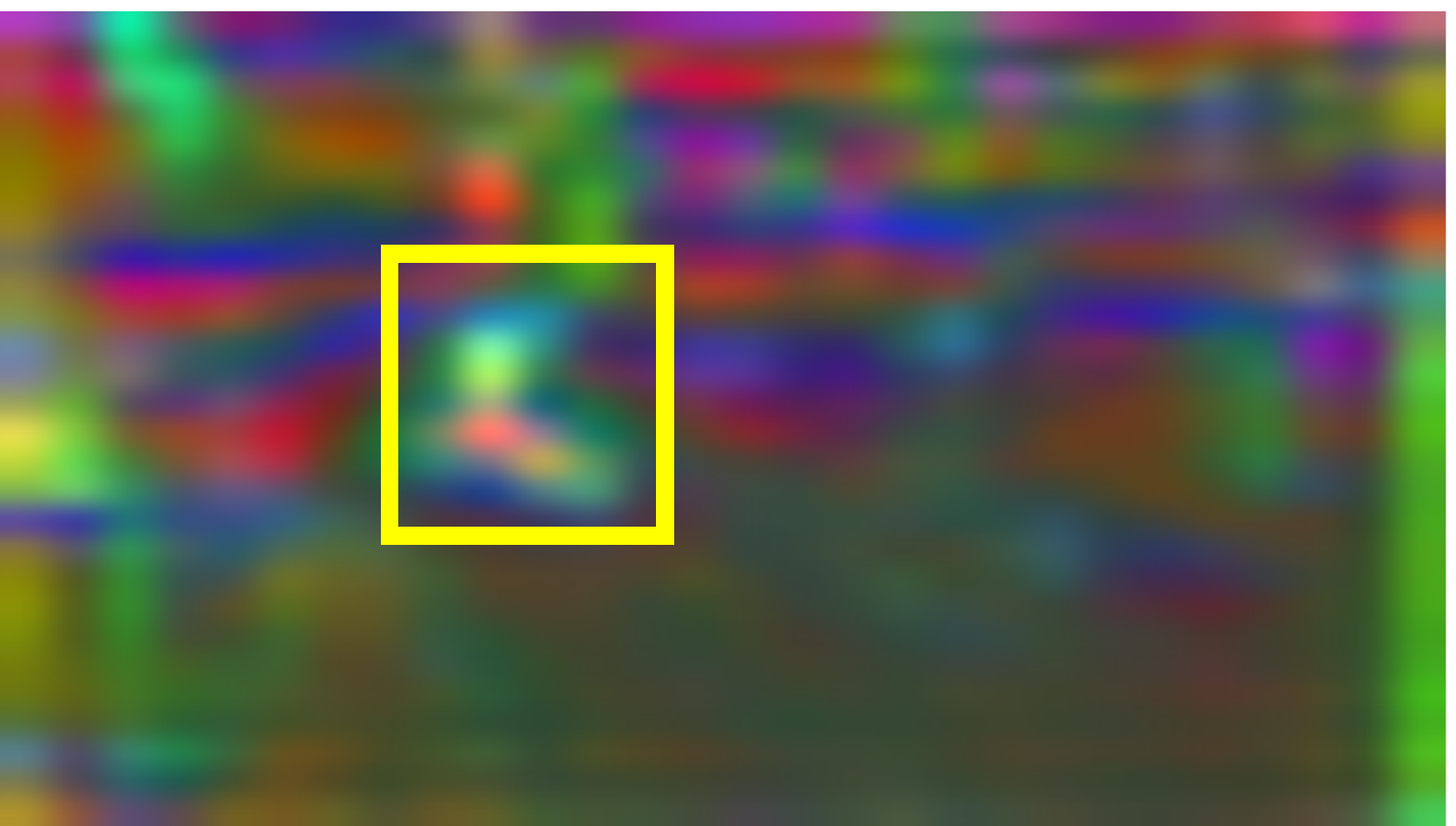}&
\includegraphics[trim = 0mm 0mm 10mm 2mm, clip,width=.11\textwidth]{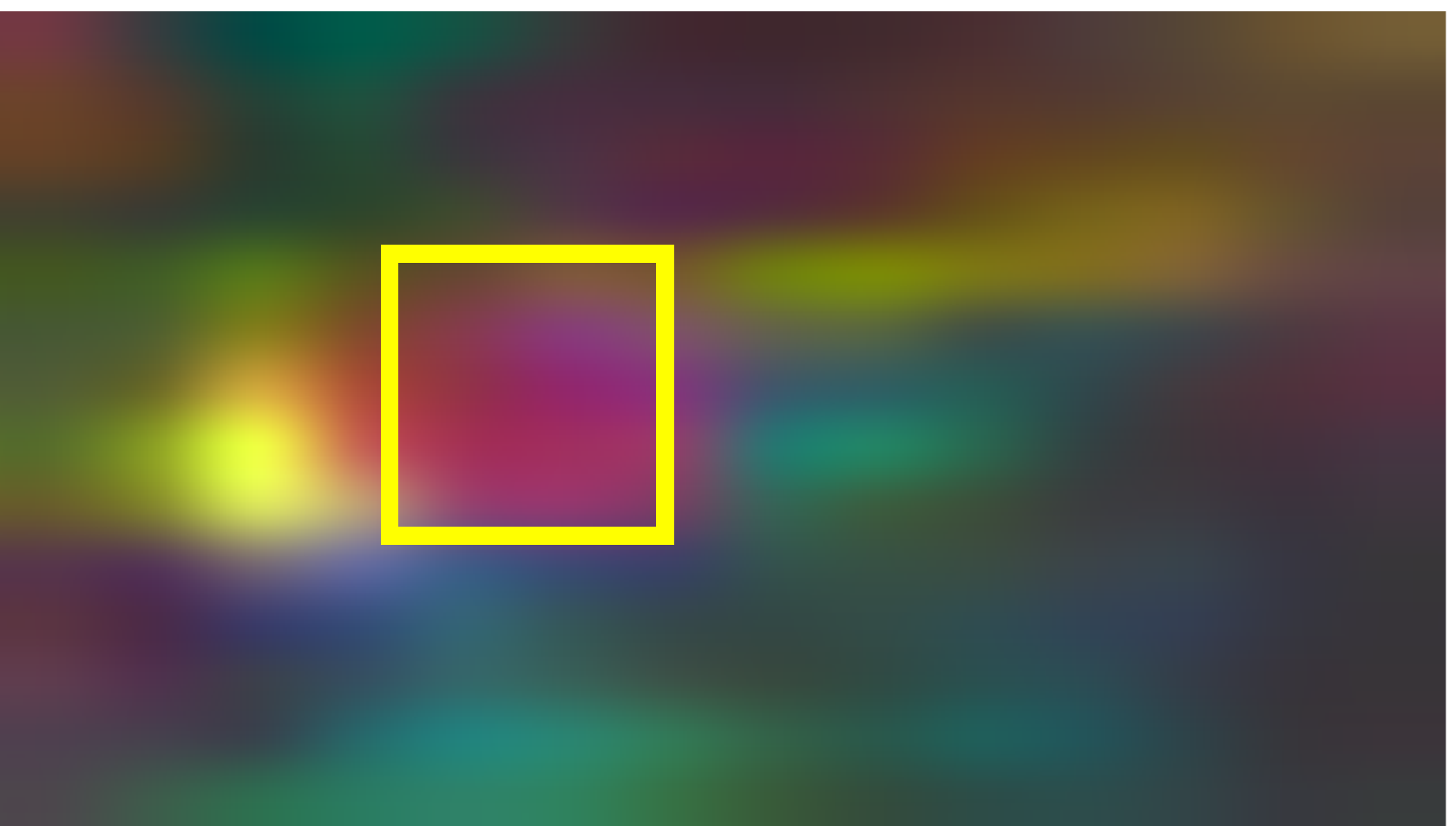}\\
\includegraphics[trim = 0mm 0mm 10mm 2mm, clip,width=.11\textwidth]{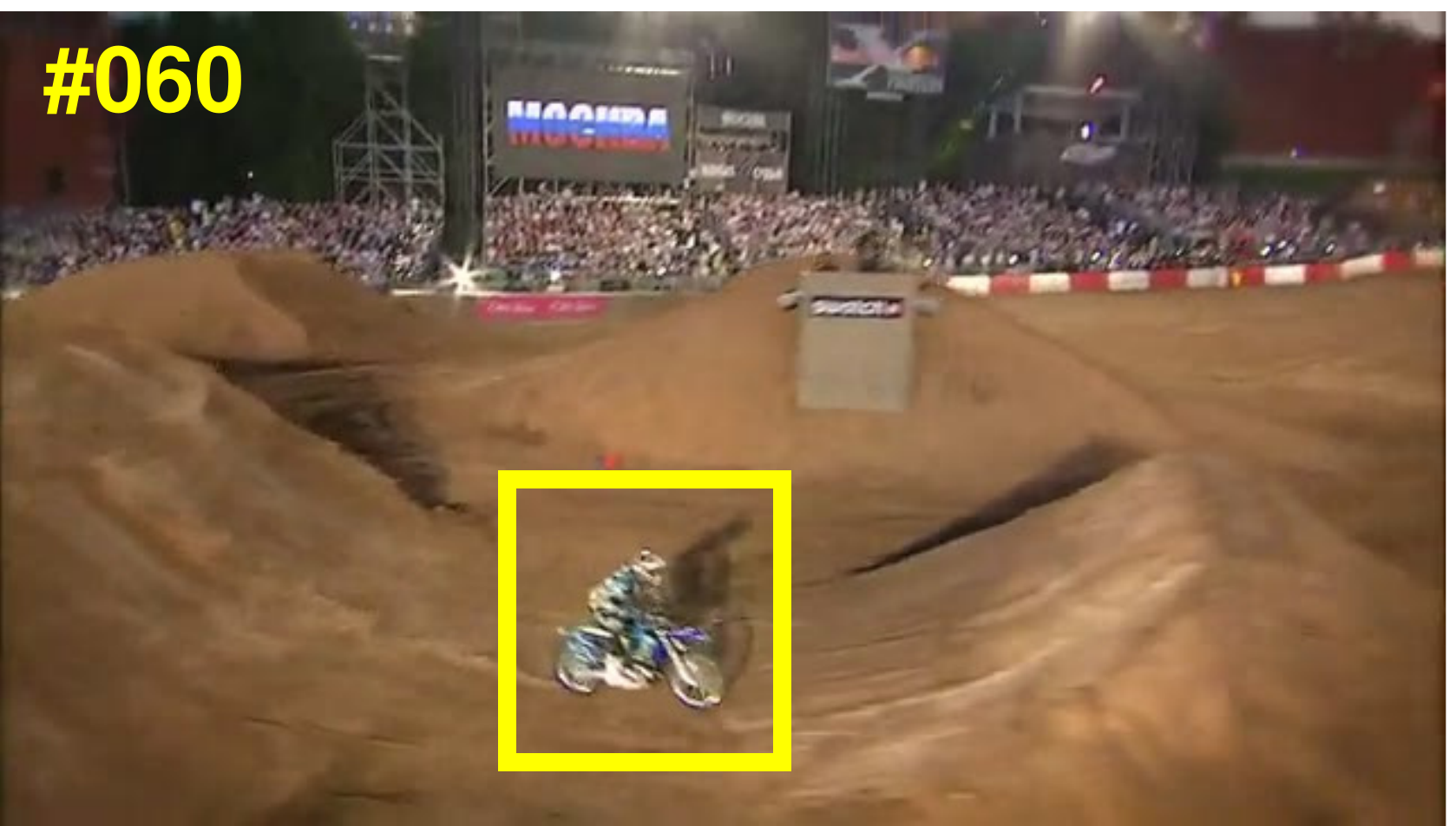}&
\includegraphics[trim = 0mm 0mm 10mm 2mm, clip,width=.11\textwidth]{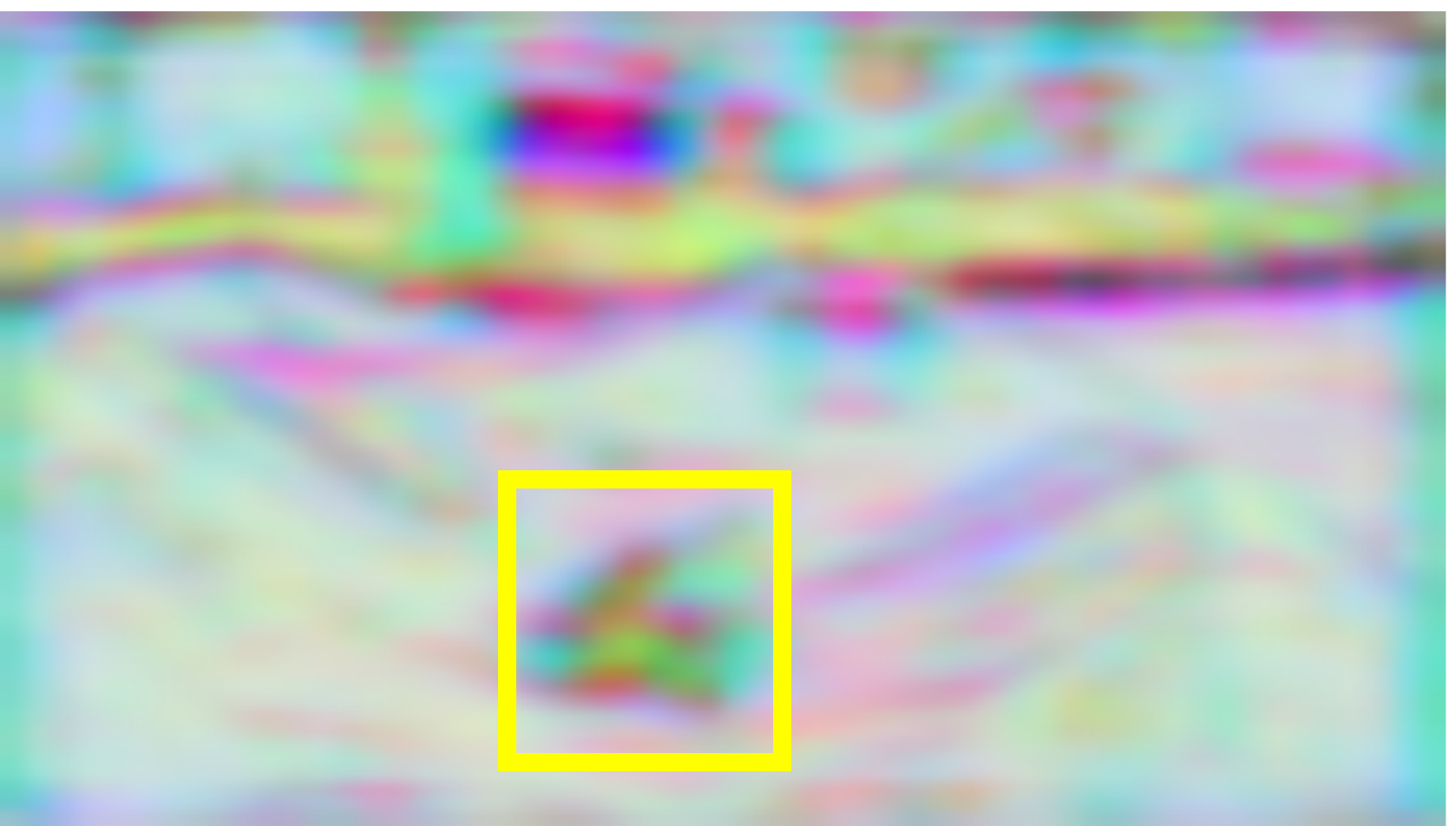}&
\includegraphics[trim = 0mm 0mm 10mm 2mm, clip,width=.11\textwidth]{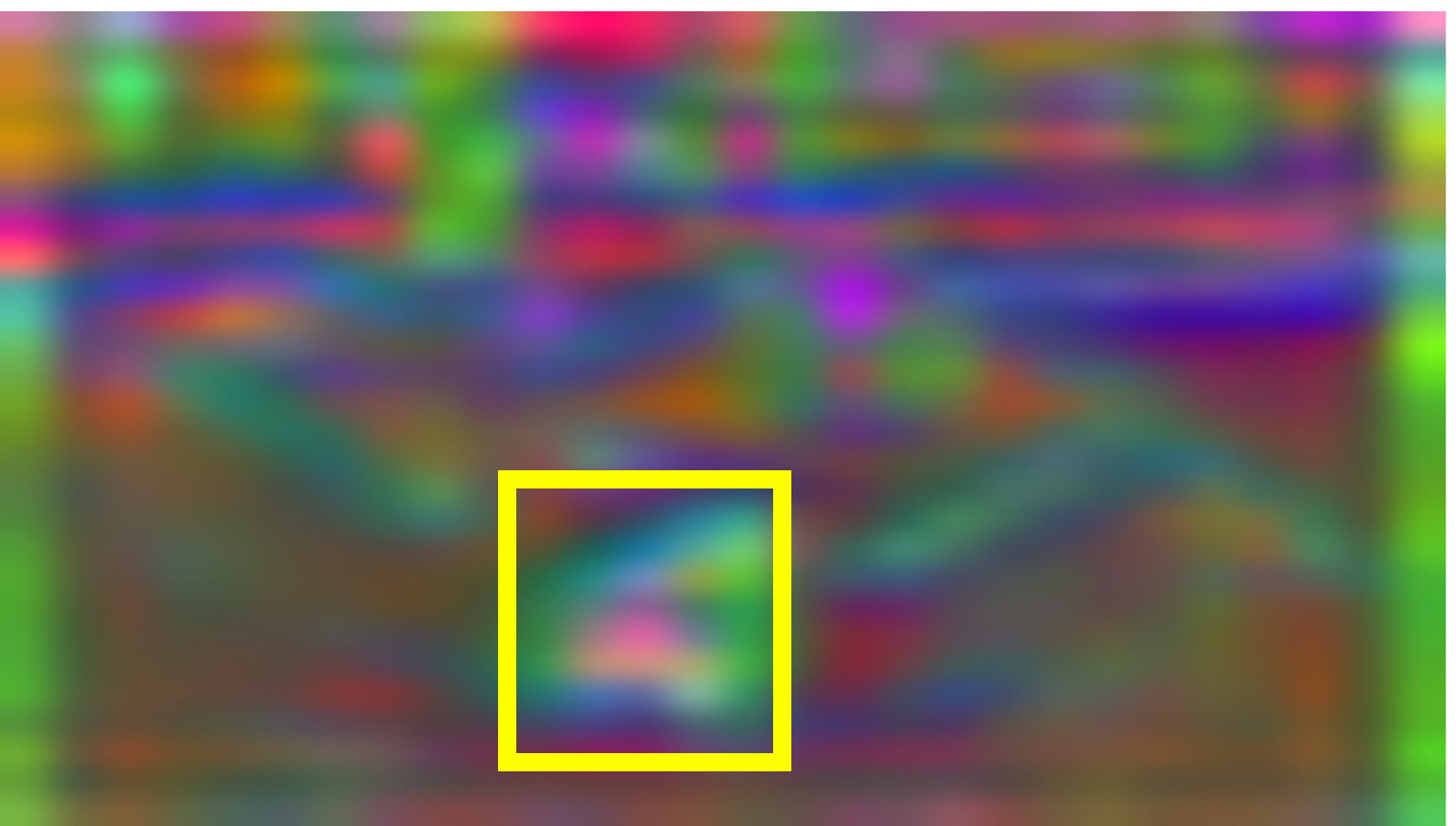}&
\includegraphics[trim = 0mm 0mm 10mm 2mm, clip,width=.11\textwidth]{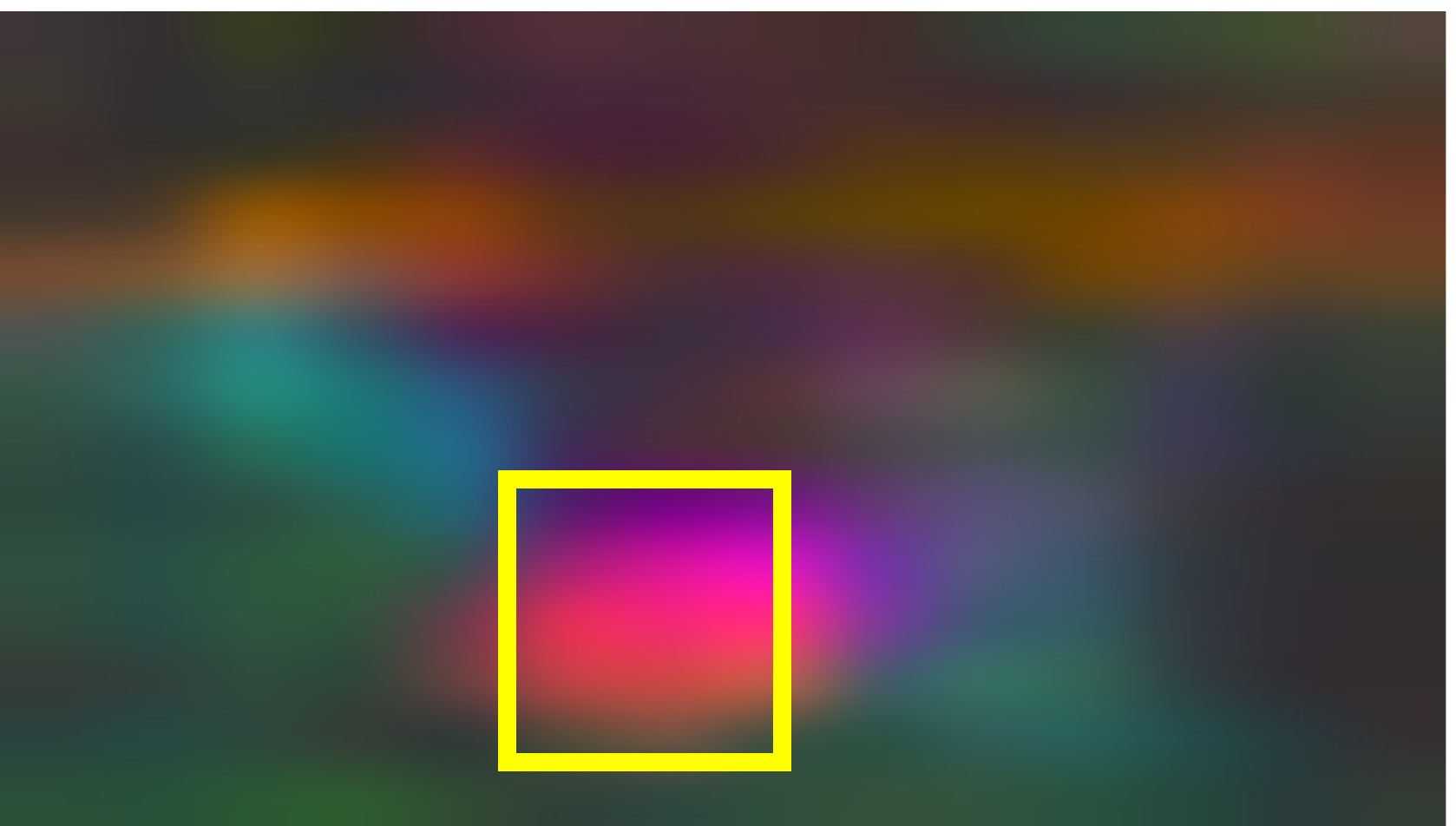}\\
\includegraphics[trim = 0mm 0mm 10mm 2mm, clip,width=.11\textwidth]{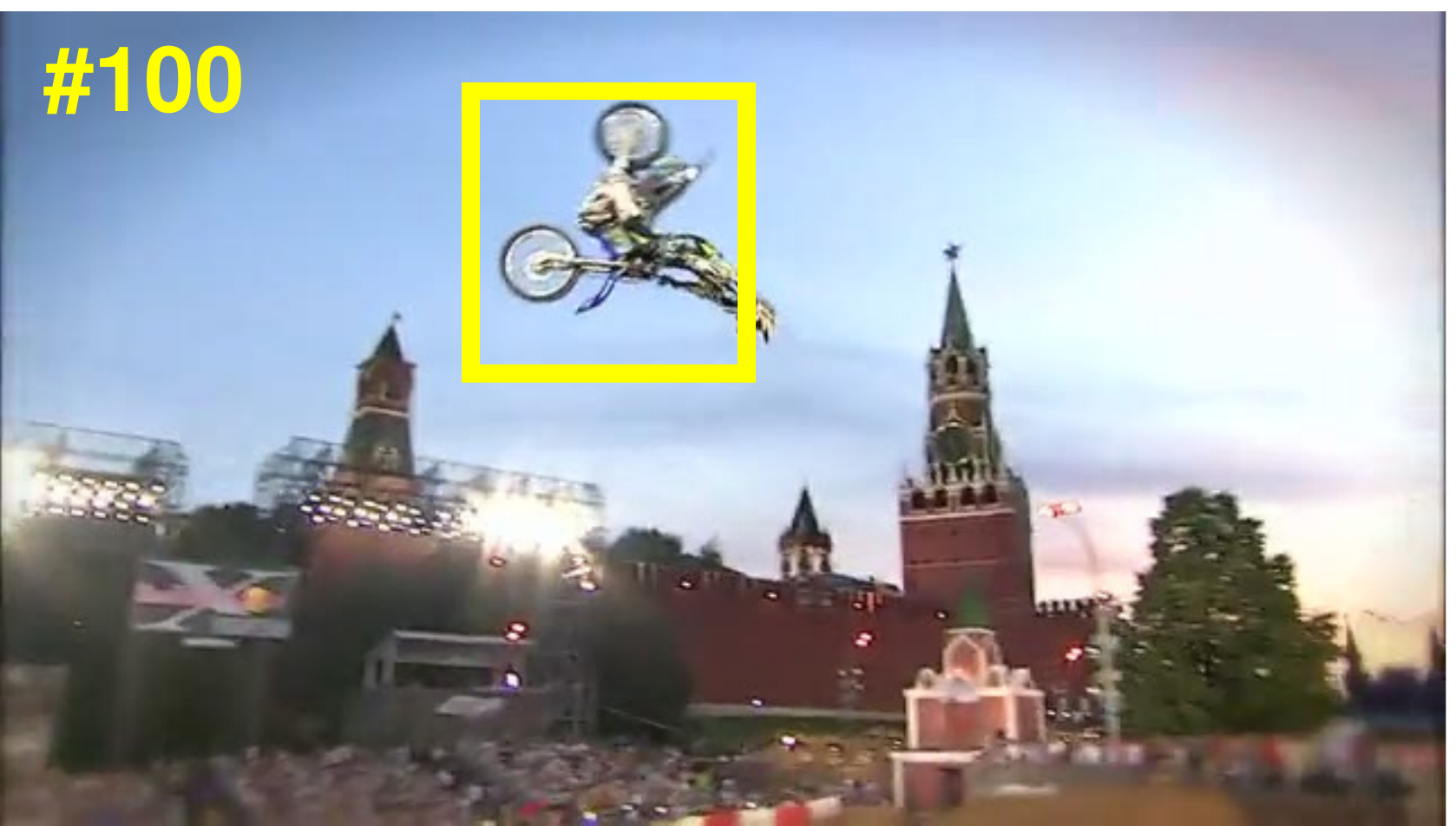}&
\includegraphics[trim = 0mm 0mm 10mm 2mm, clip,width=.11\textwidth]{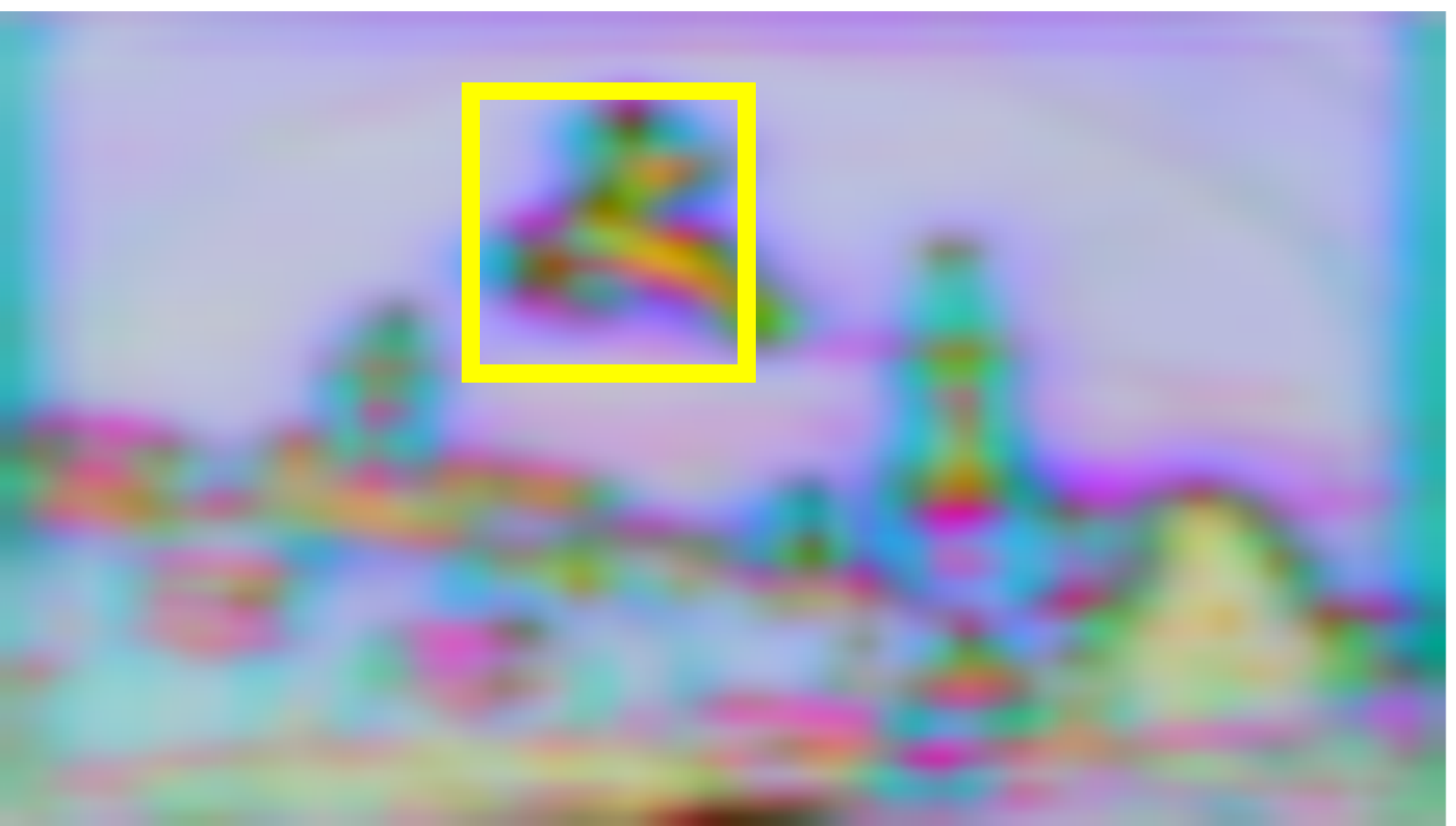}&
\includegraphics[trim = 0mm 0mm 10mm 2mm, clip,width=.11\textwidth]{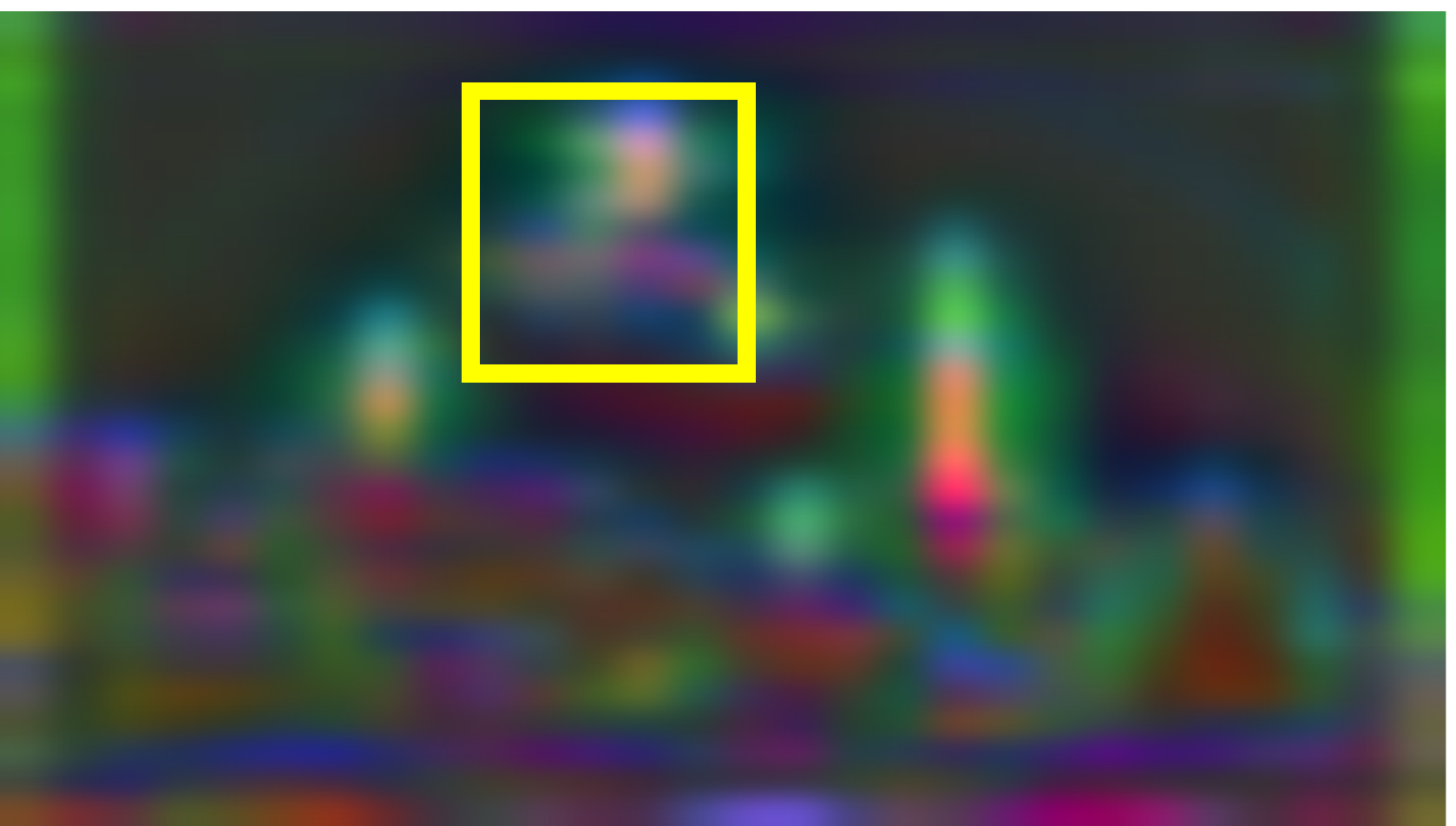}&
\includegraphics[trim = 0mm 0mm 10mm 2mm, clip,width=.11\textwidth]{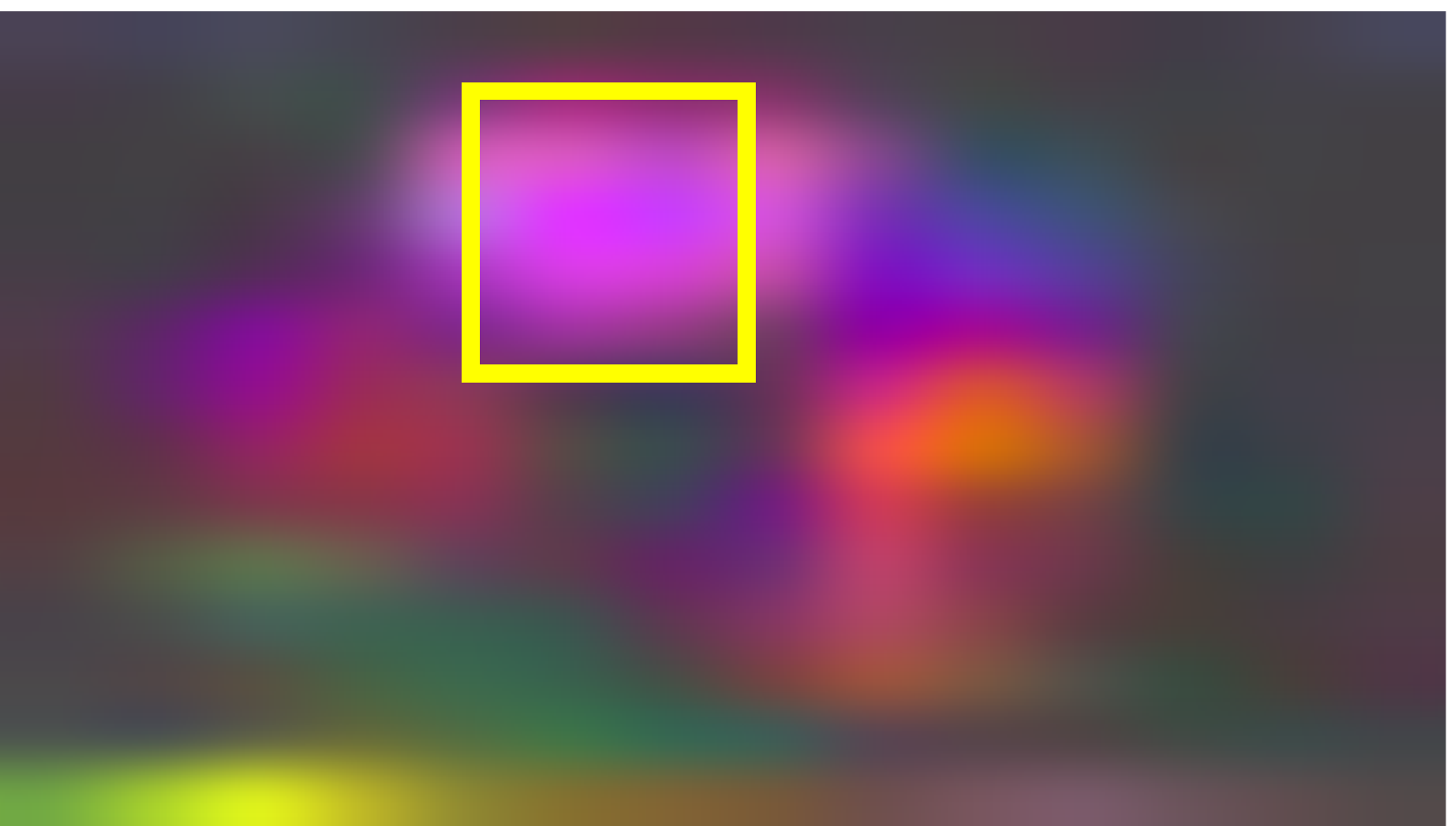}\\
\includegraphics[trim = 1mm 1mm 1mm 1mm, clip, width=.11\textwidth]{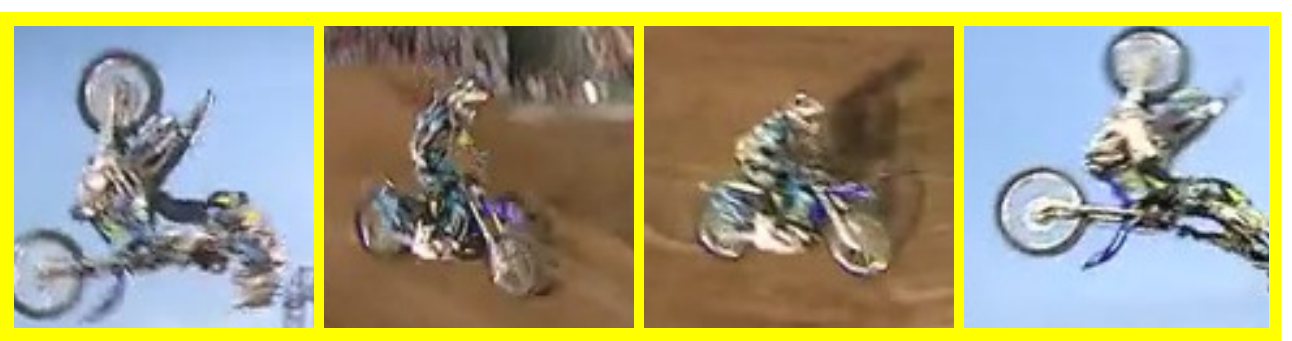}&
\includegraphics[trim = 1mm 1mm 1mm 1mm, clip, width=.11\textwidth]{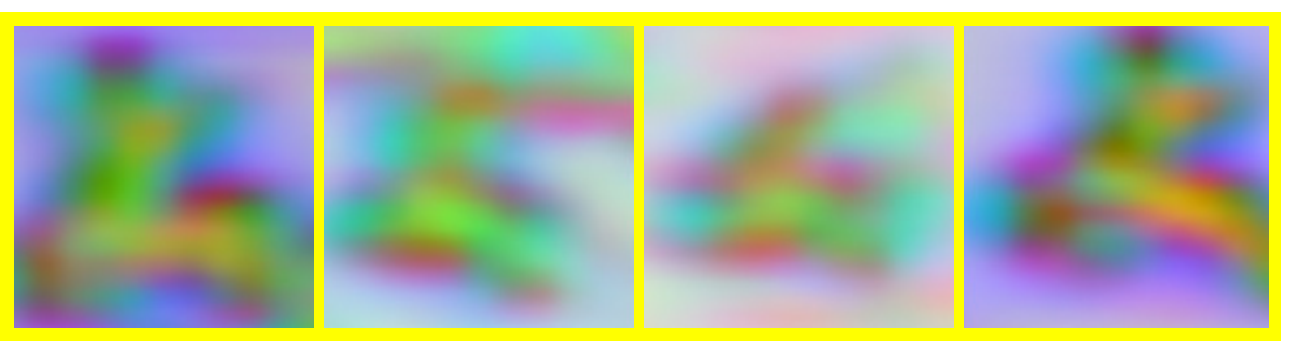}&
\includegraphics[trim = 1mm 1mm 1mm 1mm, clip, width=.11\textwidth]{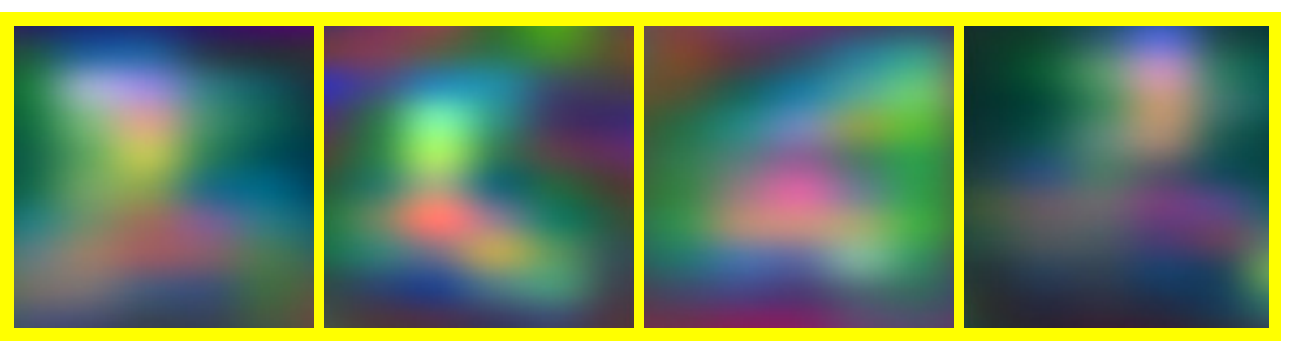}&
\includegraphics[trim = 1mm 1mm 1mm 1mm, clip, width=.11\textwidth]{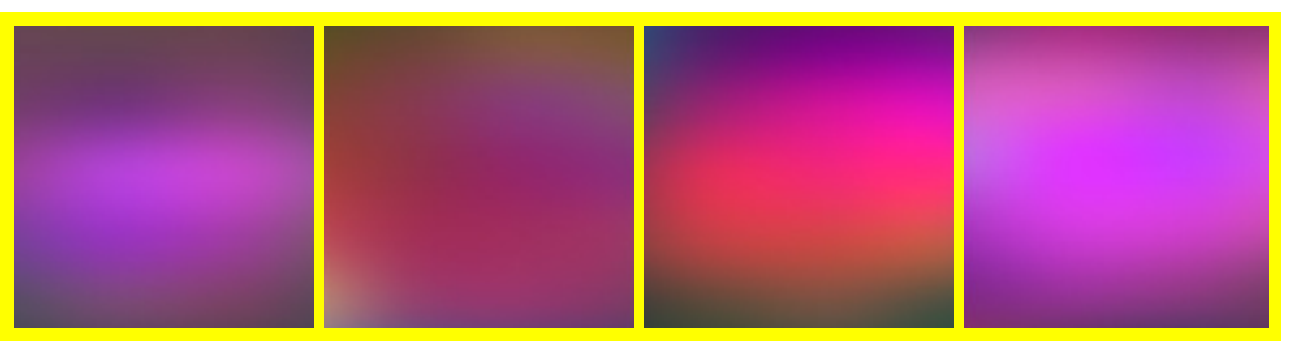}\\ 
(a) Input & (b) \textit{conv3-4} & (c) \textit{conv4-4} & (d) \textit{conv5-4} 
\end{tabular}
\caption{\tb{Visualization of convolutional layers.}
(a) Four frames from the challenging \textit{MotorRolling} sequence. 
(b)-(d) Features are from the outputs of the convolutional layers \textit{conv3-4}, \textit{conv4-4}, and \textit{conv5-4} using the VGGNet-19~\cite{DBLP:journals/corr/SimonyanZ14a}. 
The yellow bounding boxes indicate the tracking results by our method.
Note that although the appearance of the target changes significantly, the features 
using the output of the \textit{conv5-4} convolution layer (d) is able to discriminate it readily even the background has dramatically changed.
The \textit{conv4-4} (c) and \textit{conv3-4} (b) layers encode more fine-grained details and are useful to locate target precisely.
}
\label{fig:vis}
\vspace{-4mm}
\end{figure}

\vspace{-2mm}
\section{Proposed Algorithm}
We first present our approach for robust object tracking which includes
extracting CNN features, learning correlation filters, and 
developing a coarse-to-fine search strategy.
Next, we present the schemes for target re-detection and scale estimation using region proposals.
Finally, we introduce the scheme to incrementally update two types of correlation filters with different learning rates.

\vspace{-2mm}
\subsection{Hierarchical Convolutional Features}
\label{sec:feature}
We use the convolution feature maps from a CNN, e.g., AlexNet~\cite{DBLP:conf/nips/KrizhevskySH12} or VGGNet~\cite{DBLP:journals/corr/SimonyanZ14a}, to encode target appearance. 
The semantic discrimination between objects from different categories is strengthened as features are propagated to deeper layers, while there exists a gradual reduction of spatial resolution (see also Figure~\ref{fig:cnnlayer}).
For visual object tracking, we are interested in accurate locations of a target object rather than its semantic category.
We thus remove the fully-connected layers as they show little spatial resolution of $1\times1$ pixel and exploit only the hierarchical features in the convolutional layers.

Due to the use of the pooling operations, the spatial resolution of a target object 
is gradually reduced with the increase of the CNN depth.
For example, the convolutional feature maps of \textit{pool5} in the VGGNet are of 
$7\times7$ pixels, which is $\frac{1}{32}$ of the input image size of $224\times224$ pixels.
As it is not feasible to locate objects accurately with such low spatial resolution, 
we alleviate this issue by resizing each feature map to a fixed larger size using bilinear interpolation. 
Let $\mathbf{h}$ denote the feature map and $\mathbf{x}$ be the upsampled feature map.
The feature vector for the $i$-th location is:
\begin{equation}
\mathbf{x}_i=\sum_k\alpha_{ik}\mathbf{h}_k,
\label{equ:upsample}
\end{equation}
where the interpolation weight $\alpha_{ik}$ depends on the position of $i$ and $k$ neighboring feature vectors, respectively.
%
%Note that this interpolation takes place in the spatial domain and can be viewed as one kind of deconvolution process widely used for semantic image segmentation~\cite{DBLP:conf/cvpr/LongSD15,DBLP:conf/iccv/NohHH15}. 
\textcolor{black}{Note that this
	interpolation takes place in the spatial domain.} % and shares a same goal with the deconvolution process widely used for semantic
	%image segmentation~\cite{DBLP:conf/cvpr/LongSD15,DBLP:conf/iccv/NohHH15}.}
%
In Figure~\ref{fig:vis}, we use the pre-trained VGGNet-19~\cite{DBLP:journals/corr/SimonyanZ14a} as our feature extractor and visualize the upsampled outputs of the \textit{conv3-4}, \textit{conv4-4}, and \textit{conv5-4} layers on the \textit{MotorRolling} sequence. 
For the features in each convolutional layer (e.g., the \textit{conv5-4} layer has 512 channels), we convert the first three principal components of convolutional channels to RGB values.
As shown in Figure~\ref{fig:vis}(a), the target motorcyclist has significant appearance variations over time. 
Figure~\ref{fig:vis}(d) shows that the fifth convolutional features within the target bounding boxes are more or less close to dark red throughout different frames, 
while they are discriminative from background areas despite the dramatic background changes. 
This property of the \textit{conv5-4} layer allows us to deal with significant appearance changes and accurately locate the target at a coarse-grained level.
In contrast, the \textit{conv4-4} (Figure~\ref{fig:vis}(c)) and \textit{conv3-4} (Figure~\ref{fig:vis}(b)) layers encode more spatial details and help locate the target precisely on the fine-grained level. 
We note that this insight is also exploited for segmentation and fine-grained localization using CNNs~\cite{DBLP:journals/cvpr/HariharanAGM14a}, where features from multiple layers are combined by concatenation.
Since this representation ignores the coarse-to-fine feature hierarchy in the CNN architecture, we find that it does not perform well for visual tracking as shown in our experimental validation (see Section~\ref{subsec:fa}).

There are two main advantages of using pre-trained CNNs as feature extractor. 
First, as pre-trained CNNs do not require fine-tuning models, extracting CNN features becomes easier and more efficient.
Second, as pre-trained CNN models are learned from classifying 1000 object categories, the representations can describe an unseen target object well for model-free tracking.

\vspace{-2mm}
\subsection{Correlation Filters}
\label{sec:cf}
Typically, correlation trackers~\cite{DBLP:conf/cvpr/BolmeBDL10, DBLP:conf/eccv/HenriquesCMB12, DBLP:conf/cvpr/DanelljanKFW14, DBLP:conf/eccv/ZhangZLZY14, DBLP:conf/bmvc/DanelljanKFW14}
regress the circularly shifted versions of the input features to soft target scores generated by a Gaussian function, and locates target objects by searching for the maximum value on the response map. 
The underlying assumption of these methods is that the circularly shifted versions of input features 
approximate the dense samples of the target object at different locations~\cite{DBLP:journals/pami/HenriquesC0B15}. 
As learning correlation filters do not require binary (hard-threshold) samples, 
these trackers effectively alleviate the sampling ambiguity problem that adversely affects 
most tracking-by-detection approaches. 
By exploiting the complete set of shifted samples, correlation filters can be 
efficiently trained with a substantially large number of training samples using the fast Fourier transform (FFT).
This data augmentation helps discriminate the target from its surrounding background.
In this work, we use the outputs of each convolutional layer as multi-channel features~\cite{DBLP:conf/iccv/GaloogahiSL13,DBLP:conf/cvpr/BoddetiKK13,DBLP:journals/pami/HenriquesC0B15}.

Let $\mathbf{x}$ be the $l$-th layer of the feature vector of size $M \times N \times D$, 
where $M$, $N$, and $D$ indicates the width, height, and the number of feature channels, respectively. 
Here we denote $\mathbf{x}^{(l)}$ concisely as $\mathbf{x}$ and ignore the $M$, $N$, and $D$ on the layer index $l$.
We consider all the circularly shifted versions of the feature $\mathbf{x}$ along the $M$ and $N$ dimensions as training samples. 
Each shifted sample $\mathbf{x}_{ij}$, $(i,j)\in\{0,1,
\ldots, M-1\}\times\{0,1, \ldots, N-1\}$,
has a Gaussian function label $y_{ij}=e^{-\frac{(i-M/2)^2+(j-N/2)^2}{2\sigma^2}}$, where $\sigma$ is the kernel width. 
At the target center with zero shifts, we have the maximum score, $y_{(\frac{M}{2},\frac{N}{2})} = 1$.
When the position $(i, j)$ is gradually away from the target center, the score $y_{ij}$ decays rapidly from one to zero. 
We learn the correlation filter $\mathbf{w}$ with the same size of $\mathbf{x}$ by solving the following minimization problem: 
\begin{equation}
\label{equ:filter}
\mathbf{w}^*=\argmin_\mathbf{w}\sum_{i,j}\|\mathbf{w}\cdot\mathbf{x}_{ij} -y_{ij}\|^2+\lambda\|\mathbf{w}\|_2^2,
\end{equation}
where $\lambda$ is a regularization parameter ($\lambda\ge0$) and the linear product is defined as $\mathbf{w}\cdot\mathbf{x}_{ij}=\sum_{d=1}^{D}\mathbf{w}_{ijd}^\top\mathbf{x}_{ijd}$.
As the label $y_{ij}$ is softly defined, we no longer require hard-thresholded samples in filter learning.
The minimization problem in~\eqref{equ:filter} is the same as training the vector correlation filters using 
the FFT as described in~\cite{DBLP:conf/cvpr/BoddetiKK13}.
Let the capital letter be the corresponding Fourier transformed signal.
The learned filter in the frequency domain on the $d$-th ($d\in\{1,\ldots,D\}$) channel is
\begin{equation}
\mathbf{W}^d=\frac{\mathbf{Y}\odot\bar{\mathbf{X}}^d}{\sum_{n=1}^{D} \mathbf{X}^i\odot\bar{\mathbf{X}}^i+\lambda},
\label{equ:w}
\end{equation}
where $\mathbf{Y}$ is the Fourier transformation form of $\mathbf{y} = \bigl\{y_{ij} |(i,j)\in\{0,1,\ldots,M-1\}
\times\{0,1,\ldots,N-1\} \bigr\}$, and the bar denotes complex conjugation. 
The operator $\odot$ is the Hadamard (element-wise) product.
Given an image patch in the next frame, we denote $\mathbf{z}$ as the feature vector on the $l$-th layer and of size $M\times N\times D$.
We then compute the $l$-th correlation response map by
\begin{equation}
f(\mathbf{z})=\mathscr{F}^{-1}\bigl(\sum_{d=1}^{D}\mathbf{W}^d\odot\mathbf{Z}^d\bigr),
\label{equ:res}
\end{equation} 
where the operator $\mathscr{F}^{-1}$ denotes the inverse FFT transform. 
By searching for the position with the maximum value on the response map $f(\mathbf{z})$ with size $M\times N$, 
we can estimate the target location based on the $l$-th convolution layer.

\vspace{-2mm}
\subsection{Coarse-to-Fine Translation Estimation}
\label{sec:estimation}

Given the set of correlation response maps $\{f_l\}$, we hierarchically infer the target translation of each layer, i.e., the location with the maximum value in the last layer is used as a regularization to search for the maximum value of the earlier layer.
Let $f_l(m,n)$ be the response value at the position $(m,n)$ on the $l$-th layer, and $(\hat{m},\hat{n})=\argmax_{m,n}f_l(m,n)$ indicates the location of the maximum value of $f_l$. 
We locate the target in the $(l-1)$-th layer by:
\begin{align}
\label{eq:search}
\argmax_{m,n} ~ & ~ f_{l-1}(m,n) + \mu_l f_l(m,n), \\ 
\text{s.t.} ~~ & ~ |m-\hat{m}|+|n-\hat{n}| \le r. \nonumber 
\end{align}
The constraint indicates that we only search the $r\times r$ neighboring regions of $(\hat{m}, \hat{n})$ 
on the $(l-1)$-th correlation response map. 
The response values from the last layers are weighted by the regularization term $\mu_l$ and then are propagated to the response maps of early layers.
We finally estimate the target location by maximizing~\eqref{eq:search} on the layer with the finest spatial resolution.
In practice, we observe that the tracking results are not sensitive to the parameter $r$ of the neighborhood search constraint. 
This amounts to computing a weighted average of the response maps from multiple layers to infer the target location as follows:
\begin{equation}
\label{eq:sum}
\argmax_{m,n} \sum_l \mu_l f_l(m,n). \\ 
\end{equation}

To choose the weights for the response maps, we consider the following two factors.
First, we use larger weight for the response map from the last convolutional layers as they capture 
semantics that are robust to appearance changes. 
We empirically decrease the weight parameter $\mu_l$ by half from the $l$-th layer to the $(l-1)$-th layer:
\begin{equation}
\label{eq:weight}
\mu_l\propto 2^{l-5},
\end{equation}
where $l=5,4,3$. 
Second, we observe that correlation response maps from different convolutional layers are often with inconsistent ranges, e.g., the maximum values of the response map from the \textit{conv5-4} layer are generally smaller than that from the \textit{conv3-4} layer (see Figure \ref{fig:weight}(a)-(c)). % as shown in Figure~\ref{fig:kitesurf}.
We address this problem by setting the weight parameter $\mu_l$ to be inversely proportional to the maximum value of the $l$-th response map $f_l$ as:
\begin{equation}
\label{eq:softweight}
\mu_l\propto \frac{1}{\max(f_l)}.
\end{equation}

\begin{figure}[t]
	\centering
	\setlength{\tabcolsep}{.5pt}
	\begin{tabular}{cccc}
		\multicolumn{2}{c}{\includegraphics[trim = 0mm 0mm 17mm 0mm, clip, width=.22\textwidth]{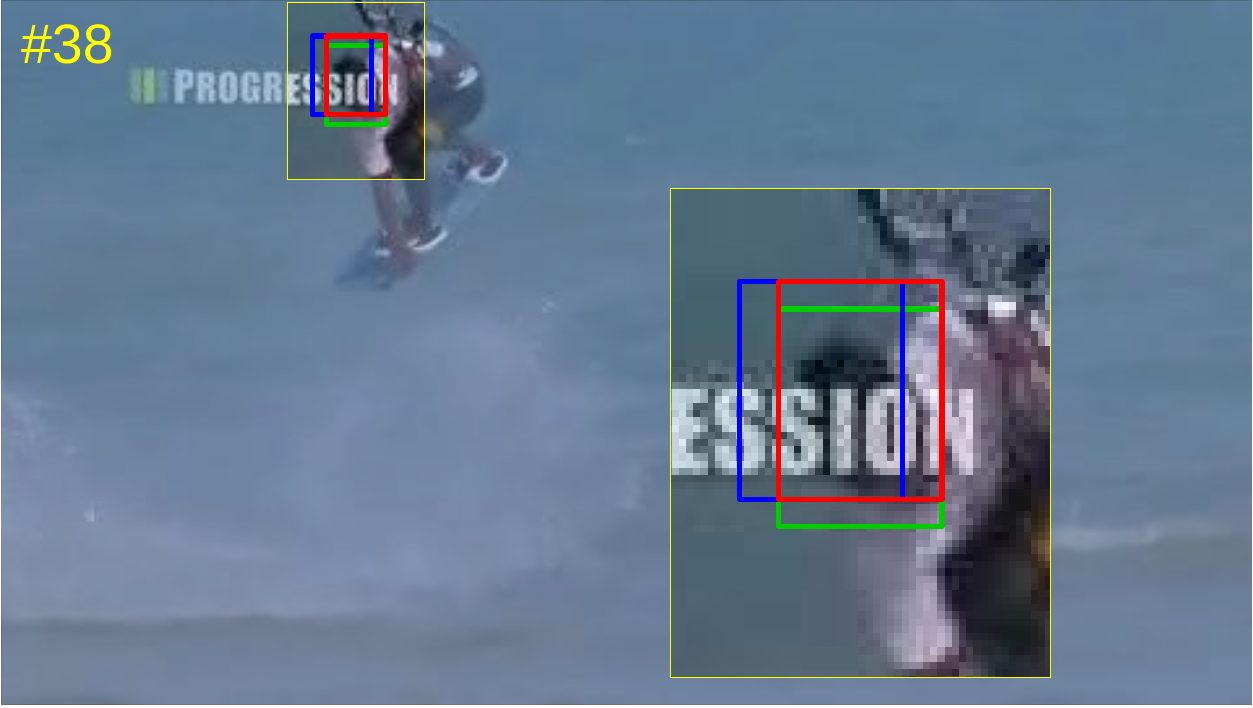}} &
		\includegraphics[width=.11\textwidth]{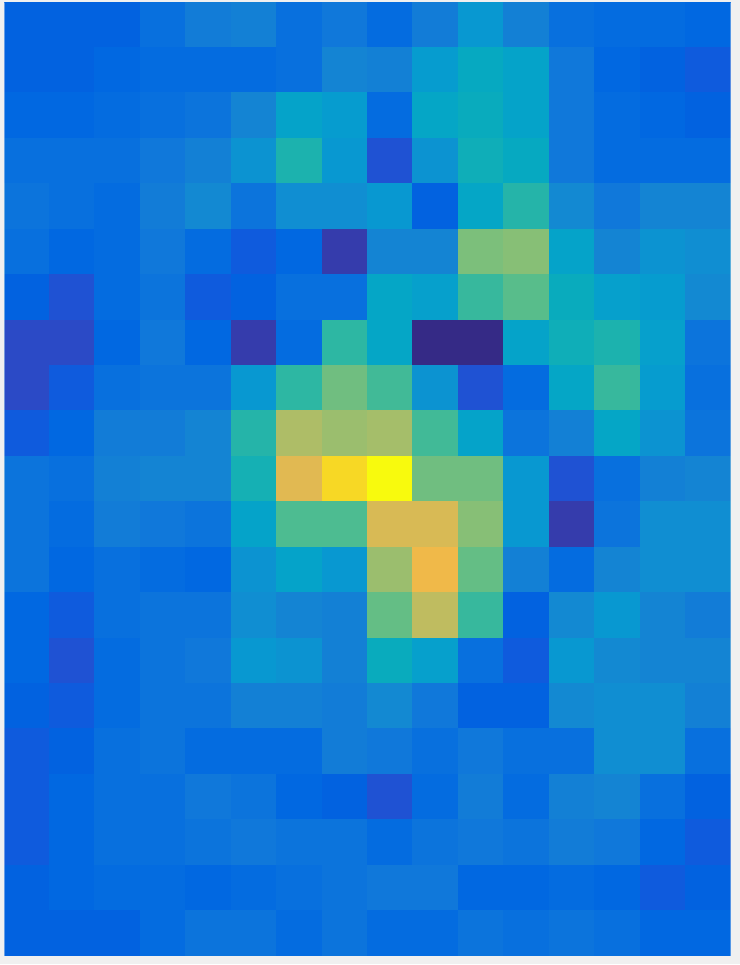} &
		\includegraphics[width=.11\textwidth]{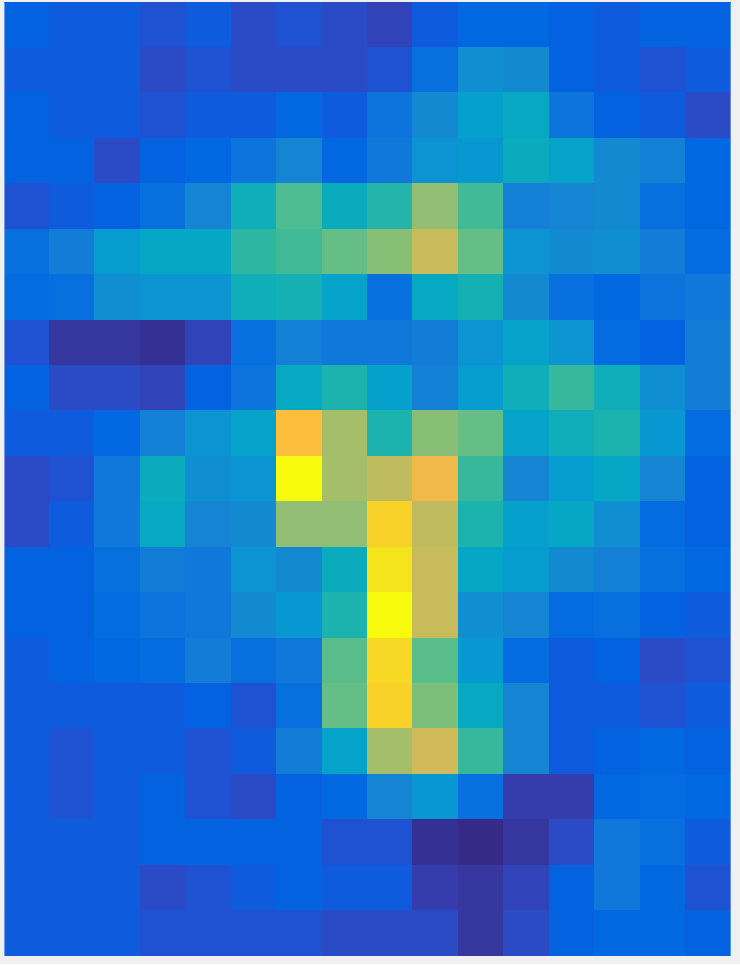} \\
		\multicolumn{2}{c}{(a) Input frame} & (b) \textit{conv5-4} & (c) \textit{conv4-4} \\ 
		\includegraphics[width=.11\textwidth]{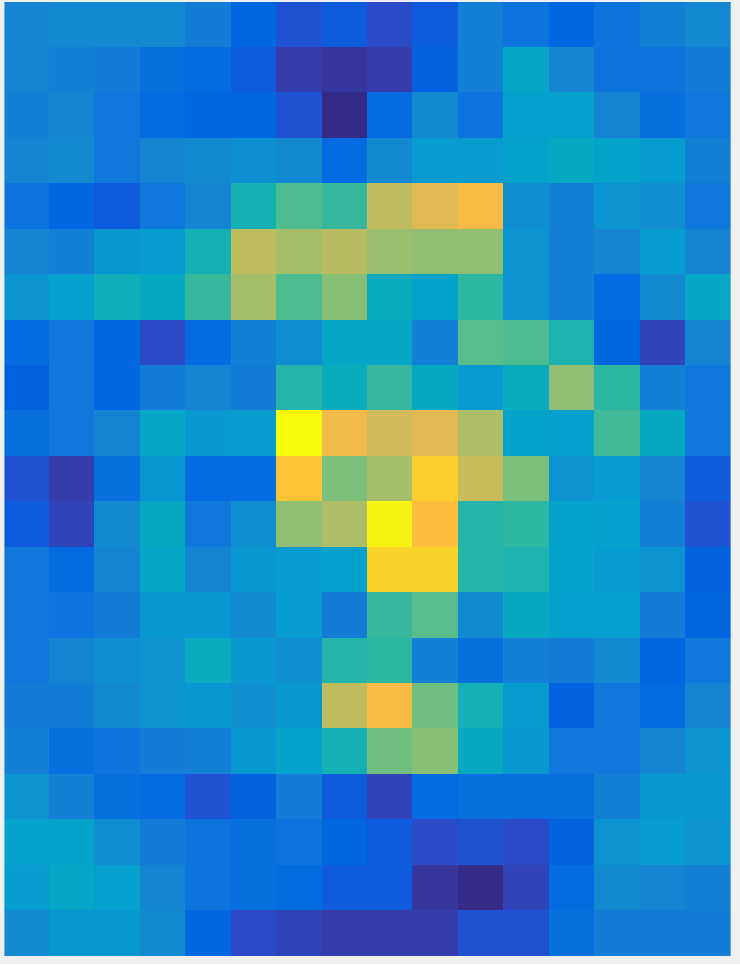} &
		\includegraphics[width=.11\textwidth]{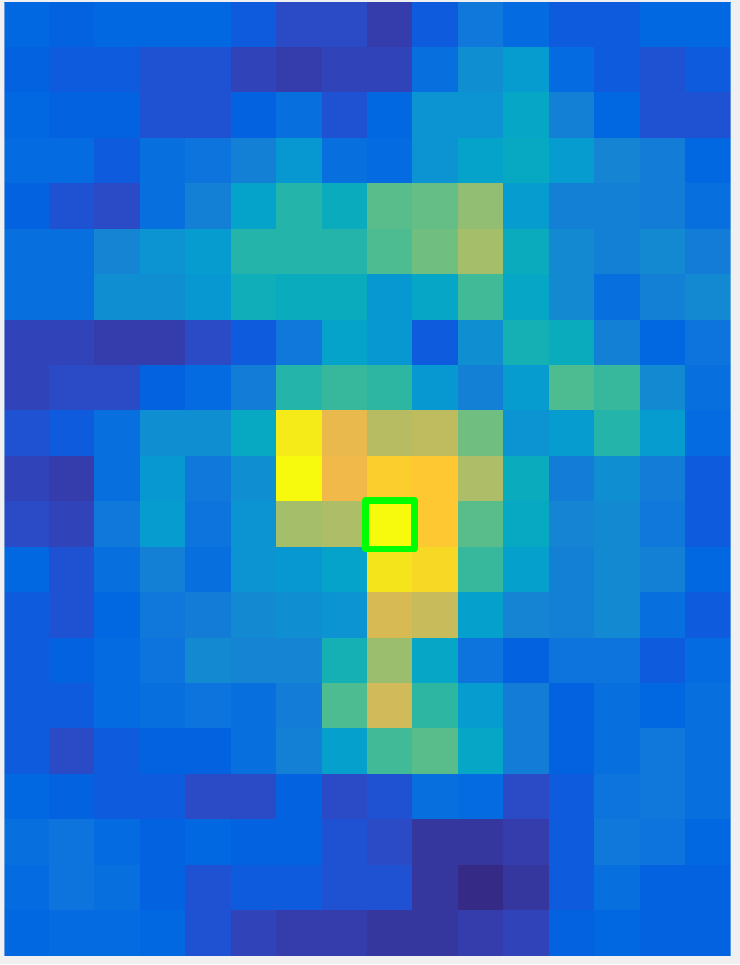} &
		\includegraphics[width=.11\textwidth]{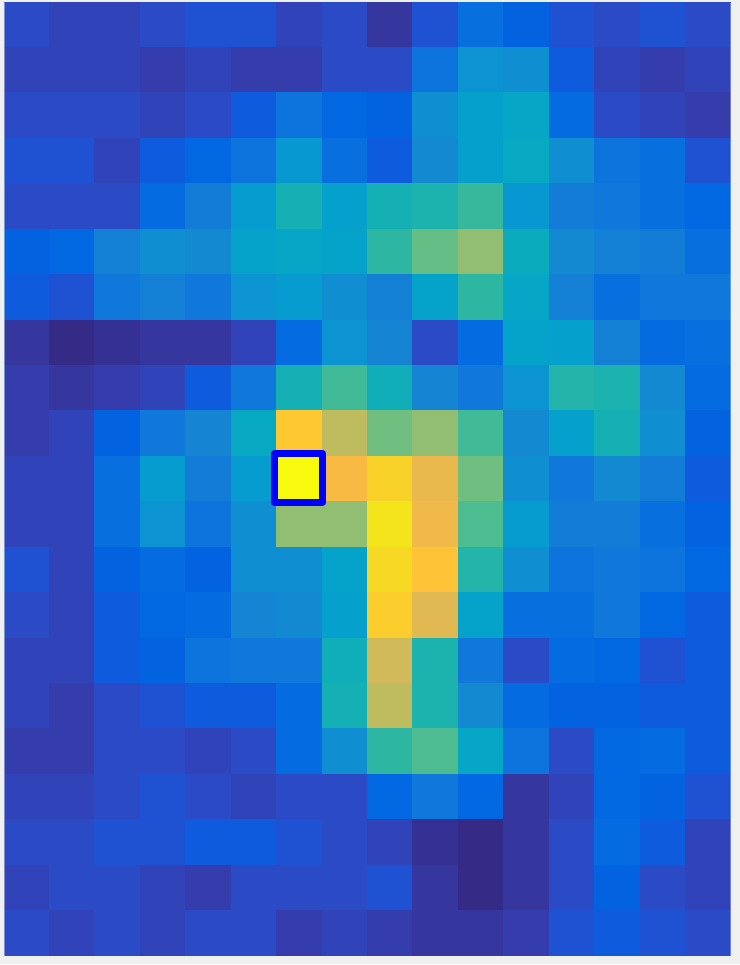} &
		\includegraphics[width=.11\textwidth]{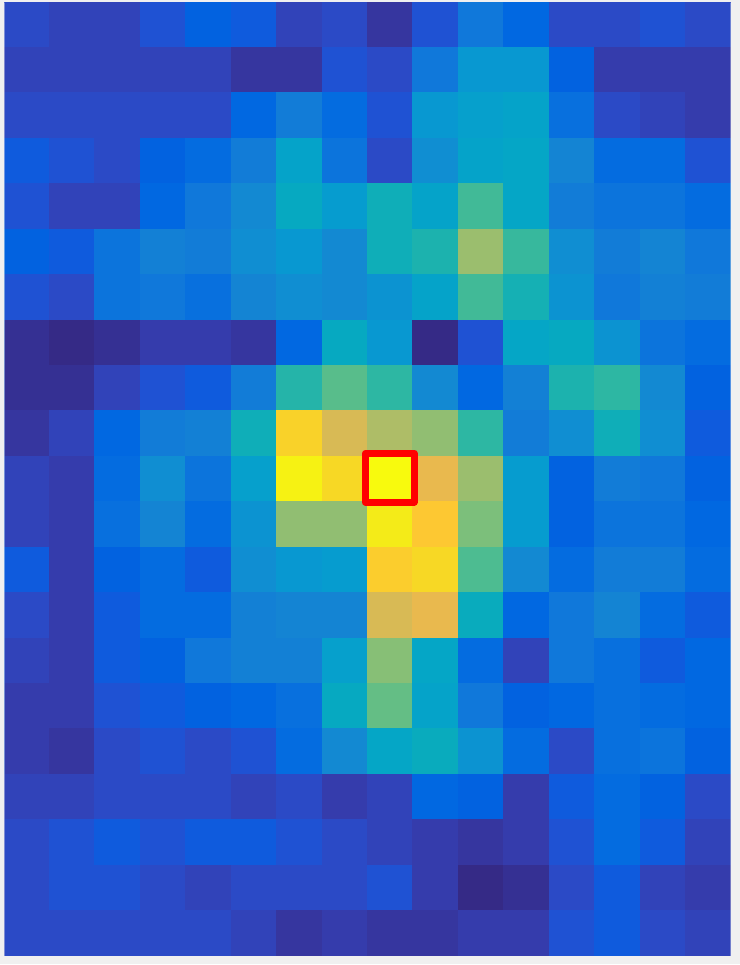} \\
		(d) \textit{conv3-4} & (e) soft mean & (f) hard weight & (g) soft weight \\
	\end{tabular} 
	\caption{\tb{Comparison of different weighting schemes} to locate the target in the search window (yellow). (a) Sample frame in the \textit{KiteSurf}~\cite{DBLP:journals/pami/WuLY15} sequence. (b)-(d) Correlation response maps from single convolutional layers. (e)-(g) Different schemes to weight (b), (c) and (d). The maximum values in (e)-(g) are highlighted by squares, and their tracking results are with the corresponding colors in (a). Note that the soft weight scheme (g) is robust to noisy maximum values in (c) and (d) when compared to the soft mean (e) and hard weight (f) schemes.}
	\label{fig:kitesurf}
	\vspace{-2mm}
\end{figure}

Note that enabling either or both of these two factors results in three different weight schemes:
\vspace{-2mm} 
\begin{itemize}
\item \emph{soft mean}: using only~\eqref{eq:softweight}, $\mu_l=\frac{1}{\max(f_l)}$.
\item \emph{hard weight}: using only~\eqref{eq:weight}, $\mu_l=2^{l-5}$.
\item \emph{soft weight}: using both \eqref{eq:weight} and \eqref{eq:softweight}.
\end{itemize}
\vspace{-2mm} 
Since~\eqref{eq:softweight} amounts to conducting maximum normalization to each response map, we rearrange~\eqref{eq:sum}-\eqref{eq:softweight} for the \emph{soft weight} scheme to locate target objects by
\begin{align}
\label{eq:sum2}
\argmax_{m,n} \sum_l \frac{\mu_l f_l(m,n)}{\max(f_l)}, 
\end{align}
where $\mu_l= 2^{l-5}$. 
In Figure~\ref{fig:kitesurf}, we compare three different weight schemes to locate the target in the 38-th frame of the \textit{KiteSurf}~\cite{DBLP:journals/pami/WuLY15} sequence. 
Compared to the soft mean (Figure~\ref{fig:kitesurf}(e)) and hard weight (Figure~\ref{fig:kitesurf}(f)) schemes, the soft weight scheme (Figure~\ref{fig:kitesurf}(g)) is robust to noisy maximum values in the \textit{conv4-4} and \textit{conv3-4} layers. 
In addition, Figure~\ref{fig:weight} shows the weighted maximum response values as well as the center location 
errors in the entire \textit{KiteSurf}~\cite{DBLP:journals/pami/WuLY15} sequence. 
The \emph{soft weight} scheme helps to track the target well over the entire sequence despite the presence of heavy occlusion and abrupt motion in the 38-th frame (see Figure~\ref{fig:kitesurf}(a)). 
Note that other alternative approaches including response maps of single convolutional layers cannot reliably track the object in the sequence. 
Our preliminary results in~\cite{DBLP:conf/iccv/MaHYY15} use the \emph{hard weight} scheme. 
Ablation studies (see Section~\ref{sec:component}) show the effectiveness of the soft weight scheme that achieves 
the accuracy gain of around +1\% on the tracking benchmark dataset with 100 sequences.

\begin{figure}[t]
\centering
\setlength{\tabcolsep}{1pt}
\begin{tabular}{cc}
\includegraphics[width=.22\textwidth]{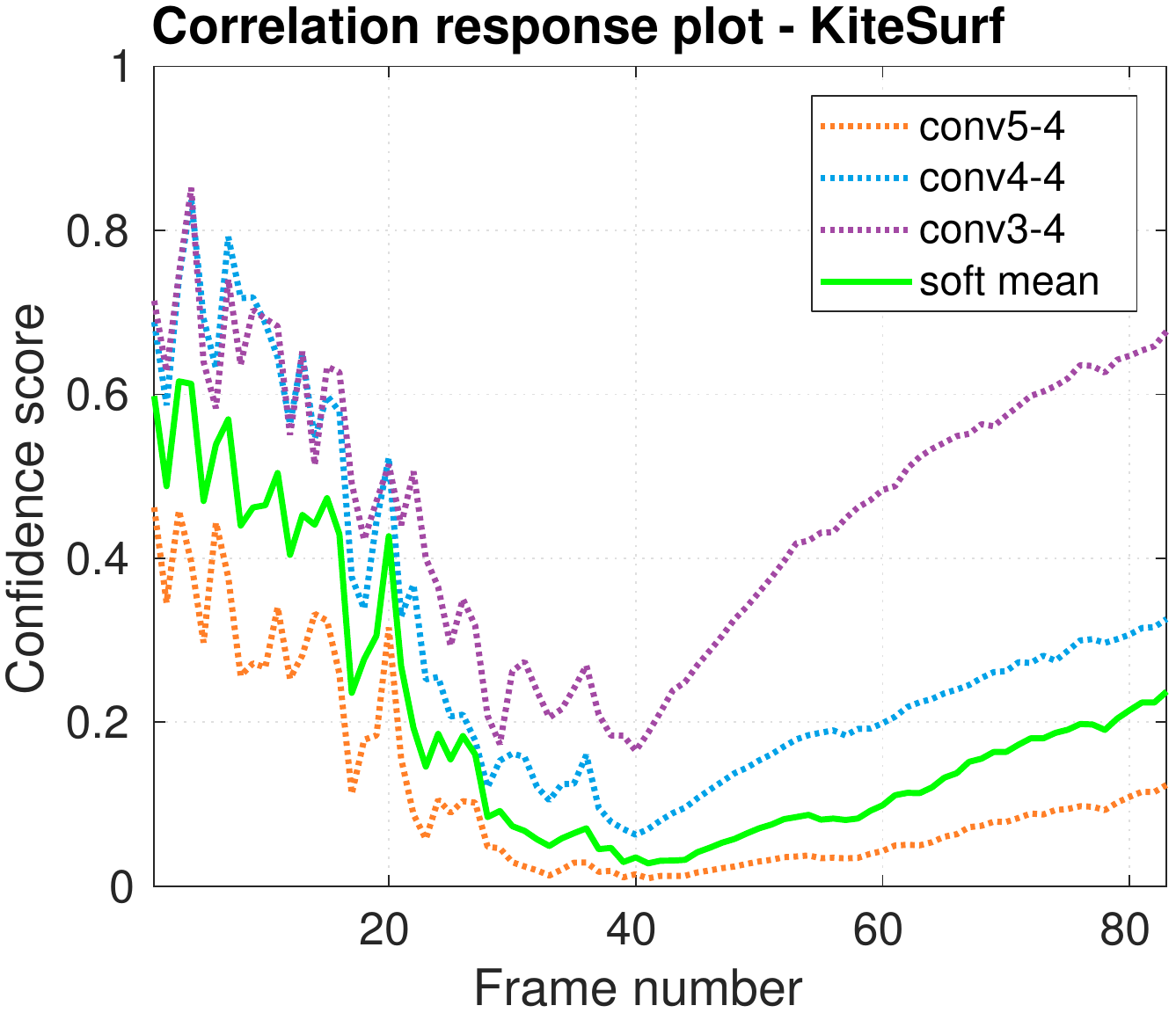} &
\includegraphics[width=.22\textwidth]{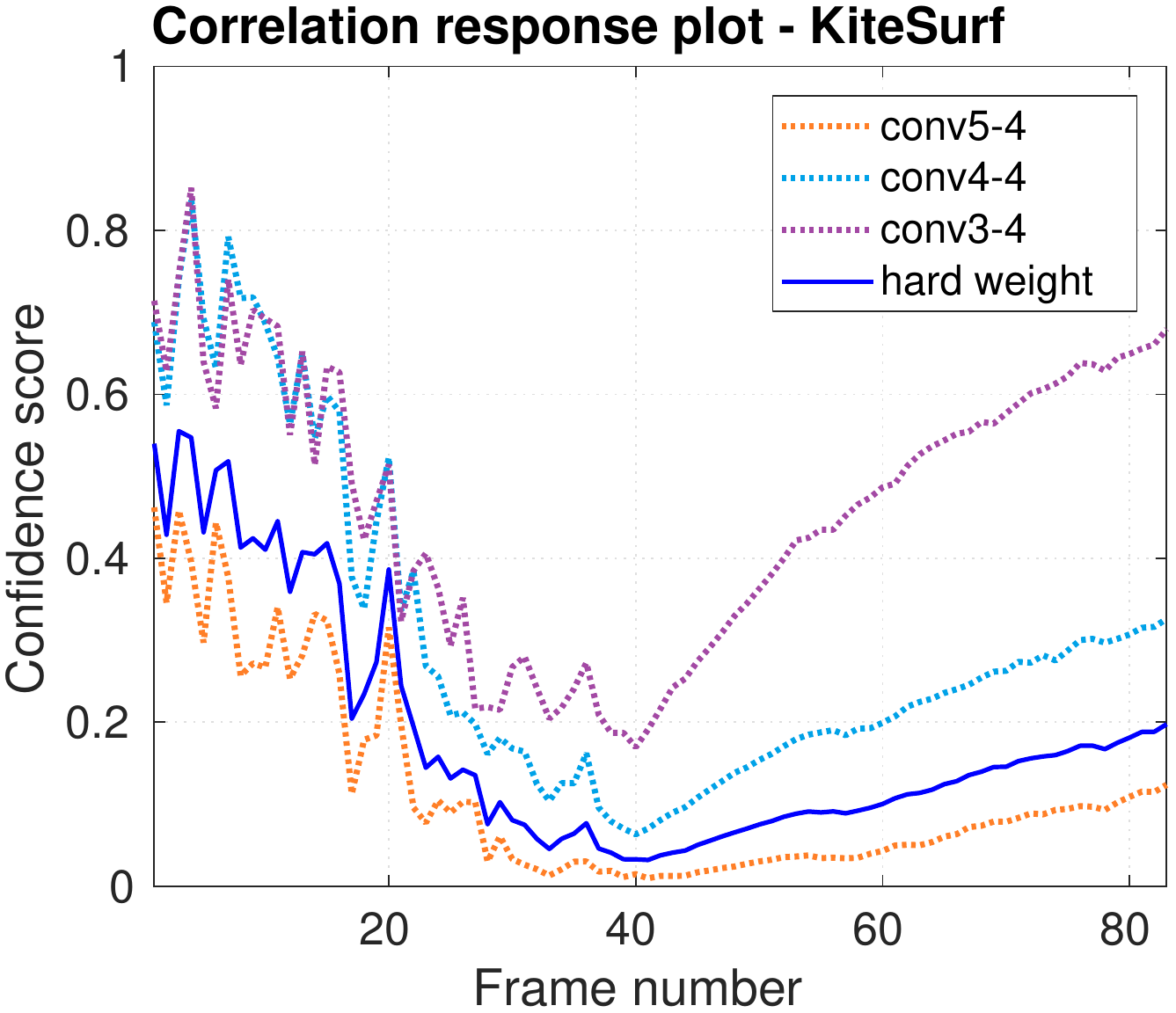} \\
(a) Soft mean & (b) Hard weight \\
\includegraphics[width=.22\textwidth]{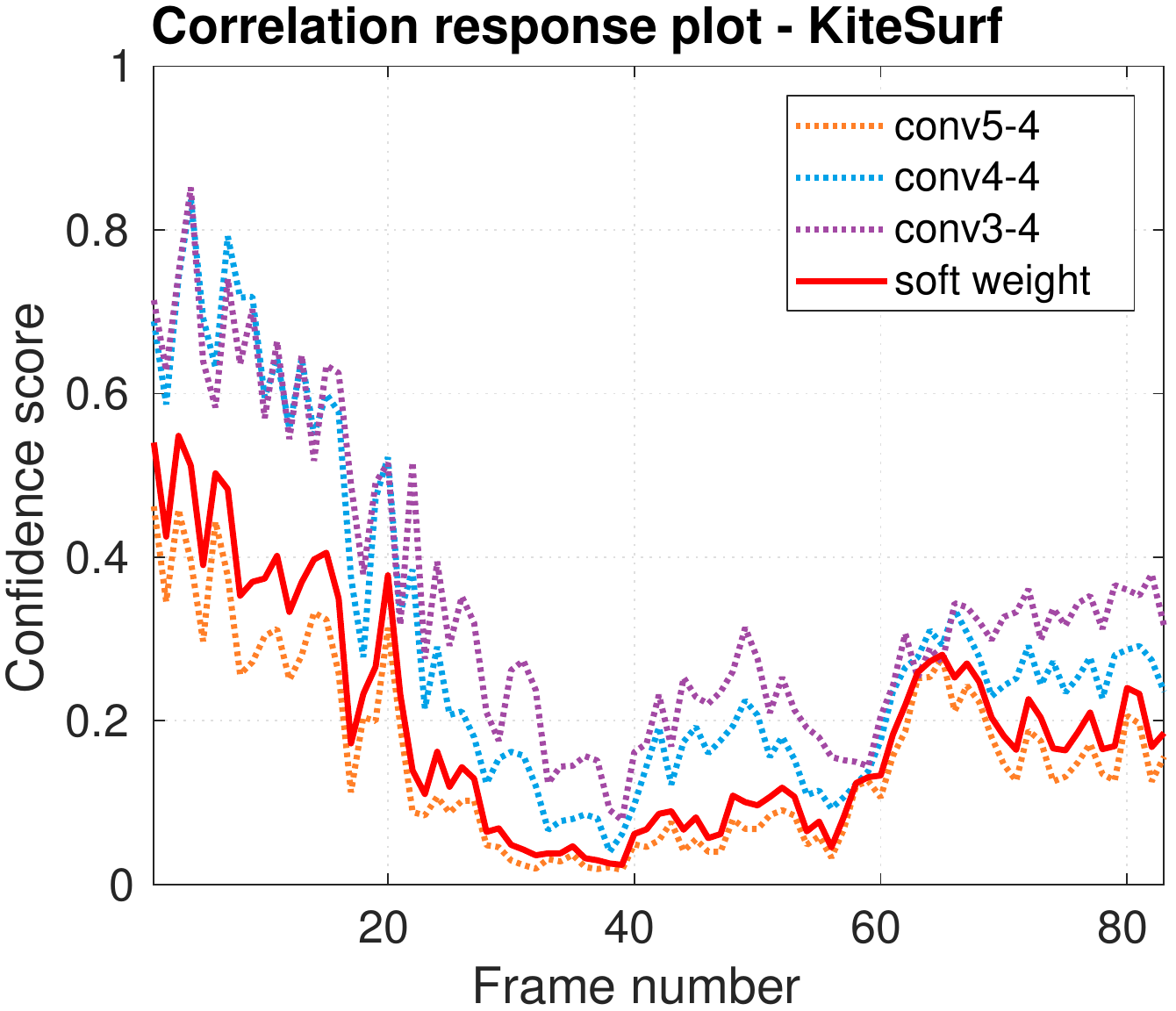} & 
\includegraphics[width=.22\textwidth]{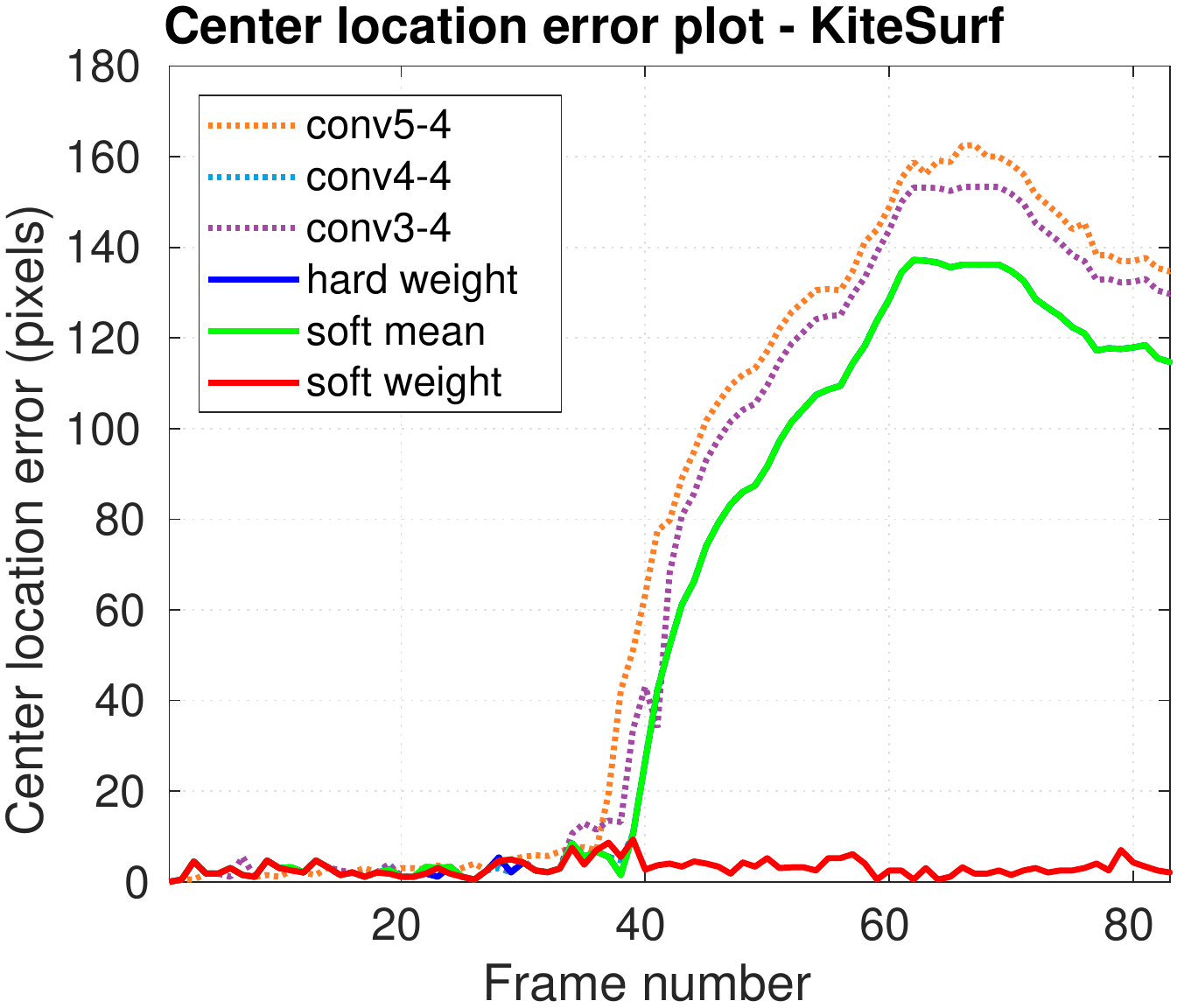} \\
(c) Soft weight & (d) Center location error\\ 
\end{tabular}
%\caption{\tb{Weighted maximum response and center location error plot} on the \textit{KiteSurf} sequence~\cite{DBLP:journals/pami/WuLY15}. Note that the soft weight scheme sets the weights to be inversely proportional to the maximum values of response maps. The weighted values are more consistent than the soft mean and hard weight schemes. The center location error plot shows that the soft weight scheme helps track the target over the entire sequence, despite the presence of heavy occlusion and abrupt motion in the 38-th frame (see Figure~\ref{fig:kitesurf}), while other alternative approaches do not produce satisfactory results.}
\caption{\tb{Frame-by-frame maximum response values and center location error plot} on the \textit{KiteSurf}~\cite{DBLP:journals/pami/WuLY15} sequence. \textcolor{black}{In (a)-(c), dash lines indicate maximum values of single correlation response maps on different CNN layers. Solid lines show maximum values of the combined response map over all CNN layers.} Note that the soft weight scheme sets the weights to be inversely proportional to the maximum values of response maps. The weighted values are more consistent than the soft mean and hard weight schemes. The center location error plot shows that the soft weight scheme helps track the target over the entire sequence, despite the presence of heavy occlusion and abrupt motion in the 38-th frame (see Figure~\ref{fig:kitesurf}), while other alternative approaches do not produce satisfactory results.}
\label{fig:weight}
\vspace{-4mm}
\end{figure}

\vspace{-2mm}
\subsection{Region Proposals}
\label{sec:proposals}
To better estimate scale change and re-detect target objects from tracking failures, we use the EdgeBox~\cite{DBLP:conf/eccv/ZitnickD14} method to generate two types of region proposals: (1) scale proposals $\mathbf{B}_s$ with small step size and tightly around the estimated target location, and (2) detection proposals $\mathbf{B}_d$ with large step size and across the whole image.
We denote each proposal $\mathbf{b}$ in either $\mathbf{B}_s$ or $\mathbf{B}_d$ as a candidate bounding box $(x,y,w,h)$, where $(x,y)$ is center axis and $(w,h)$ is width and height.
In order to compute the confidence score of each proposal $\mathbf{b}$, we learn another correlation filter with a conservative learning rate (Section~\ref{sec:cf}) to maintain a long-term memory of target appearance.
Specifically, we learn this filter over the histogram of both gradients and intensities as in \cite{ma2018adaptive} to encode more spatial details 
for differentiating minor scale changes.
Given a proposal $\mathbf{b}$, we denote the maximum filter response of the long-term memory correlation filter by $g(\mathbf{b})$.

\vspace{-2mm}
\subsubsection {Target Recovery}

Given the estimated target position $(x_t,y_t)$ in the $t$-th frame using~\eqref{eq:sum2}, 
we crop a patch $\mathbf{z}$ centered at this position with the same width and height in the previous frame. 
We set a threshold $T_0$ to determine if tracking failures occur.
When the confidence score $g(\mathbf{z})$ is below $T_0$, we flag the tracker as losing the target and proceed with target re-detection.

We generate a set of region proposals $\mathbf{B}_d$ with a large step size across the whole frame for recovering target objects. 
Rather than simply selecting the candidate bounding box with maximum confidence score as the recovered result, we also take the motion constraint into consideration for cases with drastic change between two consecutive frames.
We compute the center location distance $\mathcal{D}$ between each proposal $\mathbf{b}_t^i$ and the bounding box $\mathbf{b}_{t-1}$ in the previous frame as: 
\begin{equation}
\mathcal{D}(\mathbf{b}_t^i,\mathbf{b}_{t-1})= \exp\bigl(-\frac{1}{2\sigma^2} \|(x_t^i, y_t^i)-(x_{t-1},y_{t-1})\|^2\bigr), 
\label{equ:dis}
\end{equation}
where $\sigma$ is the diagonal length of the initial target size.
We select the optimal proposal as the re-detection result by minimizing the following problem:
\begin{align}
\label{equ:searchb}
\argmin_i ~ & ~ g(\mathbf{b}_t^i)+ \alpha \mathcal{D}(\mathbf{b}_t^i, \mathbf{b}_{t-1}), \\ 
\text{s.t.} ~~ & ~ g(\mathbf{b}_t^i)>1.5 T_0. \nonumber 
\end{align}
In~\eqref{equ:searchb}, the weight factor $\alpha$ aims to strike a balance between patch confidence and motion smoothness. 
%
%Since adopting the re-detection result will re-initiate the entire tracking process, we use a conservative scheme by increasing the threshold $T_0$ by an enlarging factor of 1.5.
% 

\vspace{-2mm}
\subsubsection{Scale Estimation}
By using a small step size and a small non-maximum suppression (NMS) threshold, we
use the EdgeBox~\cite{DBLP:conf/eccv/ZitnickD14} method to generate 
multi-scale region proposals $\mathbf{B}_s$ for scale estimation. 
We use a similar proposal rejection technique as in~\cite{bmvc15huang} to filter the proposals whose intersection over union (IoU) with the target bounding box (centered at the estimated position with the scale of the last frame) is smaller than 0.6 or larger than 0.9. 
We then resize the remaining proposals to a fixed size for computing confidence scores. 
If the maximum confidence score $\{g(\mathbf{b})|\mathbf{b}\in\mathbf{B}_s\}$ is larger than $g(\mathbf{z})$, we update the width $w_t$ and height $h_t$ of the target using a moving average:
\begin{equation}
(w_t, h_t) = \beta(w_t^*, h_t^*) + (1-\beta) ( w_{t-1} , h_{t-1} ), 
\label{equ:scale}
\end{equation}
where $w_t^*$ and $h_t^*$ denote the width and height of the proposal with maximum confidence score. 
The weight factor $\beta$ makes the target size estimates change smoothly.

\vspace{-2mm}
\subsection{Model Update}
\label{sec:update}
An optimal filter on $l$-th layer can be updated by minimizing the output error over \emph{all} tracked results 
at time $t$ as described in~\cite{DBLP:conf/cvpr/BoddetiKK13}.
However, this involves solving a $D\times D$ linear system of equations per location at $(x,y)$, which is computationally expensive as the channel number is usually large with the CNN features (e.g., $D=512$ in the \textit{conv5-4} and \textit{conv4-4} layers in the VGGNet).
To obtain a better approximation, we update the correlation filter $\mathbf{W}^d$ in~\eqref{equ:w} using a moving average:
\begin{subequations}
\label{equ:upfilter}
\begin{align}
\mathbf{A}_t^d&=(1-\eta)\mathbf{A}_{t-1}^d+\eta \mathbf{Y}\odot\bar{\mathbf{X}}_t^d; \\
\mathbf{B}_t^d&=(1-\eta)\mathbf{B}_{t-1}^d+\eta\sum_{i=1}^{D}\mathbf{X}_t^i\odot\bar{\mathbf{X}}_t^i; \\
\mathbf{W}_t^d&=\frac{\mathbf{A}_t^d}{\mathbf{B}_t^d+\lambda},
\end{align}
\end{subequations}
where $t$ is the frame index and $\eta$ is a learning rate. 

On the other hand, we conservatively update the long-term correlation filter $g(\cdot)$ only if the tracked result $\mathbf{z}$ is highly confident, i.e., $g(\mathbf{z})>T_0$. 
As such, the learned filter 
is equipped with a long-term memory of target appearance and 
robust to noisy updates that often cause rapid model degradation. 

\vspace{-2mm}
\section{Implementation Details}

\begin{algorithm}[t]
\SetKwData{Left}{left}\SetKwData{This}{this}\SetKwData{Up}{up}
\SetKwInOut{Input}{Input}\SetKwInOut{Output}{Output}
\Input{Initial target position $(x_{t-1},y_{t-1}, w_{t-1}, h_{t-1})$, the hierarchical correlation filters $\{\mathbf{W}^{l}_{t-1}|l=3,4,5\}$, and the classifier $g_{t-1}(\cdot)$ for confidence computation. }
\Output{Estimated object position $(x_t,y_t, w_t, h_t)$, 
$\{\mathbf{W}^{l}_t\}$, and $g_{t}(\cdot)$.}
\BlankLine
\Repeat{End of video sequences}{
Crop out the search window in frame $t$ centered at ($x_{t-1},y_{t-1}$) and 
extract covolutional features with spatial interpolation using~\eqref{equ:upsample}\; 
\lForEach{layer $l$}{computing confidence score $f_l$ using $\mathbf{W}^l_{t-1}$ and~\eqref{equ:res}}
Estimate the new position ($x_t, y_t$) on response map set $\{f_l\}$ using~\eqref{eq:sum2}\;
Crop out new patch $\mathbf{z}$ centered at $(x_t, y_t)$ and compute the tracking confidence $g(\mathbf{z})$\;
\lIf{$g(\mathbf{z})<T_0$}{perform target re-detection to update ($x_t, y_t$) using~\eqref{equ:searchb}}
Estimate the new scale $(w_t,h_t)$ using~\eqref{equ:scale} around ($x_t, y_t$) and compute the new confidence score $g(\mathbf{z})$\; 
\lForEach{layer $l$}{updating correlation filters $\{\mathbf{W}^{l}_t\}$ using~\eqref{equ:upfilter} with convolutioanl features}
\lIf{$g(\mathbf{z})>T_0$}{update classifier $g$}
}
\caption{Proposed tracking algorithm.}
\label{alg:pipeline}
\end{algorithm}

We present the main steps of the proposed tracking algorithm in Algorithm~\ref{alg:pipeline} and 
implementation details as follows. 
We adopt the VGGNet-19~\cite{DBLP:journals/corr/SimonyanZ14a} trained on the ImageNet~\cite{DBLP:conf/cvpr/DengDSLL009} dataset as our feature extractor.
We first remove the fully-connected layers and use the outputs of the \textit{conv3-4}, \textit{conv4-4}, and \textit{conv5-4} convolutional layers as deep features (see Figure~\ref{fig:vis}).
To retain a larger spatial resolution on each convolutional layer, we do not use the outputs of the pooling layers.
Given an image frame with search window of size $M\times N$ (e.g.,
1.8 times of the target size), we set a fixed spatial size of $\frac{M}{4}\times\frac{N}{4}$ to resize the feature channels from each convolutional layer.
For translation estimation, we keep the parameters on each convolutional layer the same for training correlation filters.
We set the regularization parameter $\lambda$ in~\eqref{equ:filter} to $10^{-4}$ and the kernel width for generating the Gaussian function labels to 0.1. 
We set the model update learning rate $\eta$ in~\eqref{equ:upfilter} to 0.01.
To remove the boundary discontinuities, 
we weight the extracted feature channels of each convolutional layer 
by a cosine box filter~\cite{DBLP:conf/cvpr/BolmeBDL10}.
We set the weight parameters $\mu_1$, $\mu_2$ and $\mu_3$ in~\eqref{eq:sum2} to 1, 0.5 and 0.25, respectively. 
We fine-tune the step size and the non-maximum suppress threshold of the EdgeBox method to generate two types of region proposals. 
For generating proposals to re-detect target objects, we use a large step size 0.85 and a larger NMS threshold 0.8.
For generating proposals to estimate scales, we set the step size to 0.75 and the NMS threshold to 0.6.
The weight parameter $\alpha$ in~\eqref{equ:searchb} is set to 0.1. 
For scale estimation, the parameter $\beta$ in~\eqref{equ:scale} is set to 0.6. 
%
%
%The current object tracking benchmarks, such as the OTB \cite{DBLP:conf/cvpr/WuLY13,DBLP:journals/pami/WuLY15} and the VOT \cite{Hadfield14d,DBLP:conf/iccvw/KristanMLFCFVHN15}, do not provide separate validation and test sets.
%
%To address the lack of validation set in the OTB \cite{DBLP:conf/cvpr/WuLY13,DBLP:journals/pami/WuLY15} and the VOT \cite{Hadfield14d,DBLP:conf/iccvw/KristanMLFCFVHN15} datasets, w
%
\textcolor{black}{The tracking failure threshold $T_0$ is set to 0.2. More details regarding parameter settings on a \emph{separate} set of video sequences introduced by the MEEM method ({\small \url{http://cs-people.bu.edu/jmzhang/MEEM/MEEM.html})} are provided in the supplementary document.  
}
%
%the discussion of hyper-parameter selection (how you use a validation set to tune the threshold) should be included in Sec 5 and refer to the supplementary material for more details

\vspace{-2mm}
\section{Experiments}

We evaluate the proposed algorithm on four benchmark datasets: OTB2013~\cite{DBLP:conf/cvpr/WuLY13}, OTB2015~\cite{DBLP:journals/pami/WuLY15}, VOT2014~\cite{Hadfield14d}, and VOT2015~\cite{DBLP:conf/iccvw/KristanMLFCFVHN15}. 
We use the benchmark protocols and the same parameters for all the sequences as well as all the sensitivity analysis.
For completeness, we also report the results in terms of distance precision rate, overlap ratio, center location error, and tracking speed in comparison with the state-of-the-art trackers.
We implement the proposed tracking algorithm in MATLAB on an Intel I7-4770 3.40 GHz CPU with 32 GB RAM and the MatConvNet toolbox~\cite{DBLP:journals/corr/VedaldiL14}, 
where the computation of forward propagation for CNN feature extraction is carried out on a GeForce GTX Titan GPU. 
The source code, as well as additional experimental results, are publicly available at our project page~\url{https://sites.google.com/site/chaoma99/hcft-tracking}.

\begin{table*}
\centering
\caption{\tb{Comparisons with state-of-the-art trackers on OTB2013 (\uppercase\expandafter{\romannumeral 1})~\cite{DBLP:conf/cvpr/WuLY13} 
and OTB2015 (\uppercase\expandafter{\romannumeral 2})~\cite{DBLP:journals/pami/WuLY15} benchmark sequences.} 
Our approach performs favorably against existing methods in distance precision rate at a threshold of $20$ pixels, overlap success rate at an overlap threshold of $0.5$ and center location error. 
The first and second best values are highlighted by bold and underline.}
\label{tb:compare}
\vspace{-2mm}
\scriptsize
\setlength{\tabcolsep}{.9em}
\begin{tabular}{r*{14}{c}} \hline
& & HCFT* & HCFT & FCNT & SRDCF & MUSTer & LCT & MEEM & TGPR & DSST & KCF & Struck & DLT \\
& & &~\cite{DBLP:conf/iccv/MaHYY15} &~\cite{DBLP:conf/iccv/WangOWL15} &~\cite{DBLP:conf/iccv/DanelljanHKF15} &~\cite{Hong_2015_CVPR} &~\cite{DBLP:conf/cvpr/MaYZY15} &~\cite{DBLP:conf/eccv/ZhangMS14} &~\cite{DBLP:conf/eccv/GaoLHX14} &~\cite{DBLP:conf/bmvc/DanelljanKFW14} &~\cite{DBLP:journals/pami/HenriquesC0B15} &~\cite{DBLP:conf/iccv/HareST11} &~\cite{DBLP:conf/nips/WangY13} \\\hline
\multirow{2}{*}{DP rate (\%)} & \uppercase\expandafter{\romannumeral 1} & \textbf{92.3} & {\ul 89.1} & 85.7 & 82.3 & 86.5 & 84.8 & 83.0 & 70.5 & 73.9 & 74.1 & 65.6 & 54.8 \\
& \uppercase\expandafter{\romannumeral 2} & \textbf{87.0} & {\ul 83.7} & 77.9 & 77.6 & 77.4 & 76.2 & 78.1 & 64.3 & 69.5 & 69.2 & 63.5 & 52.6 \\\hline
\multirow{2}{*}{OS rate (\%)} & \uppercase\expandafter{\romannumeral 1} & {\ul 79.3} & 74.0 & 75.7 & 77.6 & 78.4 & \textbf{81.3} & 69.6 & 62.8 & 59.3 & 62.2 & 55.9 & 47.8 \\
& \uppercase\expandafter{\romannumeral 2} & \textbf{72.0} & 65.5 & 66.6 & {\ul 71.5} & 68.3 & 70.1 & 62.2 & 53.5 & 53.7 & 54.8 & 51.6 & 43.0 \\\hline
\multirow{2}{*}{CLE (pixel)} & \uppercase\expandafter{\romannumeral 1} & \textbf{9.43} & {\ul 15.7} & 17.4 & 35.8 & 17.3 & 26.9 & 20.9 & 51.3 & 40.7 & 35.5 & 50.6 & 65.2 \\
& \uppercase\expandafter{\romannumeral 2} & \textbf{16.7} & {\ul 22.8} & 23.6 & 39.2 & 31.5 & 67.1 & 27.7 & 55.5 & 47.7 & 45.0 & 47.1 & 66.5 \\\hline
\multirow{2}{*}{Speed (FPS)} & \uppercase\expandafter{\romannumeral 1} & 6.63 & 11.0 & 2.31 & 6.09 & 9.32 & 21.6 & 20.8 & 0.66 & {\ul 43.0} & \textbf{246} & 10.0 & 8.59 \\
& \uppercase\expandafter{\romannumeral 2} & 6.70 & 10.4 & 2.48 & 5.69 & 8.46 & 20.7 & 20.8 & 0.64 & {\ul 40.9} & \textbf{243} & 9.84 & 8.43 \\\hline
\end{tabular}
\vspace{-2mm}
\end{table*}

\vspace{-2mm}
\subsection{OTB datasets}

The object tracking benchmark (OTB) dataset contains two sets:
(1) OTB2013~\cite{DBLP:conf/cvpr/WuLY13} with 50 sequences and (2) OTB2015~\cite{DBLP:journals/pami/WuLY15} with 100 sequences. 
We mainly report the results on the OTB2015 dataset since the other one is a subset.

We compare the proposed algorithm with 10 state-of-the-art trackers which can be categorized as follows:
\begin{itemize}
\item Deep learning trackers including FCNT~\cite{DBLP:conf/iccv/WangOWL15} and DLT~\cite{DBLP:conf/nips/WangY13} as well as the HCFT~\cite{DBLP:conf/iccv/MaHYY15} method which is our earlier work;
\item Correlation filter trackers including SRDCF~\cite{DBLP:conf/iccv/DanelljanHKF15}, MUSTer~\cite{Hong_2015_CVPR}, LCT~\cite{DBLP:conf/cvpr/MaYZY15}, DSST~\cite{DBLP:conf/bmvc/DanelljanKFW14} and KCF~\cite{DBLP:journals/pami/HenriquesC0B15};
\item Representative tracking algorithms using single or multiple online classifiers, including the MEEM~\cite{DBLP:conf/eccv/ZhangMS14}, TGPR~\cite{DBLP:conf/eccv/GaoLHX14}, and Struck~\cite{DBLP:conf/iccv/HareST11} methods.
\end{itemize}
We use the benchmark protocol~\cite{DBLP:conf/cvpr/WuLY13} and 
two metrics to evaluate the tracking performance.
%\begin{itemize}
%\item 
The overlap success rate is the percentage of frames where the overlap ratio between predicted bounding box ($B_1$) and ground truth bounding box ($B_0$) is larger than a given threshold $T$, i.e., $\frac{B_1\cap B_0}{B_1\cup B_0}>T$.
%\item 
The distance precision rate is the percentage of frames where the estimated center location error is smaller than a given distance threshold, e.g., 20 pixels in this work. 
%\end{itemize} 

\subsubsection{Overall Performance}

\begin{figure}[t]
\centering
\setlength{\tabcolsep}{.25mm}
\begin{tabular}{cc}
\includegraphics[width=.22\textwidth]{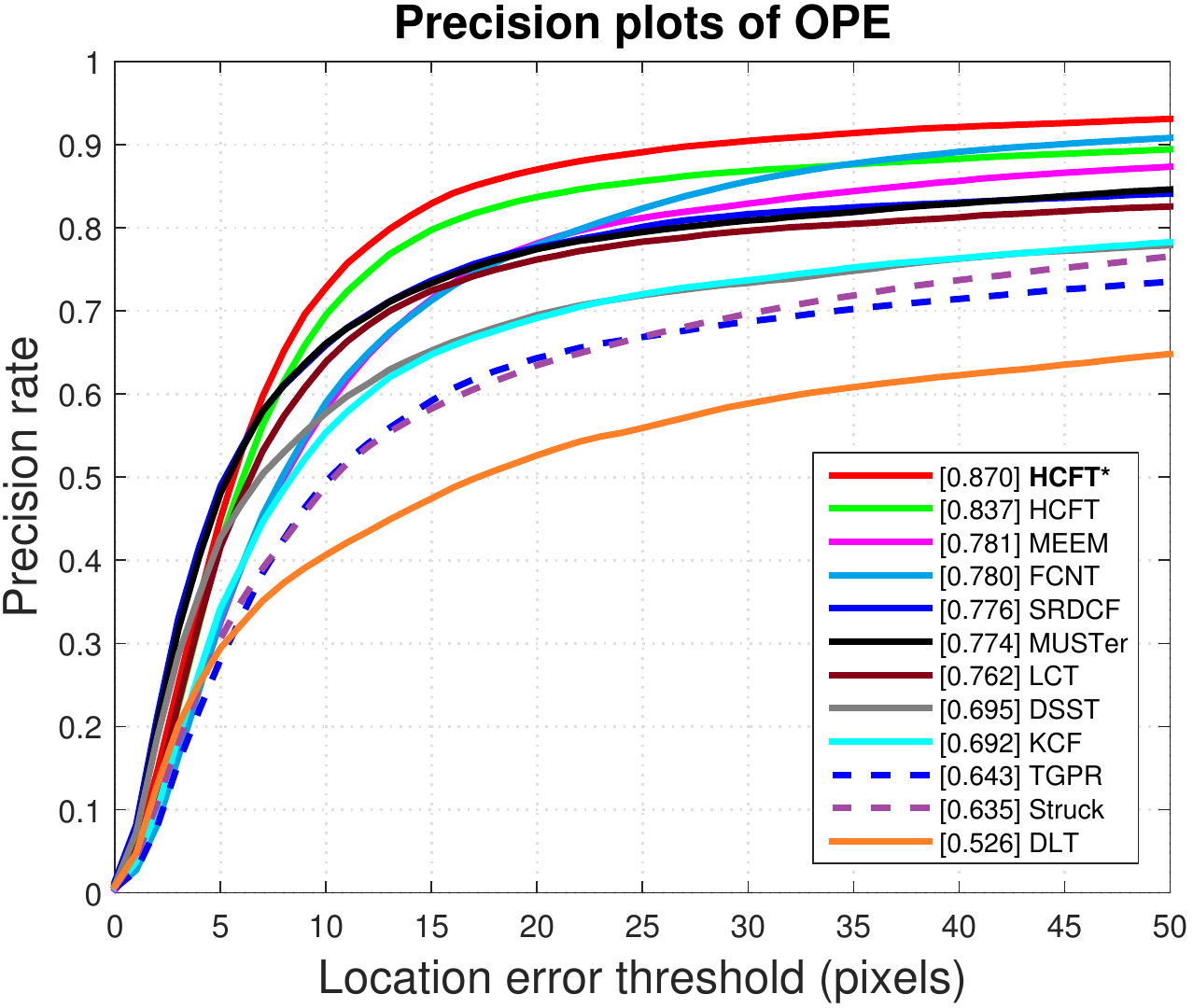} &
\includegraphics[width=.22\textwidth]{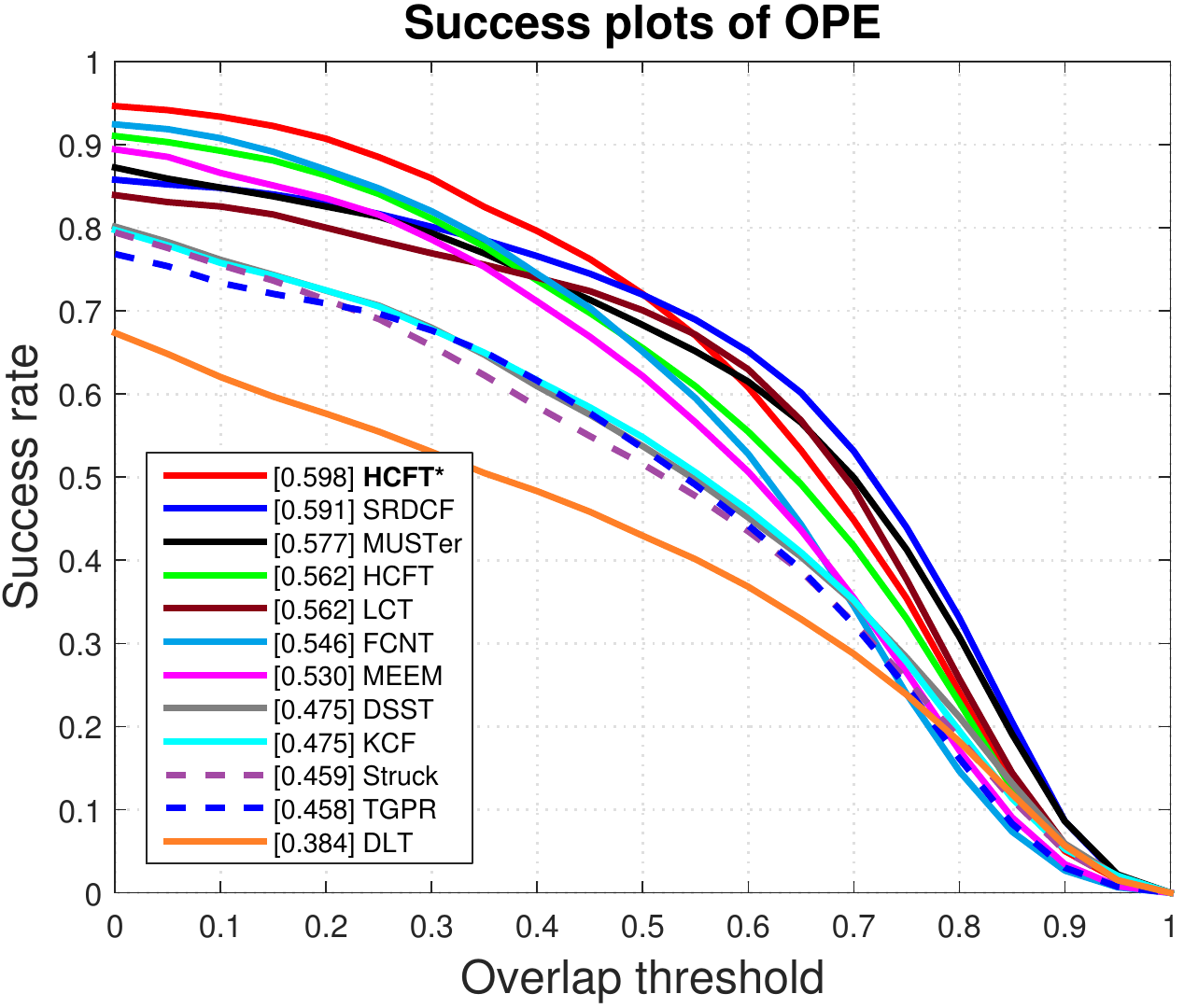} \\
\includegraphics[width=.22\textwidth]{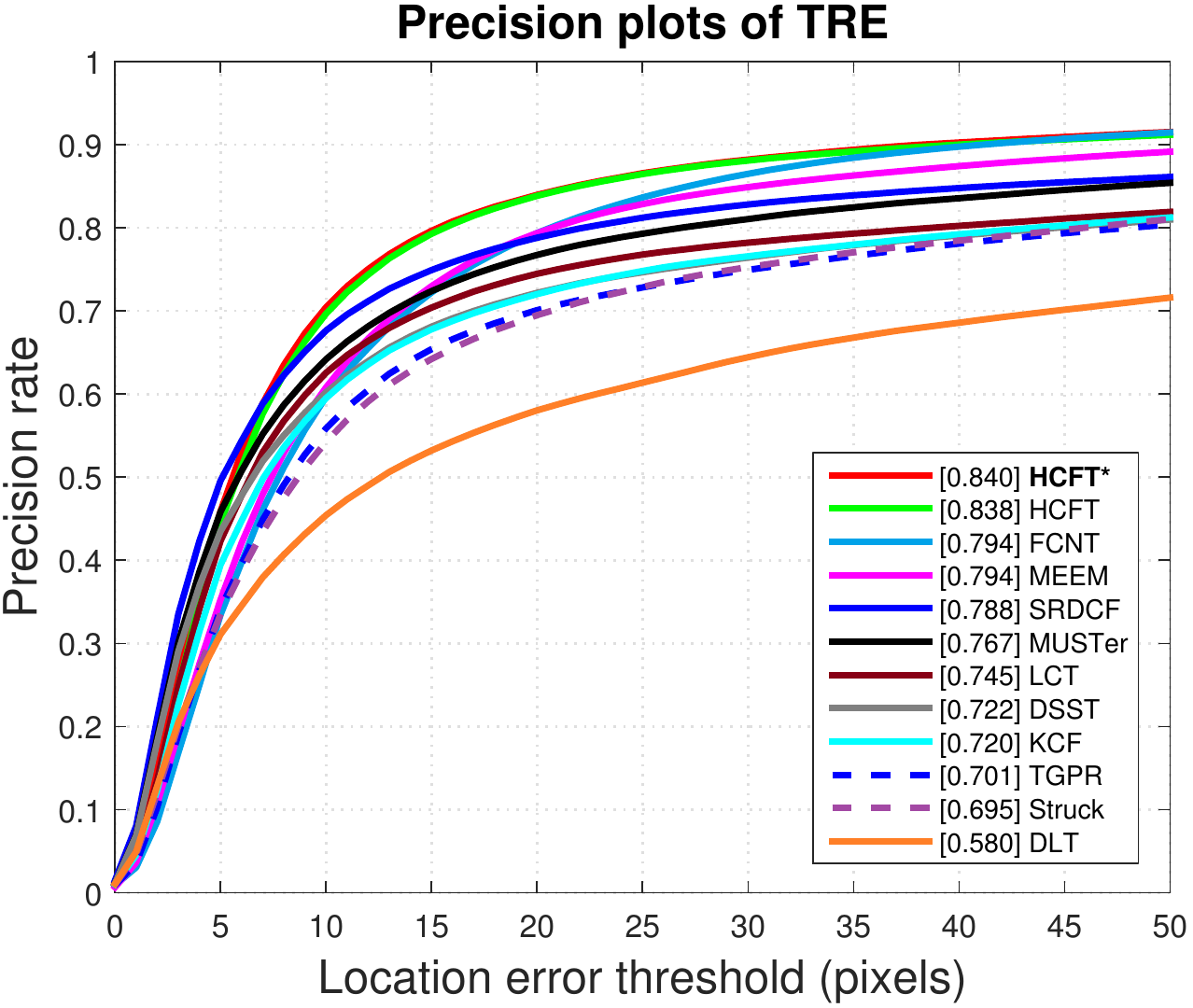} &
\includegraphics[width=.22\textwidth]{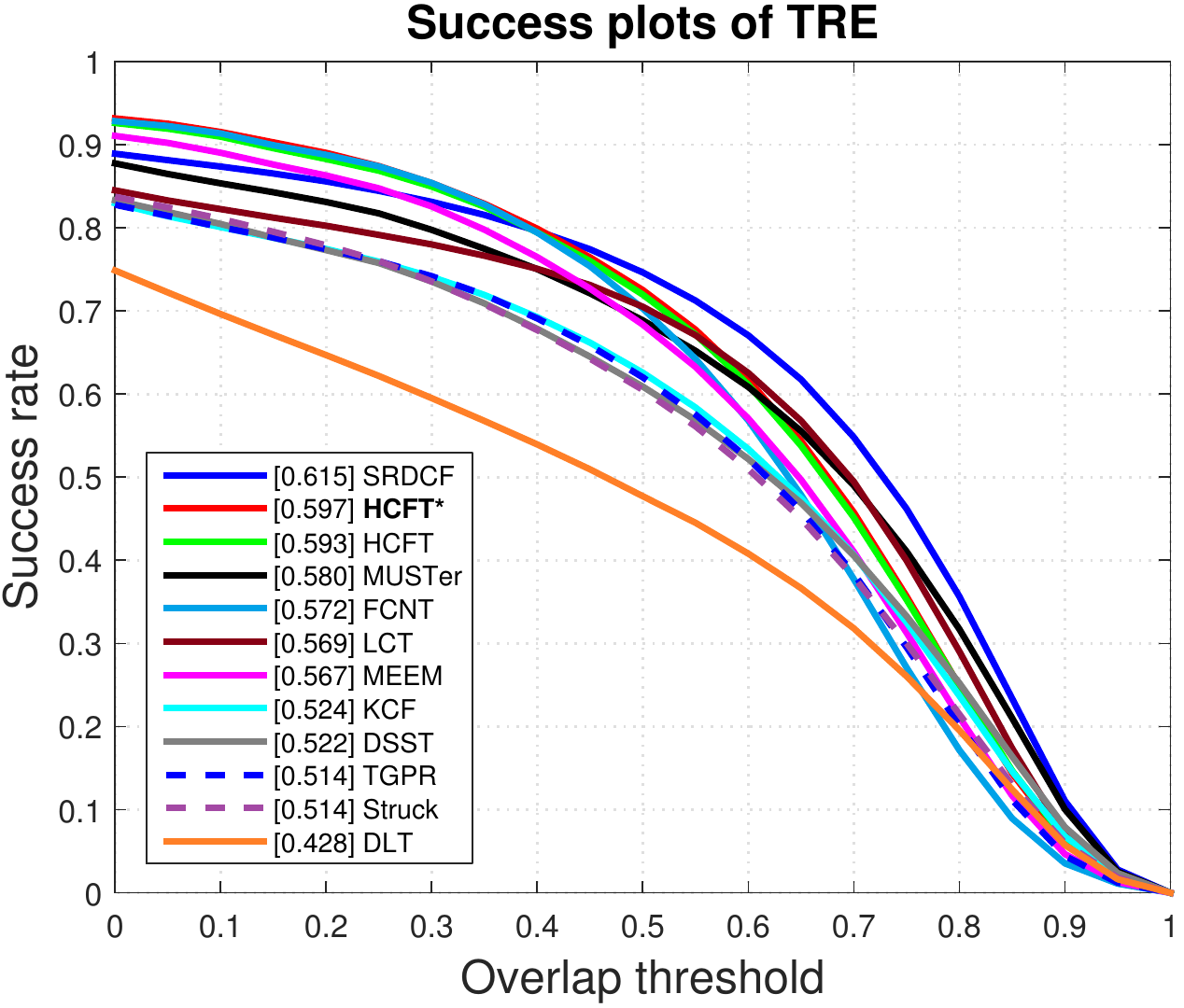} \\
\includegraphics[width=.22\textwidth]{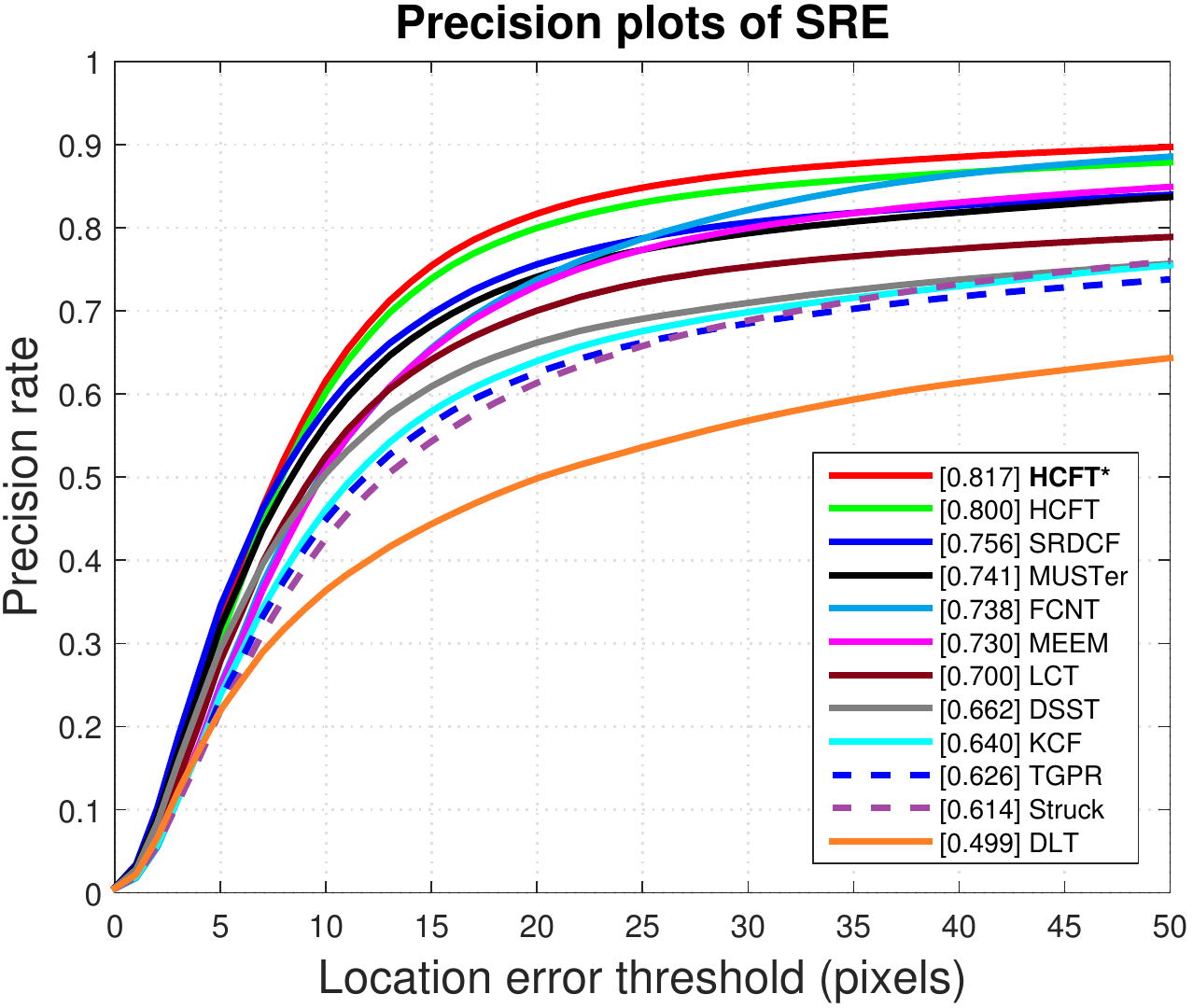} &
\includegraphics[width=.22\textwidth]{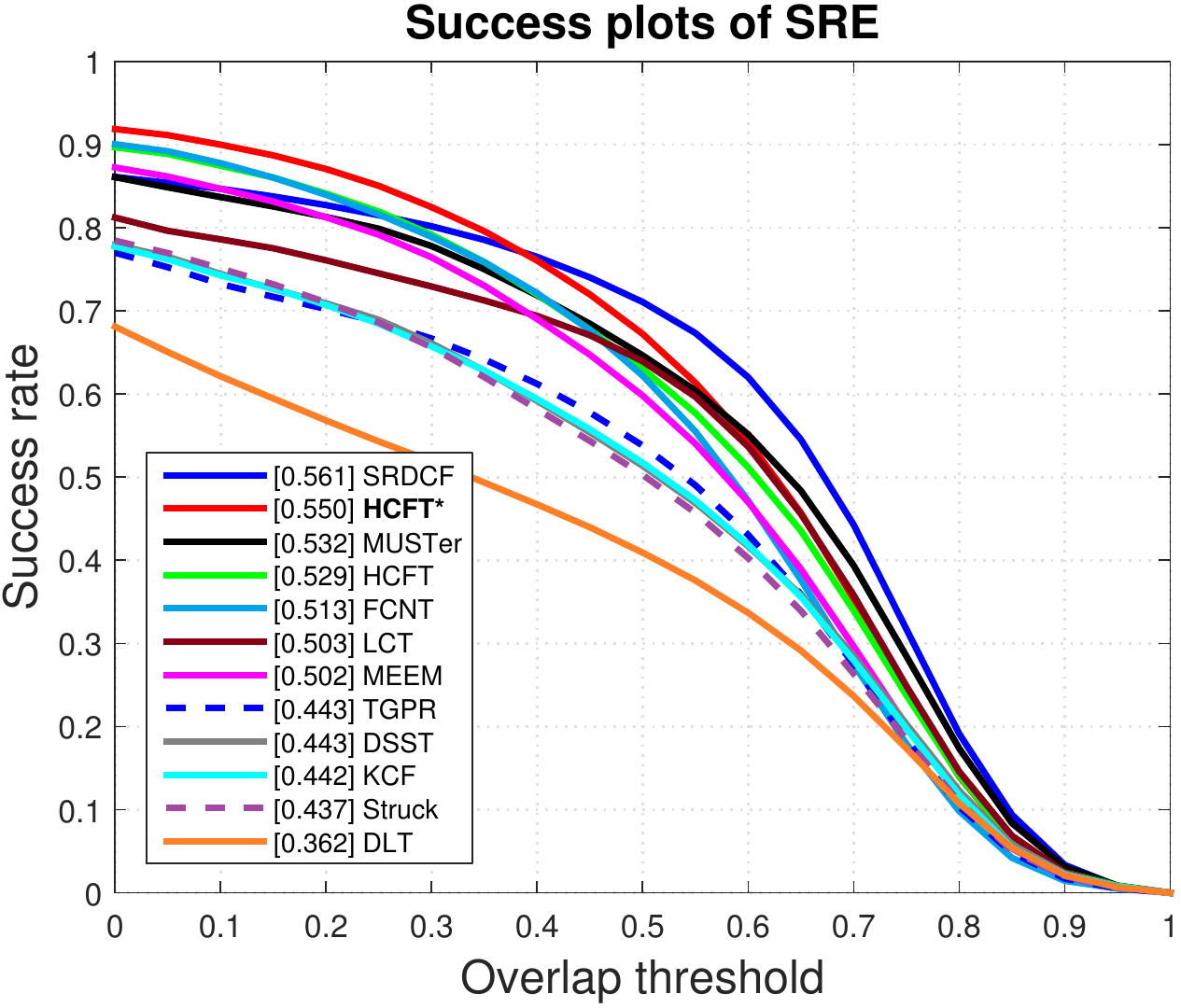} \\
\end{tabular}
\caption{\tb{Distance precision and overlap success plots on the OTB2015~\cite{DBLP:journals/pami/WuLY15} dataset.} Quantitative results on the 100 benchmark sequences using OPE, SRE and TRE.
The legend of distance precision contains threshold scores at 20 pixels, while the legend of overlap success contains area-under-the-curve score
for each tracker. 
The proposed algorithm, HCFT*, performs favorably against the state-of-the-art trackers.
} 
\label{fig:otb100}
\vspace{-2mm}
\end{figure}

\begin{figure}[t]
\centering
\setlength{\tabcolsep}{.25mm}
\begin{tabular}{ccc}
\includegraphics[width=.22\textwidth]{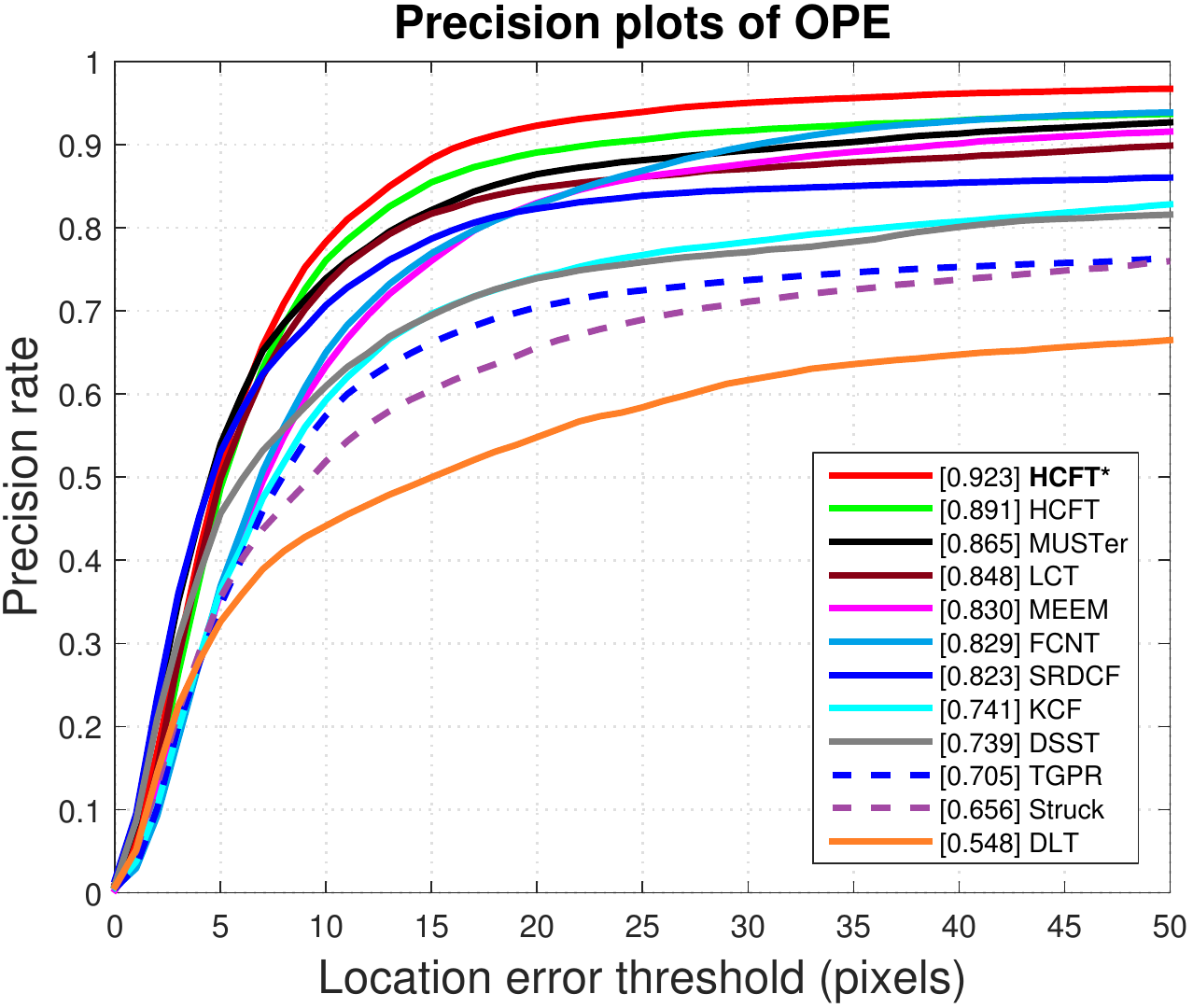} &
\includegraphics[width=.22\textwidth]{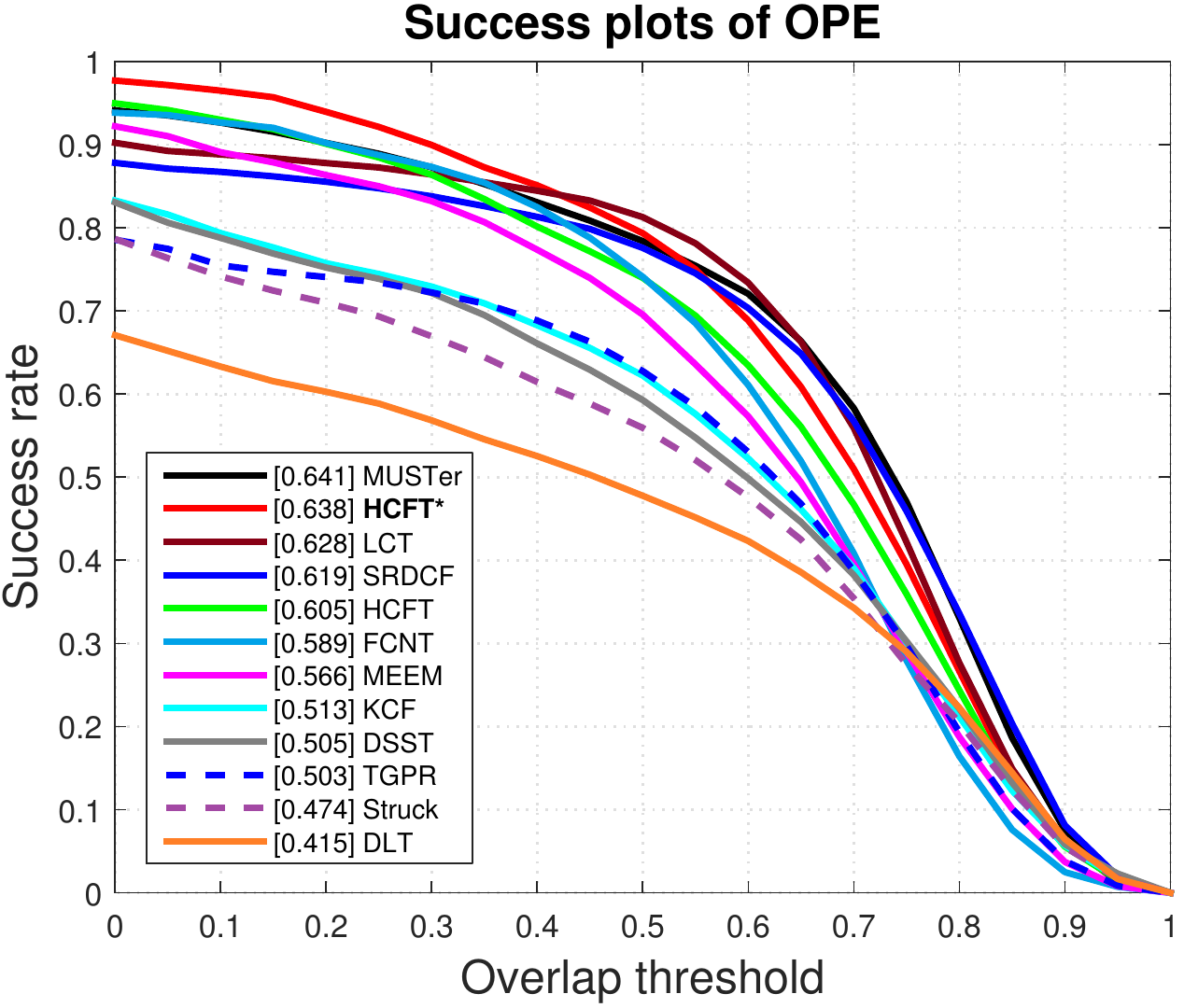} \\
\end{tabular}
\caption{\tb{Distance precision and overlap success plots on the OTB2013~\cite{DBLP:conf/cvpr/WuLY13} dataset.} Quantitative results on the 50
benchmark sequences using OPE. 
The legend of distance precision contains the threshold scores at 20 pixels 
while the legend of overlap success contains area-under-the-curve score
for each tracker. 
Our tracker HCFT* performs well against the state-of-the-art algorithms.
} 
\label{fig:otb50}
\vspace{-2mm}
\end{figure}

%MH: how can a dataset validate algorithms?
%The OTB datasets validate tracking algorithms using three different types of evaluation procedures.
%
Tracking algorithms are evaluated on the OTB datasets using three protocols with either distance precision or overlap success rates \cite{DBLP:conf/cvpr/WuLY13}.
%
%MH: understand the object and verbs
%Given a test sequence, one-pass evaluation (OPE) initiates trackers at the first frame and computes the one-pass accuracy for performance evaluation; temporal robustness evaluation (TRE) splits the sequence into several fragments and computes the average accuracy over all fragments; while spatial robustness evaluation (SRE) spatially shifts the initial bounding box to validate the spatial robustness. 
%
In the one-pass evaluation (OPE) protocol, each tracker is initialized with the ground truth location for evaluation until the end of each sequence. 
For the spatial robustness evaluation (SRE) protocol, each tracker is initialized with the perturbed object location for evaluation of the entire sequence.
In the temporal robustness evaluation (TRE) protocol, each tracker is re-initialized at fixed frame interval until the end of a sequence. 
Figure~\ref{fig:otb100} shows the plots of the distance precision rate and the overlap success rate using all three
protocols. 
Overall, 
the proposed algorithm HCFT*
performs favorably against the state-of-the-art methods in all three 
protocols on the OTB2015 dataset.
For completeness, we present the OPE results against other methods on the OTB2013 dataset in Figure~\ref{fig:otb50}.
We list the quantitative results of distance precision rate at 20 pixels, overlap success rate at 0.5 intersections over union (IoU), center location errors, and tracking speed in Table~\ref{tb:compare}. 
Table~\ref{tb:compare} shows that the proposed algorithm performs well against the state-of-the-art trackers in distance precision (DP) rate, overlap success (OS) rate and center location error (CLE). 
Notice that the OTB2015 dataset is more challenging where 
all the evaluated trackers perform worse than on the OTB2013 dataset.

\begin{figure}[!t]
\centering
\includegraphics[width=.4\textwidth]{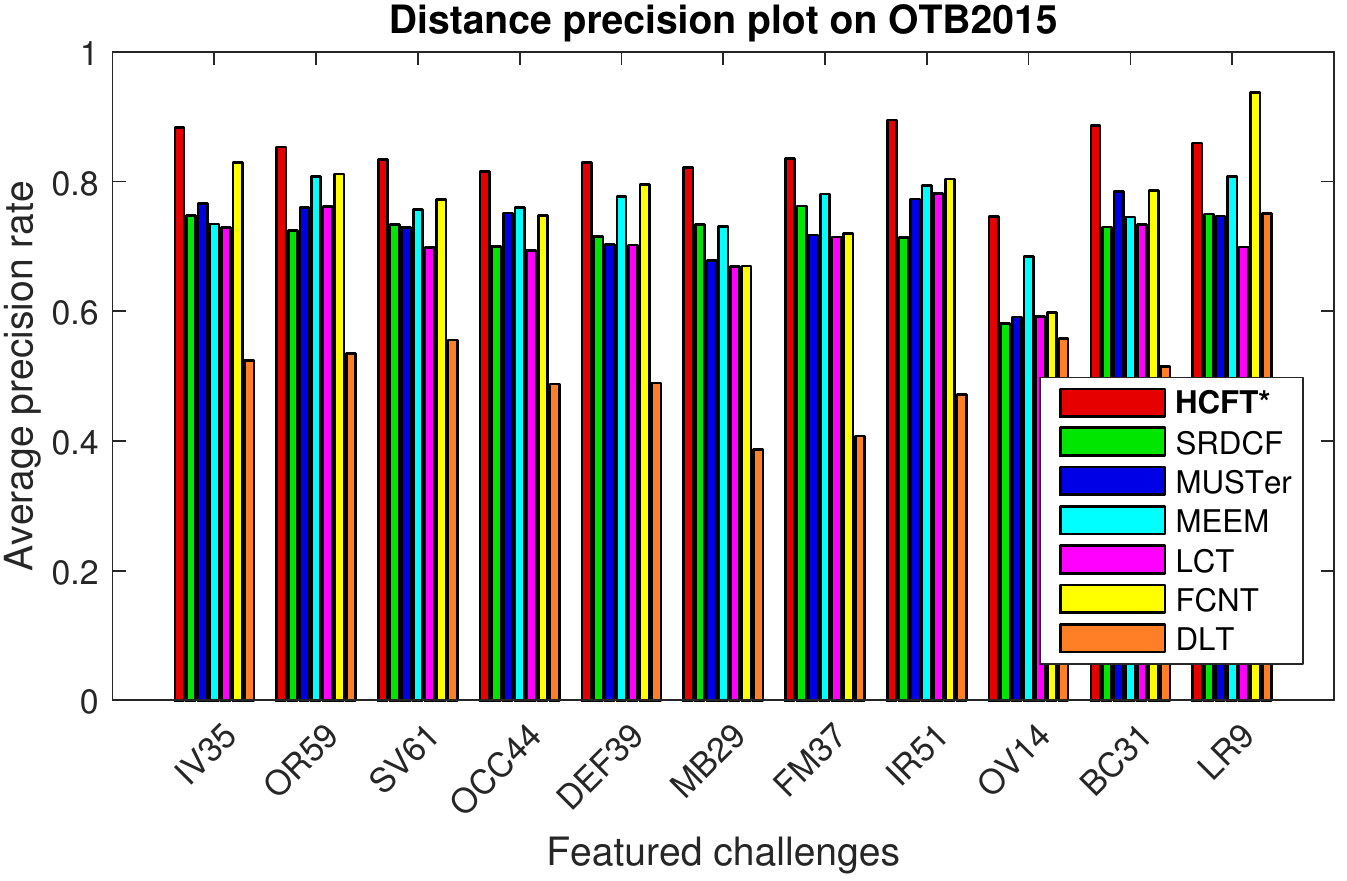} \\ \includegraphics[width=.4\textwidth]{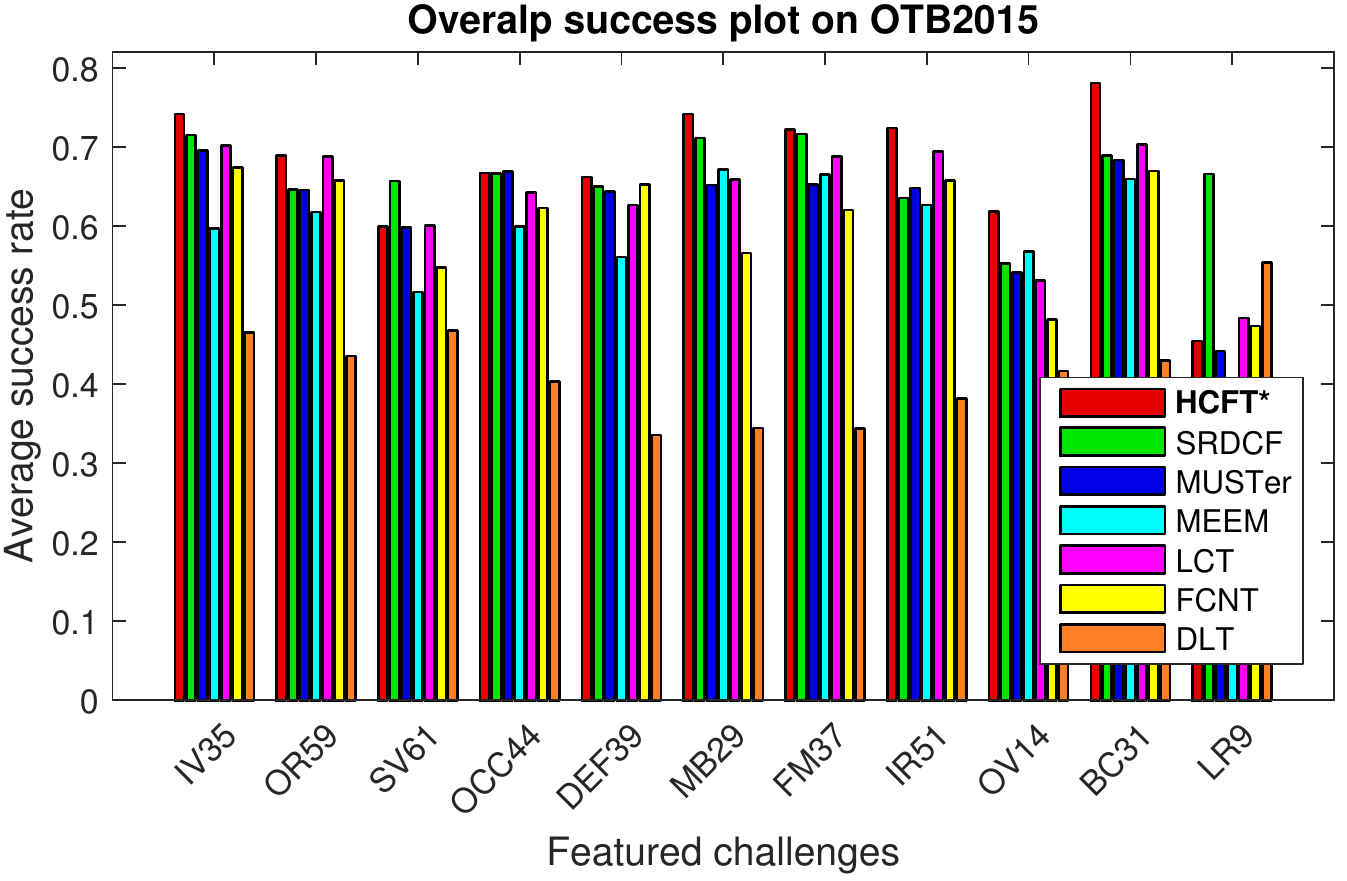} 
\vspace{-2mm}
\caption{\tb{Performance evaluation on benchmark attributes:} illumination variation (IV35), 
out-of-plane rotation (OR59), scale variation (SV61), occlusion (OCC44), deformation (DEF39), motion blur (MB29), fast motion (FM37), in-plane rotation (IR51), out-of-view (OV14), background clutter (BC31), and low resolution (LR9). The later digits in each acronym mean the number of videos with that attribute. 
The proposed algorithm HCFT* performs well against state-of-the-art results.}
\label{fig:attribute} 
\vspace{-4mm}
\end{figure}

Among the state-of-the-art tracking methods, 
the HCFT~\cite{DBLP:conf/iccv/MaHYY15}, FCNT~\cite{DBLP:conf/iccv/WangOWL15} and SRDCF~\cite{DBLP:conf/iccv/DanelljanHKF15} are three concurrently proposed trackers. 
The HCFT* consistently performs well against these three baseline methods.
With the use of coarse-to-fine localization and region proposals for target re-detection as
well as scale estimation, the proposed HCFT* significantly improves the performance in both distance precision (3.3\%) and overlap success (6.5\%) when compared to the HCFT on the OTB2015 dataset. 
For the OTB2013 dataset, the HCFT* gains 
3.2\% in distance precision and 5.3\% in overlap success.
Compared to the HCFT~\cite{DBLP:conf/iccv/MaHYY15}, 
the HCFT* reduces the center location error by 6.3 pixels on the OTB2013 dataset
and 6.1 pixels on the OTB2015 dataset.
%
%The average speed of our tracker is around 7 frames per second.
The runtime performance of the HCFT* is 7 frames per second where
the main computational load lies in the forward propagation
process to extract deep convolutional features 
(around 40\% of the runtime for each frame). 
%
%We maintain another network that only generates \textit{conv3} features for every region proposal for more efficiency. 
%
The components using region proposals for scale estimation and target recovery 
from tracking failures account for another 35\% of the runtime.

%\vspace{-2mm}
\subsubsection{Attribute-based Evaluation}
We analyze the tracker performance using 11 annotated attributes in the OTB2015~\cite{DBLP:journals/pami/WuLY15} dataset: 
illumination variation (IV35),
out-of-plane rotation (OR59),
scale variation (SV61),
occlusion (OCC44),
deformation (DEF39),
motion blur (MB29),
fast motion (FM37),
in-plane rotation (IR51),
out-of-view (OV14),
background clutter (BC31), and
low resolution (LR9) 
(number of videos for each attribute is appended to the end of each abbreviation). 
Figure~\ref{fig:attribute} presents the results under one-pass evaluation regarding these challenging attributes for visual object tracking.
%
%From Figure~\ref{fig:attribute}, we have the following observations.
%
First, the proposed tracker HCFT*, as well as the HCFT~\cite{DBLP:conf/iccv/MaHYY15} and FCNT~\cite{DBLP:conf/iccv/WangOWL15}, generally perform well 
in scenes with background clutters, which can be attributed to 
the use of deep feature hierarchies capturing semantics and spatial details.
The DLT~\cite{DBLP:conf/nips/WangY13} pre-trains neural networks in an unsupervised 
manner for visual tracking, and it does not perform well in all attributes. 
This shows that CNN features trained with category-level supervision 
for object representations are more discriminative. 
Second, as the last layer of the pre-trained CNN model retains semantic information insensitive to scale changes (Figure~\ref{fig:vis}(d)), even without the scale estimation component, the HCFT still achieves favorable performance in the presence of scale variations.
Third, correlation filters trained over each convolutional layer can be viewed 
as holistic templates of target objects on different abstraction levels. 
%
%For scenes with occlusion and targets moving out-of-view, 
%the HCFT does not perform as well as in the other attributes. 
On the other hand, the HCFT does not perform well for scenes with occlusion and out-of-view.
With the use of recovery module based on object proposals, the proposed HCFT* 
achieves significant gains of 4.0\% and 6.1\% for scenes with occlusion and out-of-view attributes. 

\begin{figure}[t]
\centering
\small
\setlength{\tabcolsep}{0em}
\begin{tabular}{cc}
\includegraphics[width=0.22\textwidth]{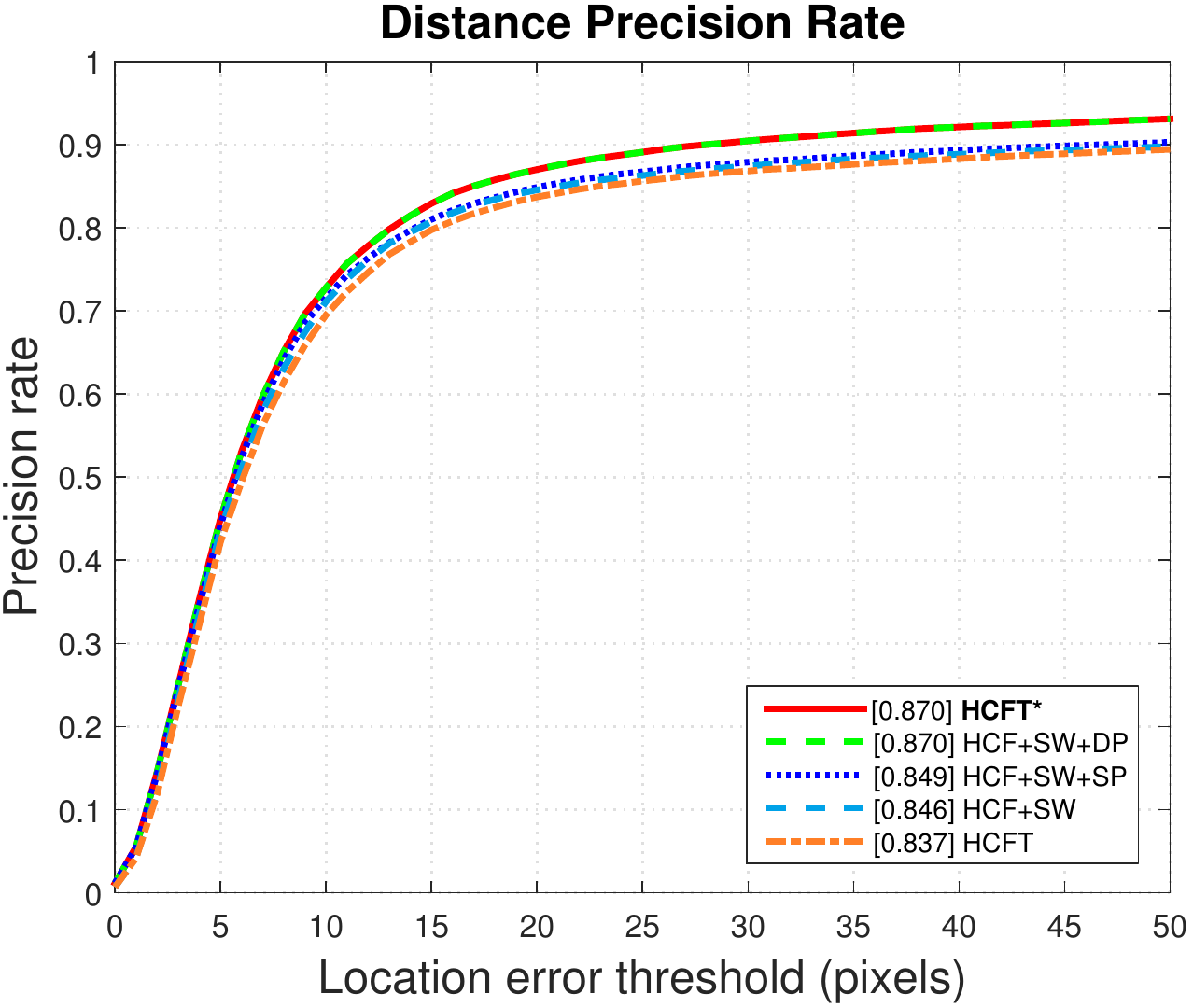} &
\includegraphics[width=0.22\textwidth]{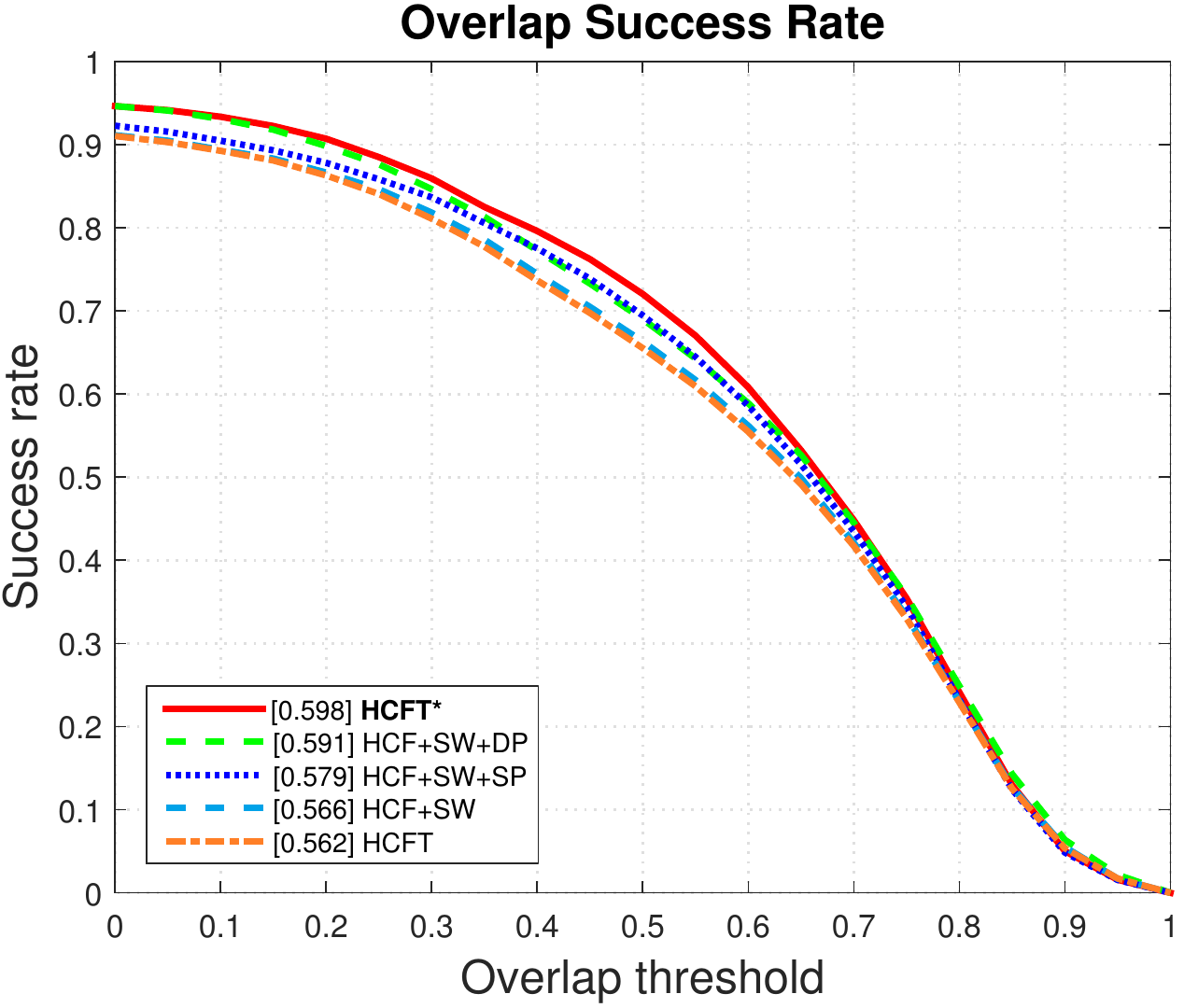}\\
\end{tabular}
\caption{\tb{Ablation study.} Component analysis on the OTB2015~\cite{DBLP:journals/pami/WuLY15} dataset with comparisons to the HCFT~\cite{DBLP:conf/iccv/MaHYY15}. SW: soft weight. SP: scale proposals for scale estimation. DP: detection proposals for target re-detection. The values at the legends with distance precision is based on the threshold of 20 pixels while values at the legend of overlap success are based on area under the curve.The proposed HCFT* incorporates all of these components and achieves performance gains of +3.3\% and +3.6\% in distance precision and overlap success when compared to the HCFT.}
\label{fig:component2}
\vspace{-2mm}
\end{figure} 

%\vspace{-2mm}
\subsubsection{Component Analysis}
\label{sec:component}
Compared with the HCFT, there
are three additional components in HCFT*: 
(1) a soft weight scheme to combine hierarchical correlation response maps, 
(2) object proposal for scale estimation, and 
(3) object proposals for target re-detection. 
In Figure~\ref{fig:component2}, we demonstrate the effectiveness 
of these three schemes on the OTB2015~\cite{DBLP:journals/pami/WuLY15} dataset.
The baseline HCF-SW method 
indicates the hierarchical convolutional features (HCF) with the soft weight (SW) scheme. 
The HCF-SW-SP scheme adds the scale estimate module using 
object proposals (SP) based on the HCF-SW approach, 
while the HCF-SW-DP method 
adds the re-detection scheme using object proposals (DP) based on the HCF-SW module. 
The proposed HCFT* incorporates all of these components (HCF-SW-SP-DP) and achieves considerably large performance gains in both distance precision and overlap success when compared to the HCFT~\cite{DBLP:conf/iccv/MaHYY15}.

\begin{figure}[t]
\centering
\small
\setlength{\tabcolsep}{0em}
\begin{tabular}{cc}
\includegraphics[width=0.22\textwidth]{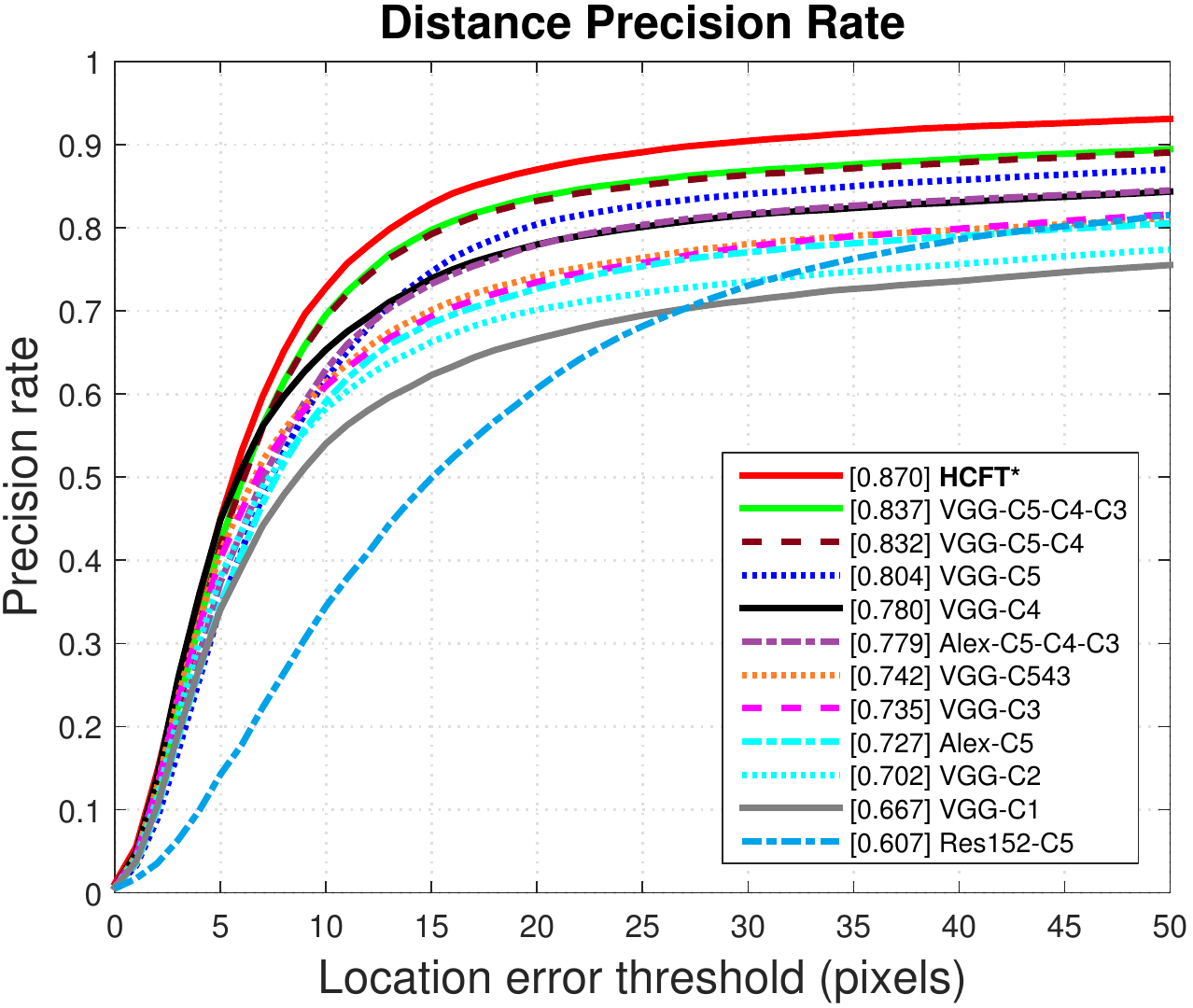} &
\includegraphics[width=0.22\textwidth]{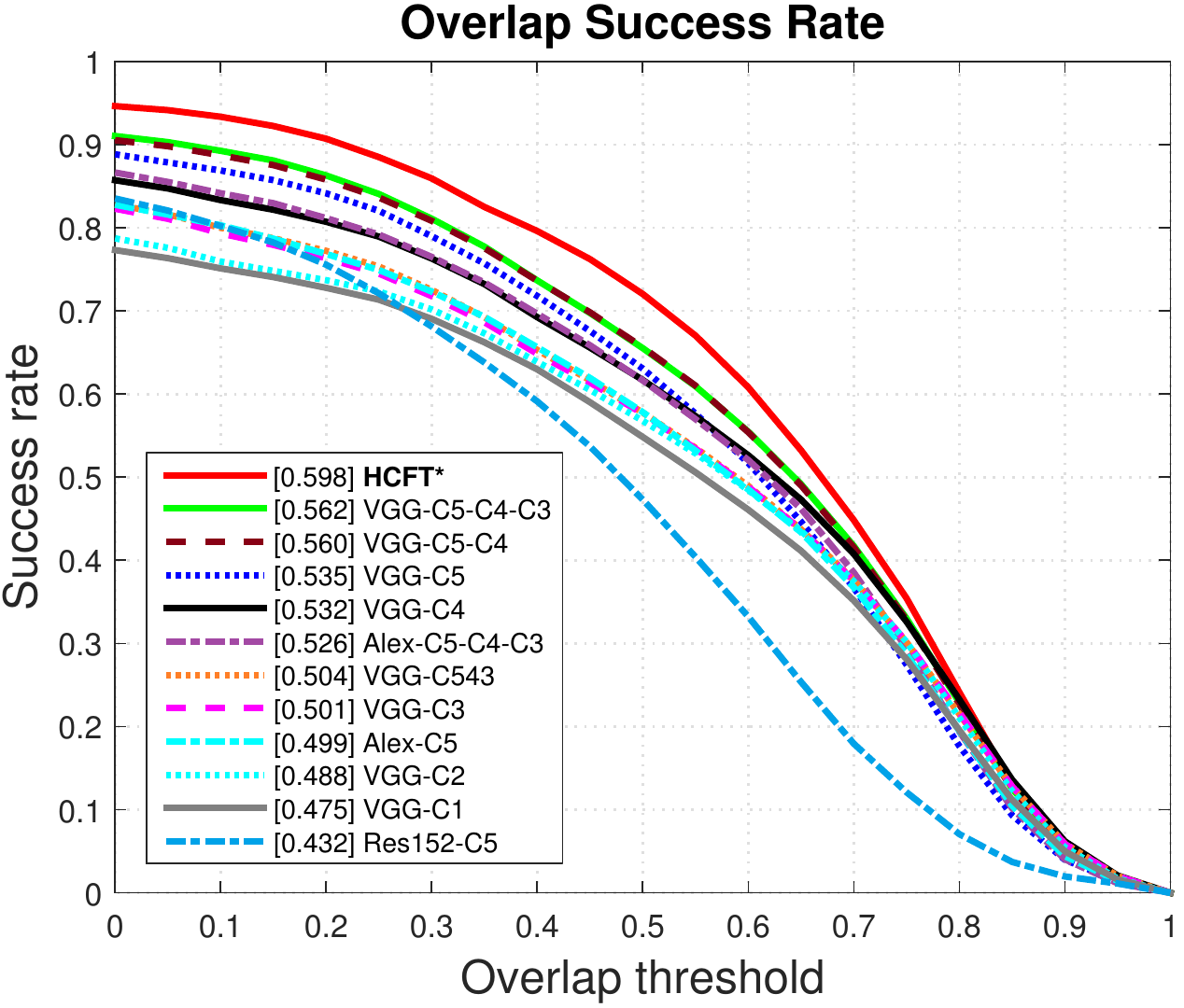}\\
\end{tabular}
\caption{\tb{Performance evaluation using different convolutional layers 
from the VGGNet-19 (VGG)~\cite{DBLP:journals/corr/SimonyanZ14a},
AlexNet (Alex)~\cite{DBLP:conf/nips/KrizhevskySH12}
and ResNet-152 (Res152)~\cite{DBLP:conf/cvpr/HeZRS16}
on the OTB2015~\cite{DBLP:journals/pami/WuLY15} dataset.}
Each single layer (C5, C4, C3, C2 and C1), the combination of correlation response maps from the \textit{conv5-4} and \textit{conv4-4} layers (C5-C4), and concatenation of three convolutional layers (C543) are evaluated. 
By exploiting the hierarchical information from multiple convolutional layers, the VGG-C5-C4-C3 (HCFT~\cite{DBLP:conf/iccv/MaHYY15}) tracker performs well against alternative baseline implementations. 
Overall, the proposed HCFT* performs favorably
with the use of region proposals for scale estimation and target recovery.}
\label{fig:component}
\vspace{-2mm}
\end{figure}

\subsubsection{Feature Analysis}
\label{subsec:fa}
To analyze the effectiveness of the proposed algorithm, we 
compare the tracking results 
using different convolutional layers as features on the OTB2015 dataset.
Figure~\ref{fig:component} shows 11 alternative baseline methods using one-pass evaluation, where the values at the legends with distance precision is based on the threshold of 20 pixels while values at the legend of overlap success are based on area under the curve (AUC). 
We first use each single convolutional layer (C5, C4, C3, C2 and C1) of the VGGNet-19 model 
to represent objects in the proposed algorithm.
Note that if one convolutional layer contains multiple sub-layers, we use features from the last sub-layer, e.g., C5 indicates the \textit{conv5-4} layer in VGGNet-19.
%
%Figure~\ref{fig:component} shows that the deeper the VGGNet layer is, the better the tracking performance is. 
Figure~\ref{fig:component} shows better tracking performance can be obtained when a deeper layer of the VGGNet is used, which 
can be attributed to the semantic abstraction from a deeper network. 
We then perform the coarse-to-fine search scheme on the fifth and fourth layers (C5-C4).
The VGG-C5-C4 method performs better than VGG-C5 and worse than the VGG-C5-C4-C3 scheme (as used in HCFT~\cite{DBLP:conf/iccv/MaHYY15}). 
The quantitative results show that hierarchical inference on the translation cues from multiple CNN layers helps improve tracking performance.
Moreover, we concatenate the fifth, fourth, and third layers together (C543) as multiple channel features similar to the hypercolumns used in~\cite{DBLP:journals/cvpr/HariharanAGM14a}.
However, such concatenation breaks down the hierarchical features over CNN layers (learning a single correlation filter over C543 equally weighs each CNN layer) and thus does not perform well for visual tracking.
Given that the second (C2) and first (C1) layers do not perform well independently and there exists slight performance gain from VGG-C5-C4 to HCFT (C5-C4-C3), we do not explore the feature hierarchies from the second or first layers.
Since features extracted from the VGGNet are generally more effective than the AlexNet in visual tracking, we attribute the performance gain to a deeper network architecture. 
We further implement an alternative baseline method (Res-C5) using the 
last convolutional layer of the ResNet-152~\cite{DBLP:conf/cvpr/HeZRS16}. 
However, we find that Res-C5 method does not perform well. 
We also evaluate the performance of other layers of the ResNet and do not obtain 
performance gain as well.
These results may be explained by the fact that the ResNet uses skip connections to combine different layers. 
Thus, the ResNet does not maintain an interpretable coarse-to-fine feature hierarchy 
from the last layer to early layers as the VGGNet does. 
Each layer of ResNet contains both semantical abstraction and spatial details 
from earlier layers.

\begin{figure}[t]
\centering
\includegraphics[width=.44\textwidth]{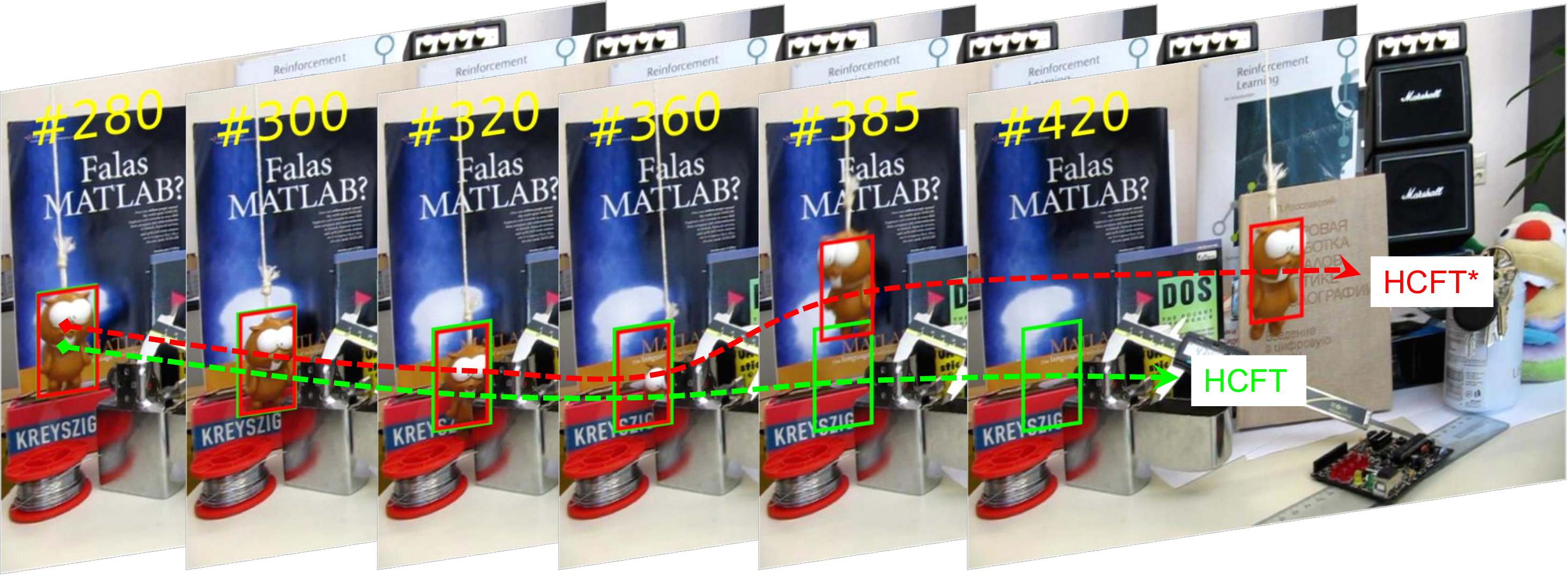} \\
\vspace{1em}
\setlength{\tabcolsep}{.1em}
\begin{tabular}{cc}
\includegraphics[width=.22\textwidth]{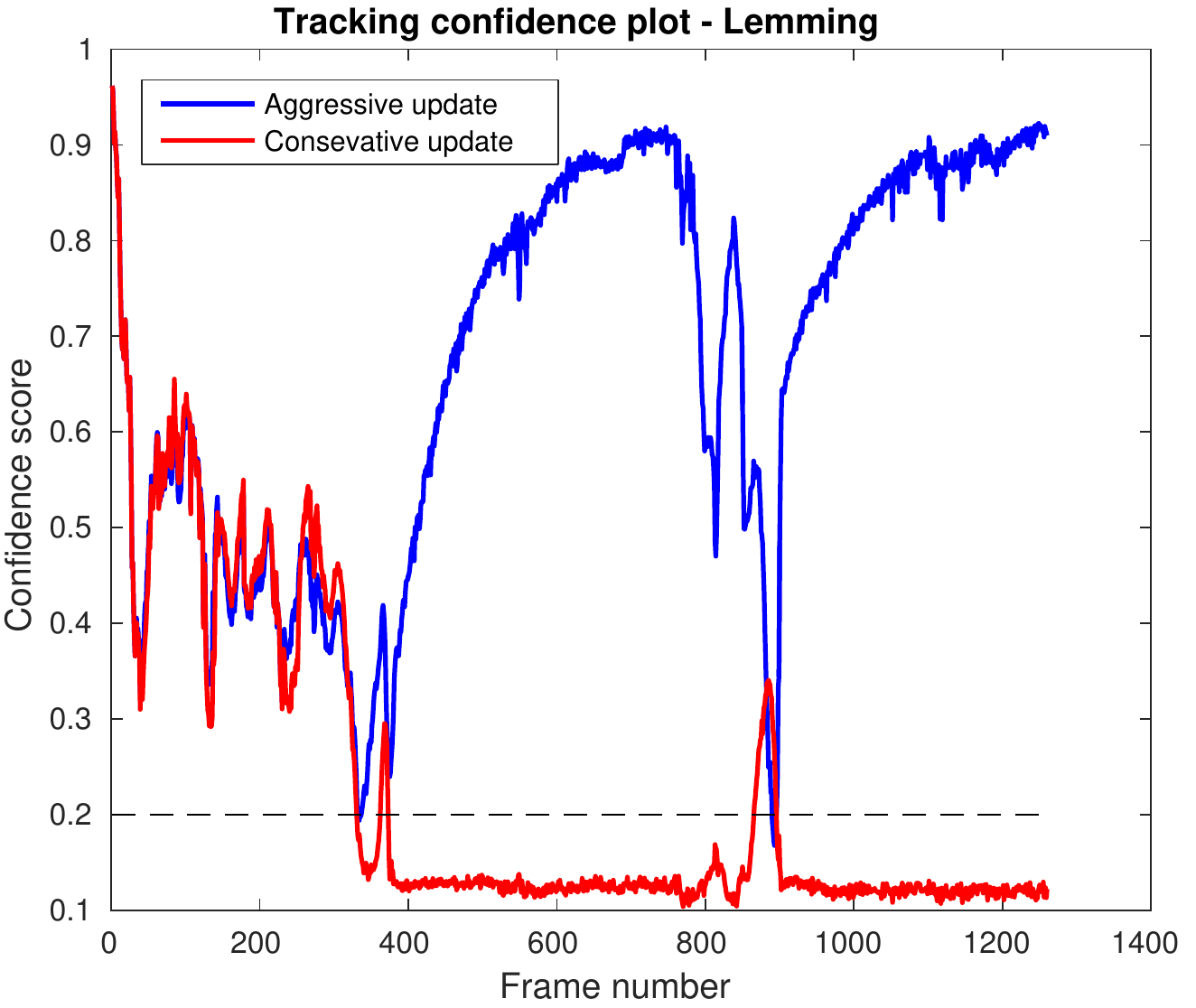} &
\includegraphics[width=.22\textwidth]{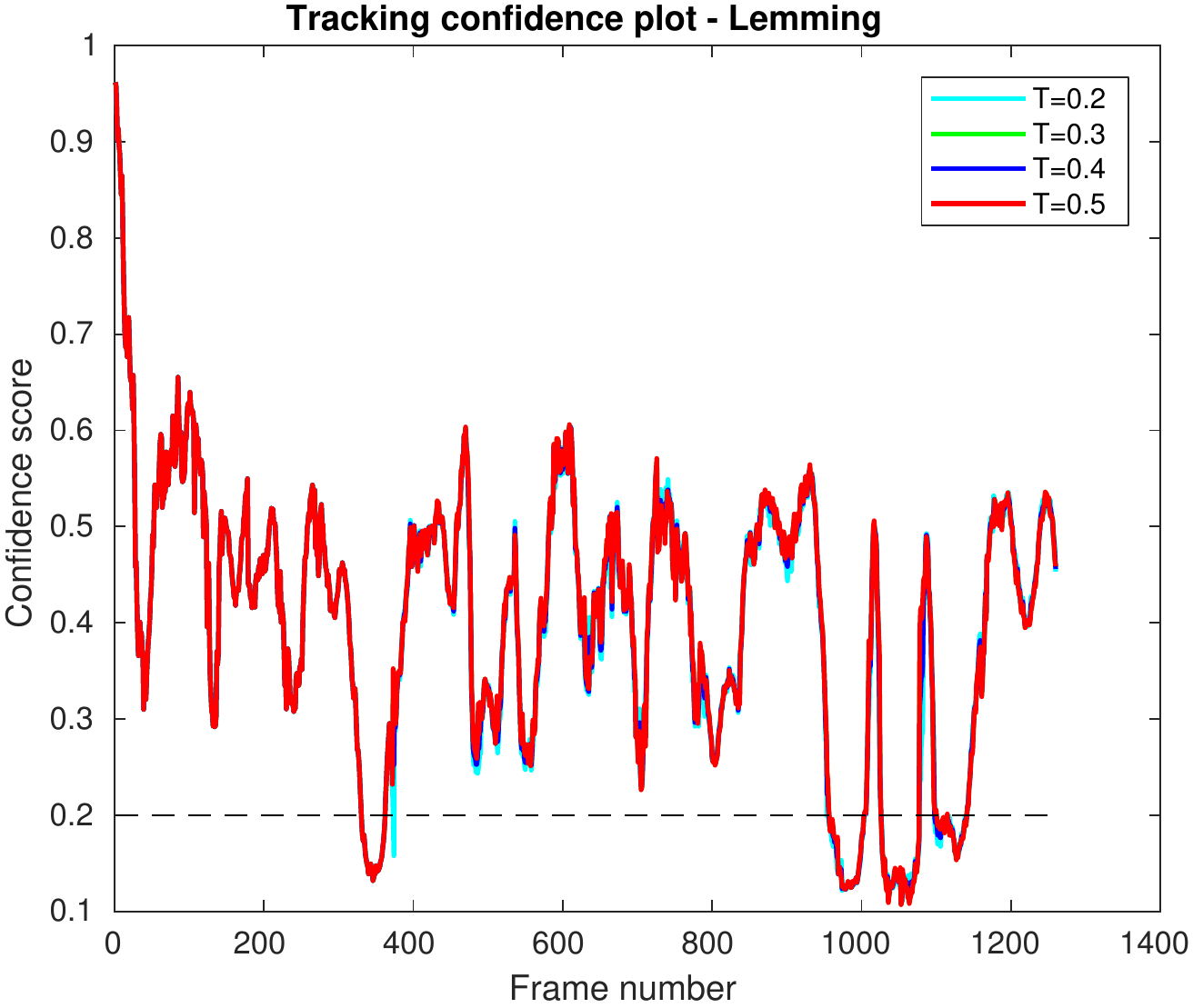}\\
% (b) $g(\cdot)$ with different learning rate & (c) \\
\end{tabular}
\caption{\tb{Classifier analysis on the \textit{Lemming} sequence.}
The target object is under heavy occlusion in the 360th frame. \textbf{Bottom left:} Without a re-detection module, the HCFT~\cite{DBLP:conf/iccv/MaHYY15} fails after the 360th frame. 
A correlation filter $g(\cdot)$ with an aggressive learning rate cannot predict tracking failures after the 360th frame due to noisy updates. 
In contrast, a conservatively learned filter $g(\cdot)$ accurately predicts the confidence score of every tracking result. 
\textbf{Bottom right:} For the proposed HCFT* with a re-detection module, we test the update threshold for learning classifier $g(\cdot)$. Confidence scores are consistent with threshold values between 0.2 and 0.5.}
\label{fig:lemming}
\vspace{-2mm}
\end{figure}

\begin{figure*}[t]
\centering
\footnotesize
\setlength{\tabcolsep}{0pt}
\begin{tabular}{cccccc}
\includegraphics[trim = 0mm 3mm 2mm 1mm, clip,width=.16\textwidth]{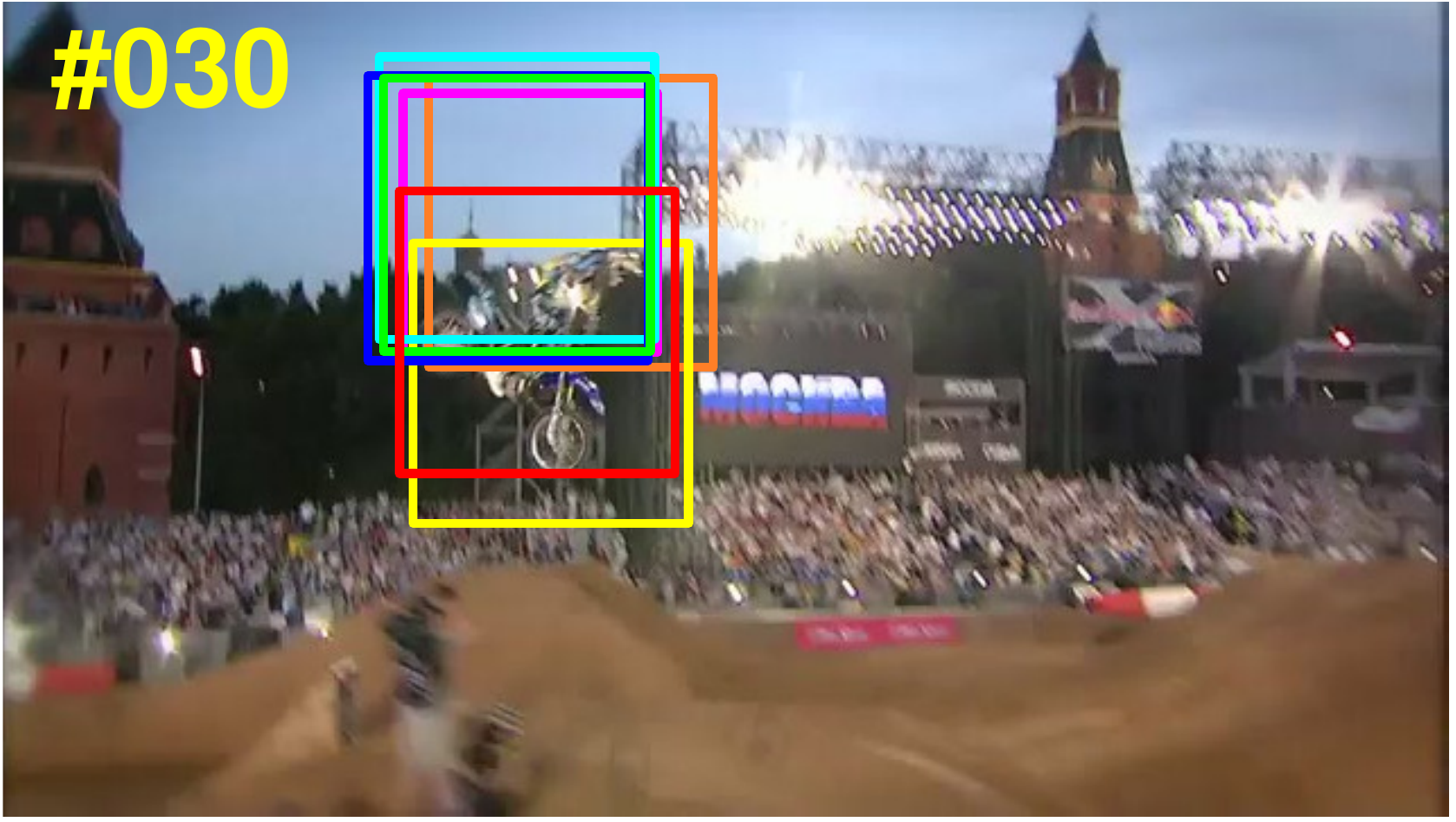} &
\includegraphics[trim = 0mm 3mm 2mm 1mm, clip,width=.16\textwidth]{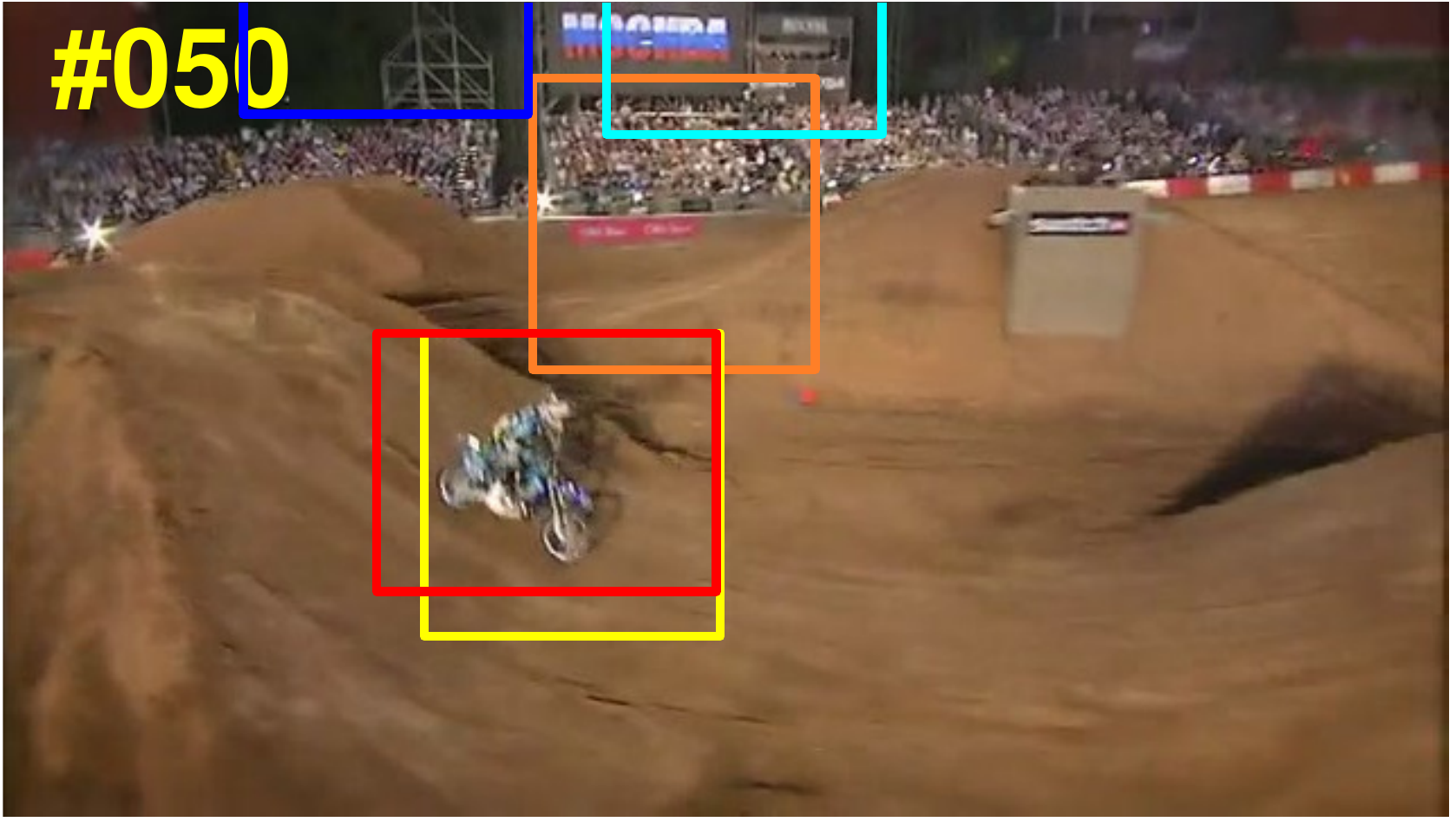} &
\includegraphics[trim = 0mm 3mm 2mm 1mm, clip,width=.16\textwidth]{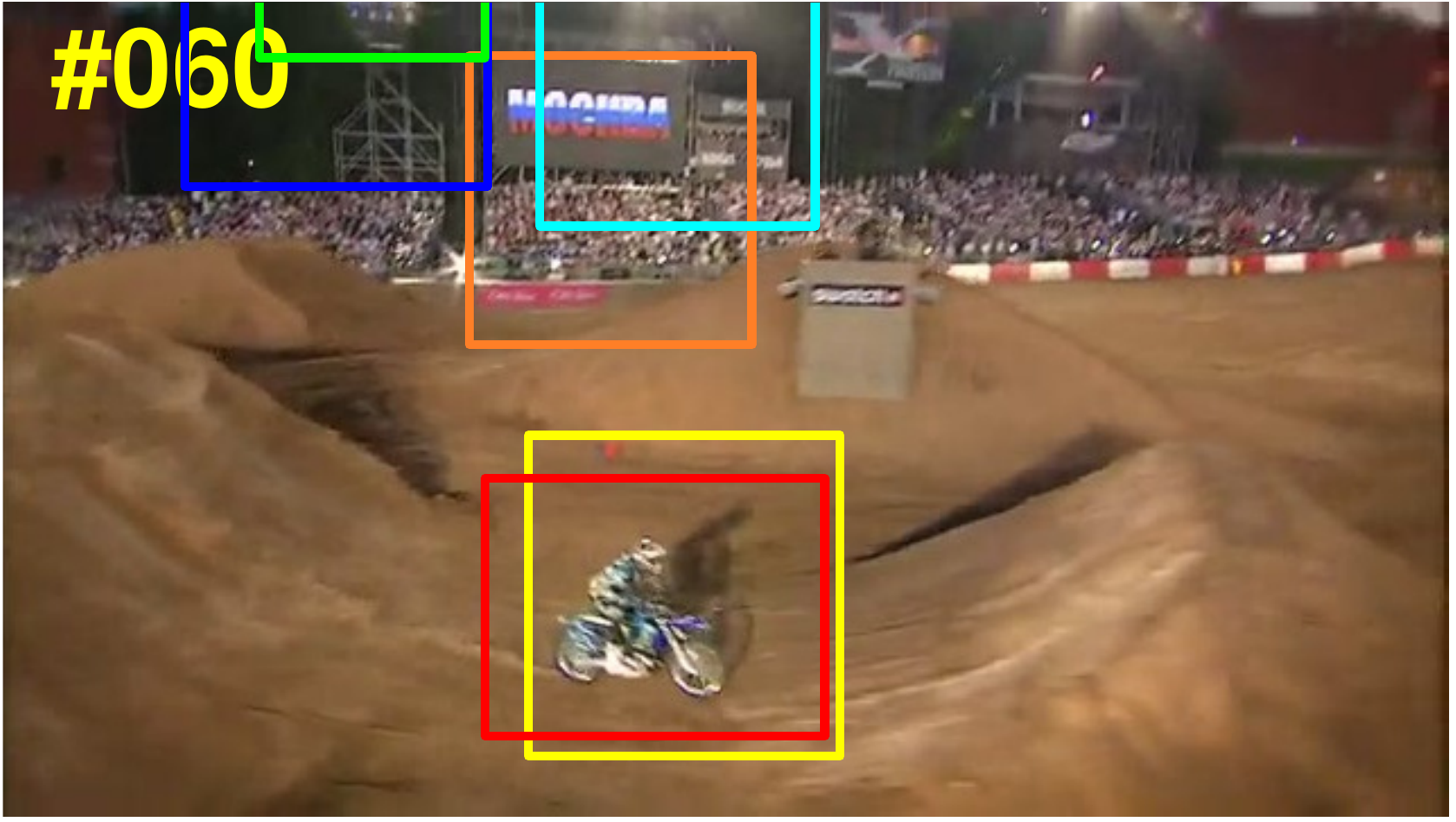} &
\includegraphics[trim = 0mm 3mm 2mm 1mm, clip,width=.16\textwidth]{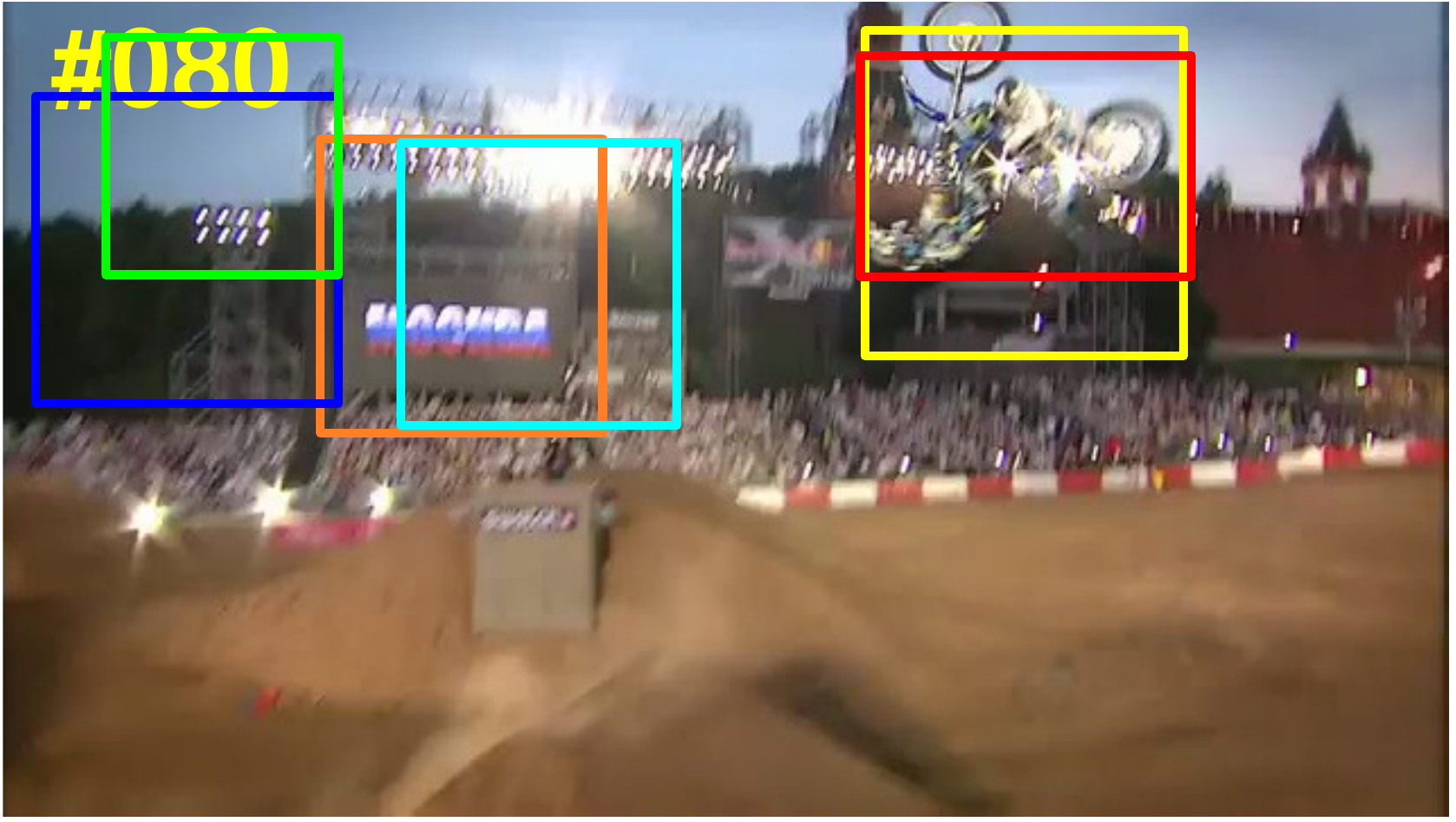}& 
\includegraphics[trim = 0mm 3mm 2mm 1mm, clip,width=.16\textwidth]{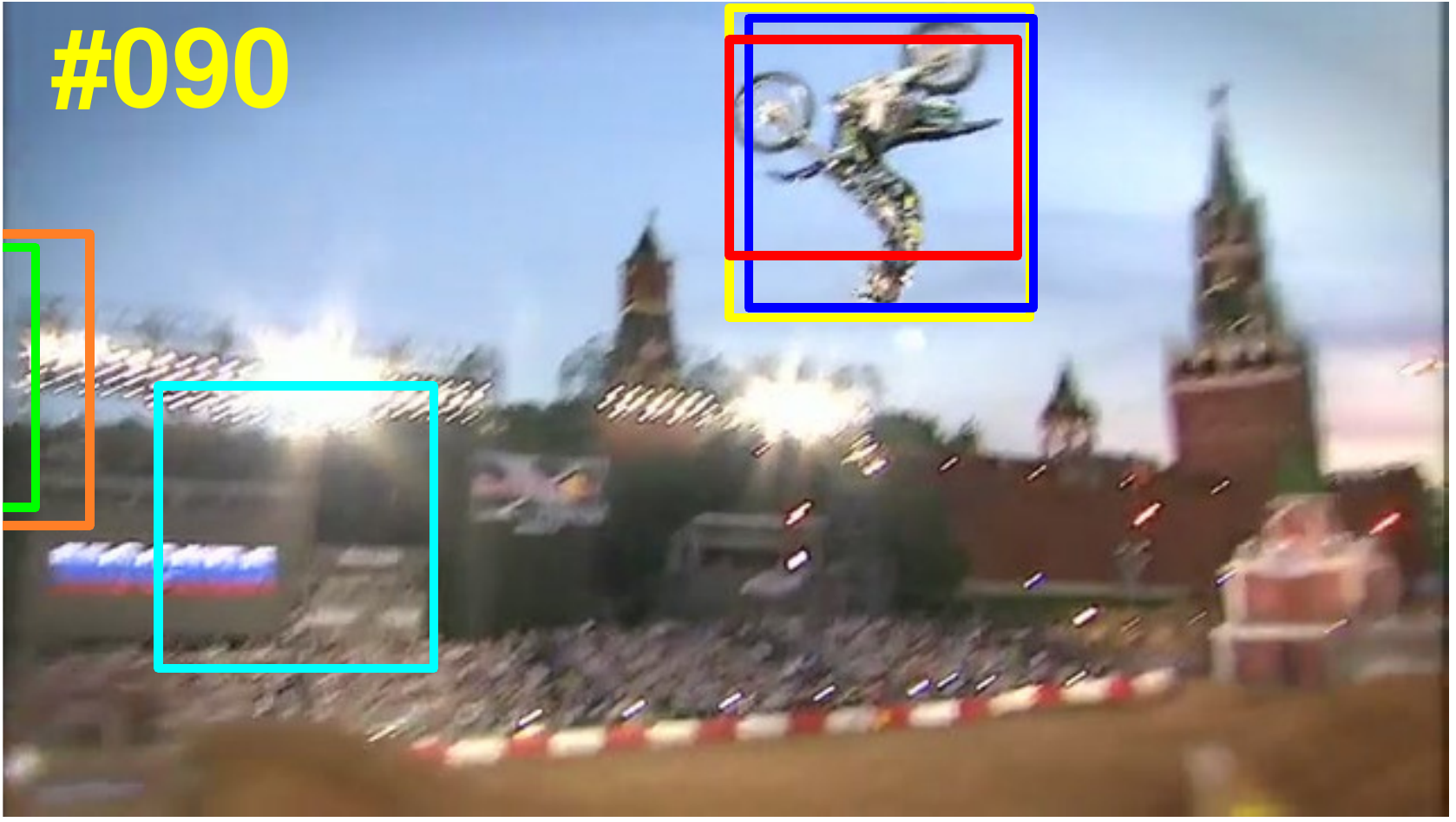} &
\includegraphics[trim = 0mm 3mm 2mm 1mm, clip,width=.16\textwidth]{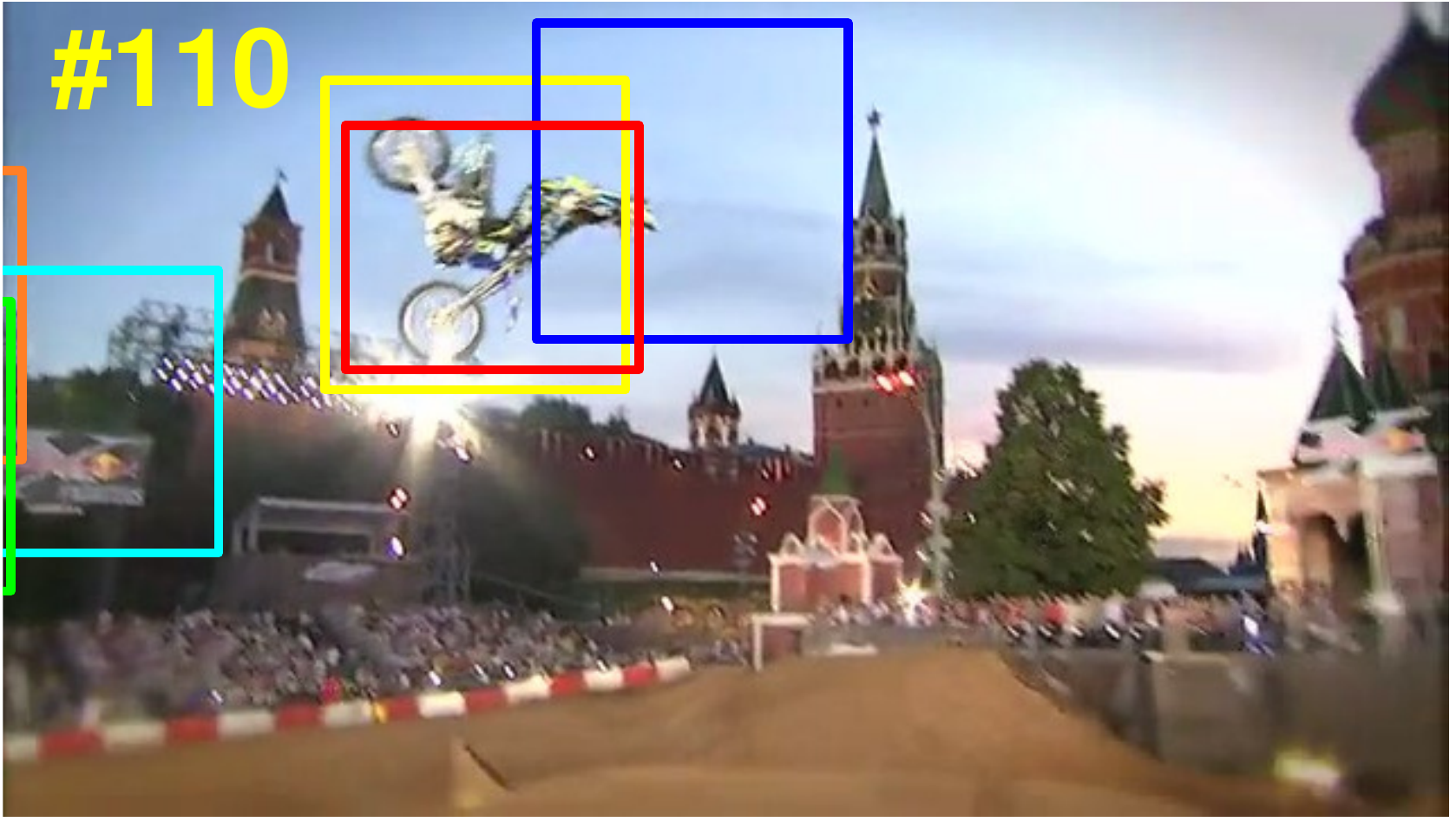}\\ 
\includegraphics[trim = 30mm 2mm 30mm 1mm,
clip,width=.16\textwidth]{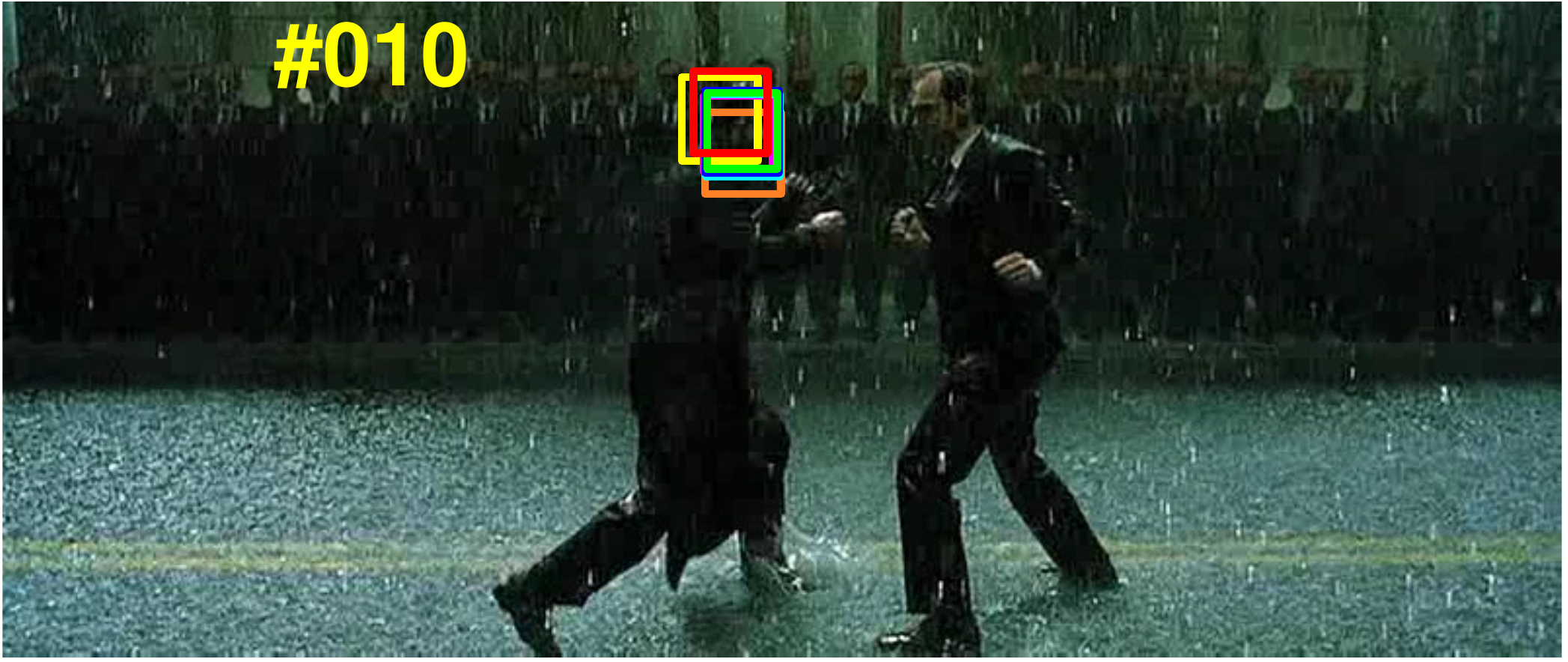} &
\includegraphics[trim = 30mm 2mm 30mm 1mm, clip,width=.16\textwidth]{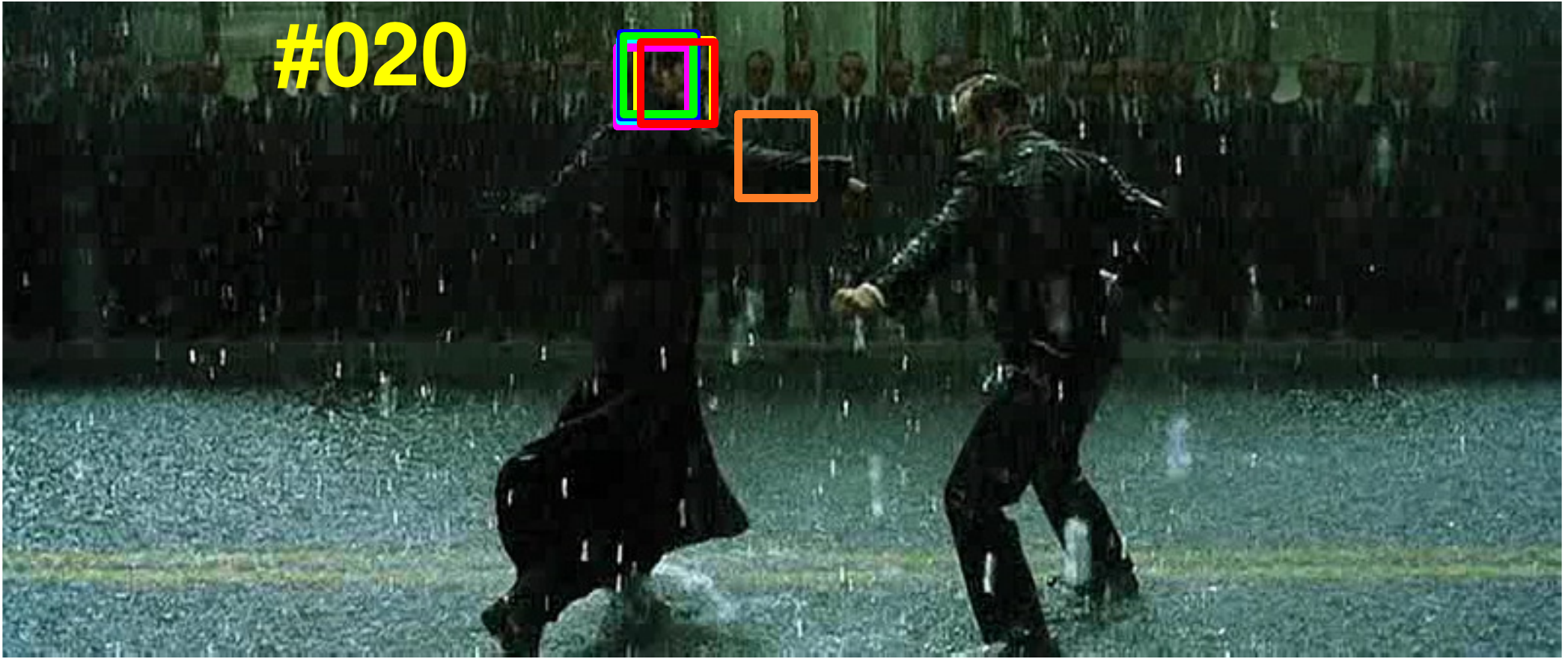} &
\includegraphics[trim = 30mm 2mm 30mm 1mm,
clip,width=.16\textwidth]{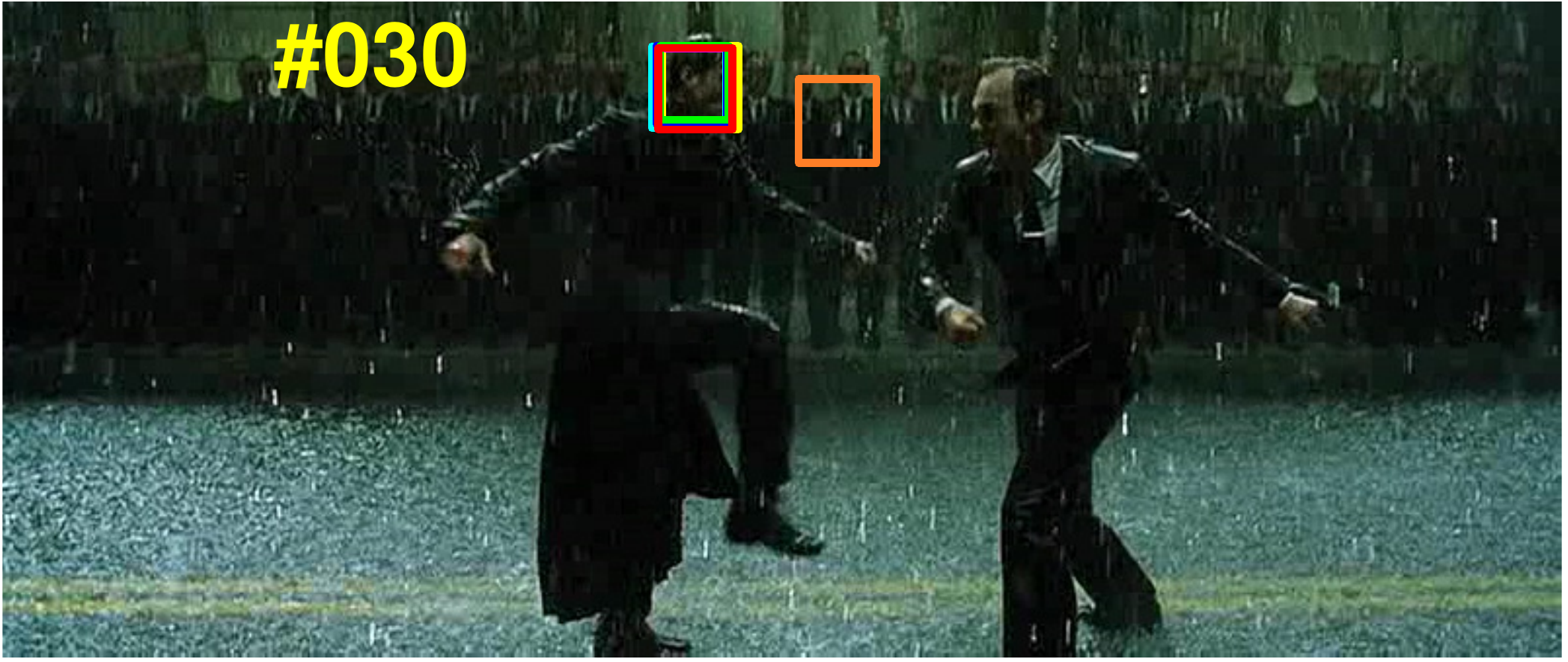} &
\includegraphics[trim = 30mm 2mm 30mm 1mm, clip,width=.16\textwidth]{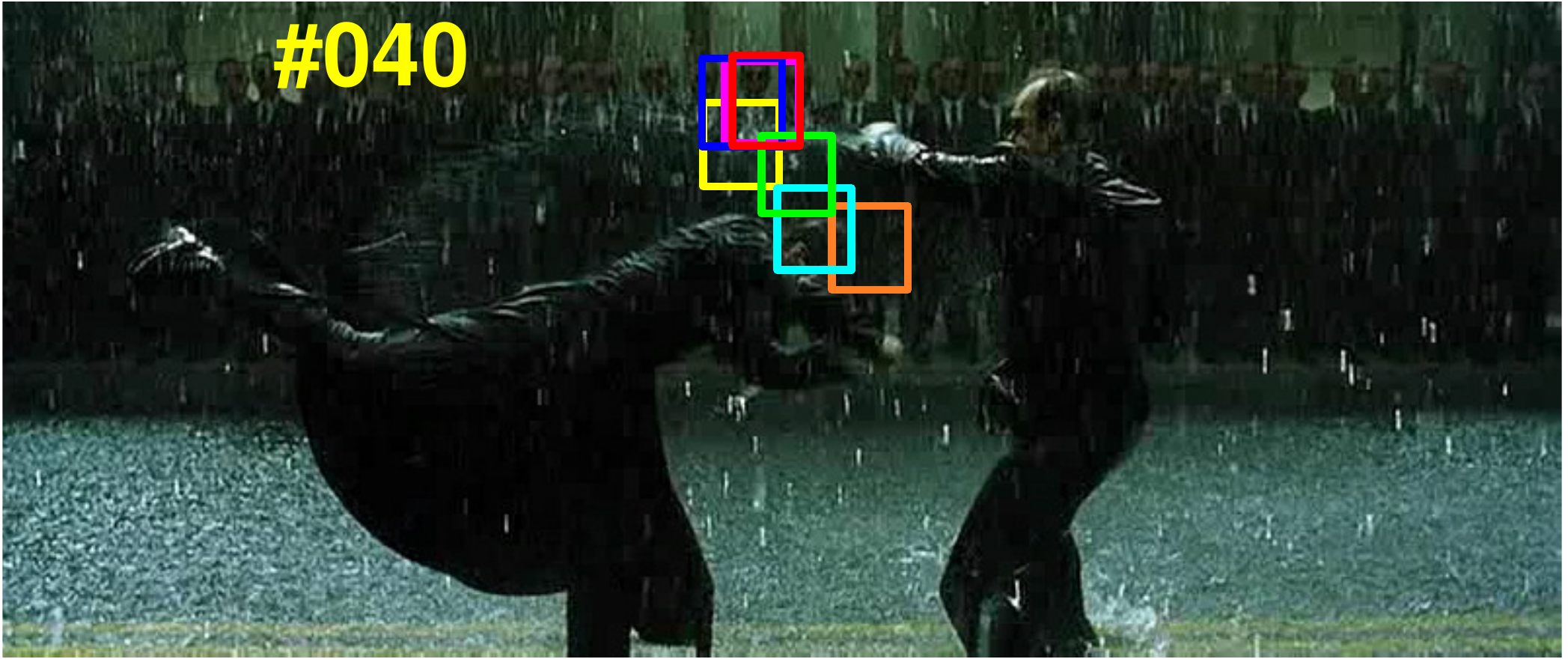} & 
\includegraphics[trim = 30mm 2mm 30mm 1mm, clip,width=.16\textwidth]{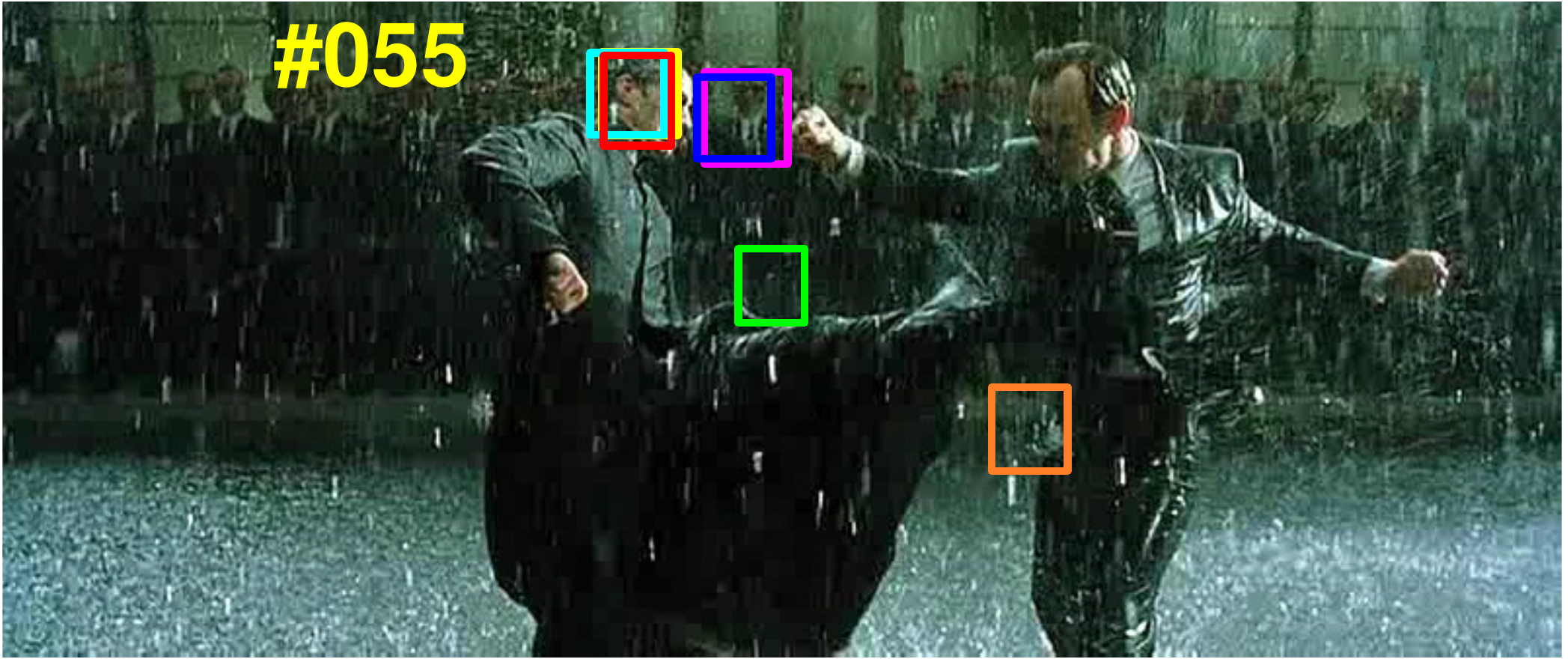} &
\includegraphics[trim = 30mm 2mm 30mm 1mm, clip,width=.16\textwidth]{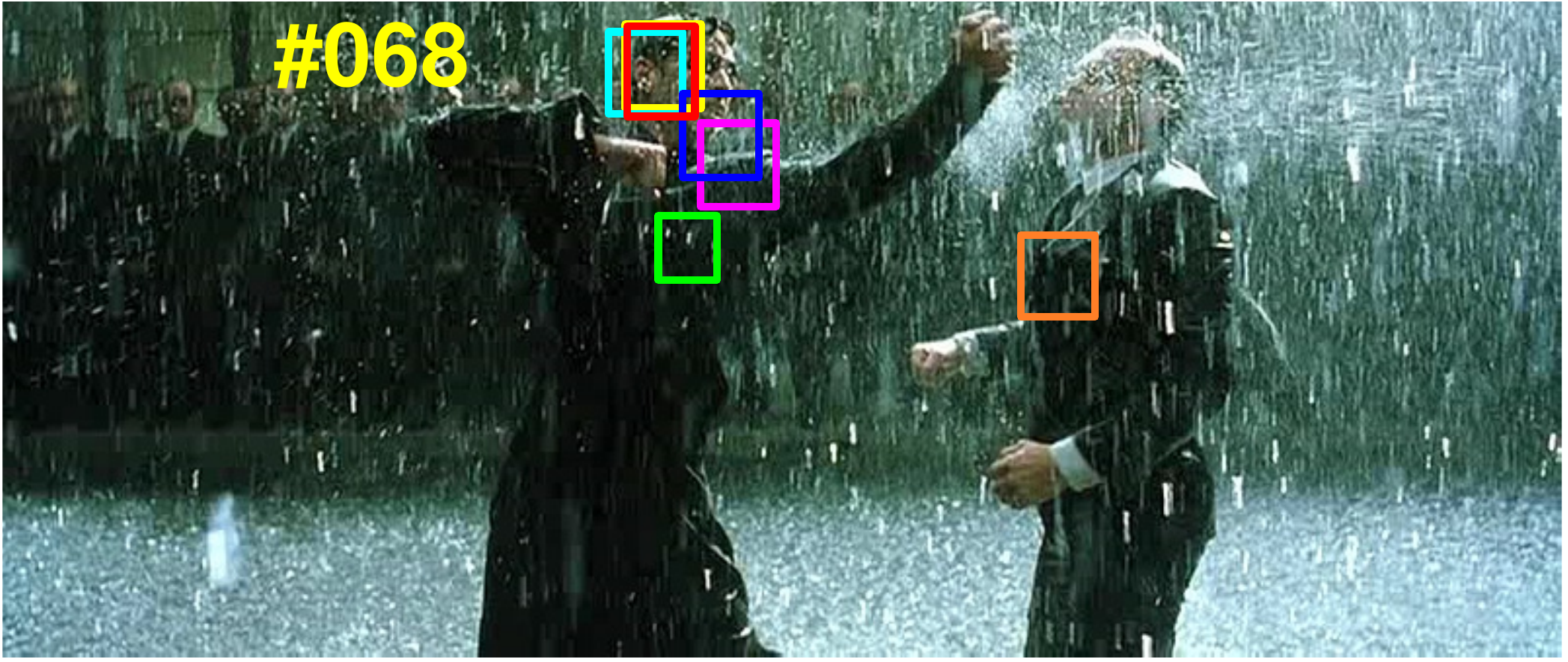} \\
\includegraphics[trim = 2mm 2mm 5mm 1mm, clip,width=.16\textwidth]{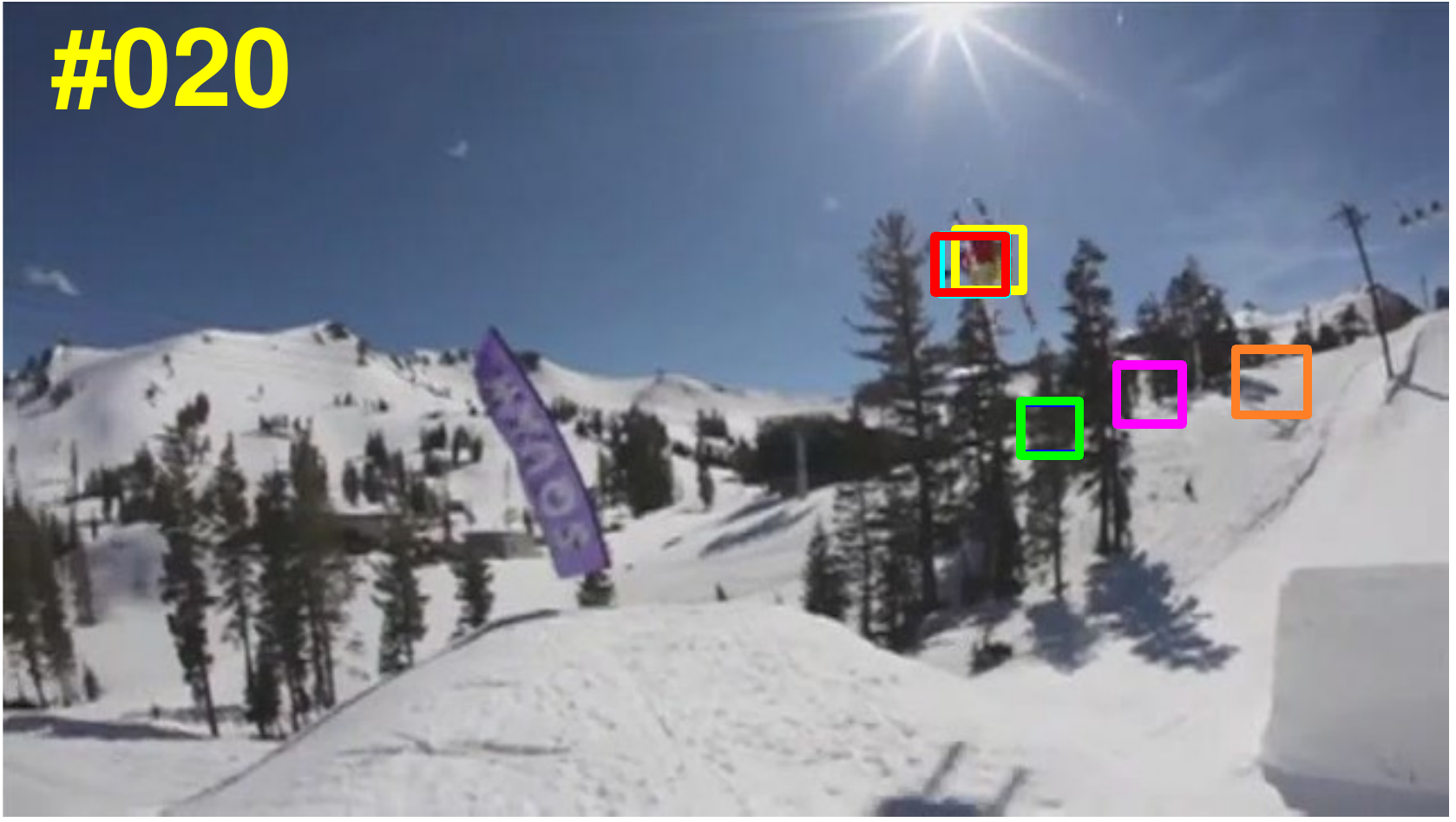}&
\includegraphics[trim = 2mm 2mm 5mm 1mm, clip,width=.16\textwidth]{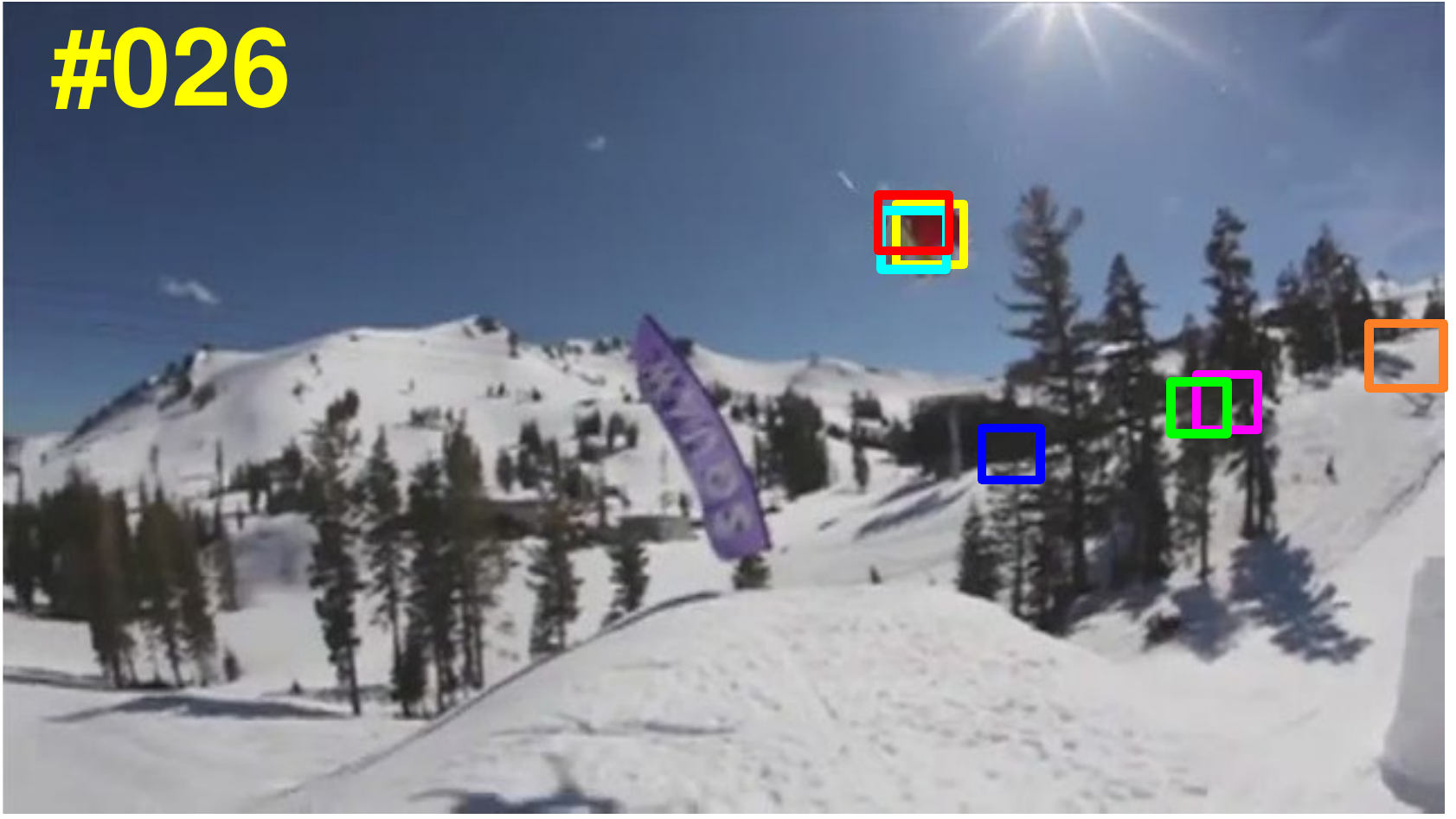}& 
\includegraphics[trim = 2mm 2mm 5mm 1mm, clip,width=.16\textwidth]{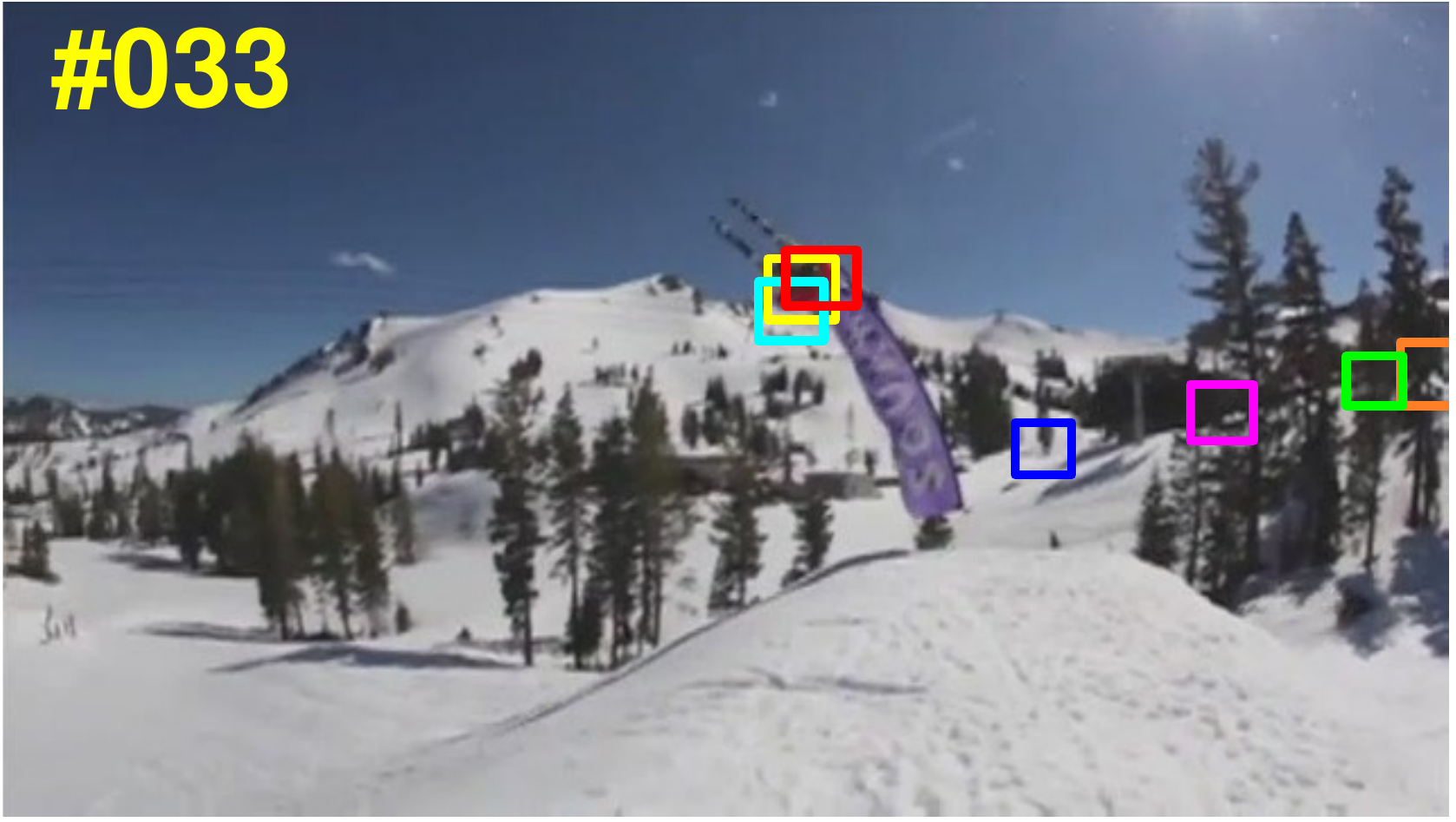} &
\includegraphics[trim = 2mm 2mm 5mm 1mm, clip,width=.16\textwidth]{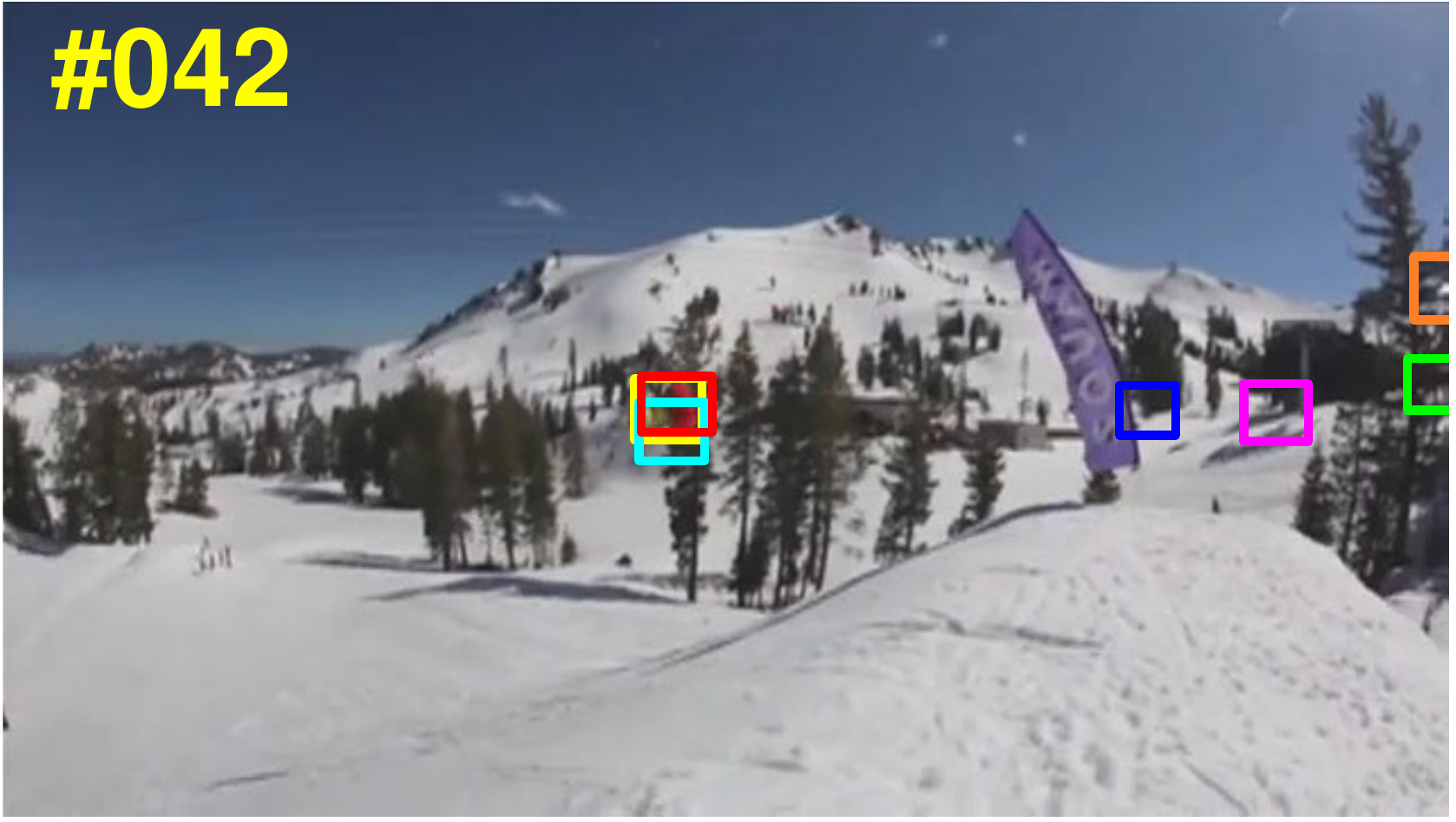} &
\includegraphics[trim = 2mm 2mm 5mm 1mm, clip,width=.16\textwidth]{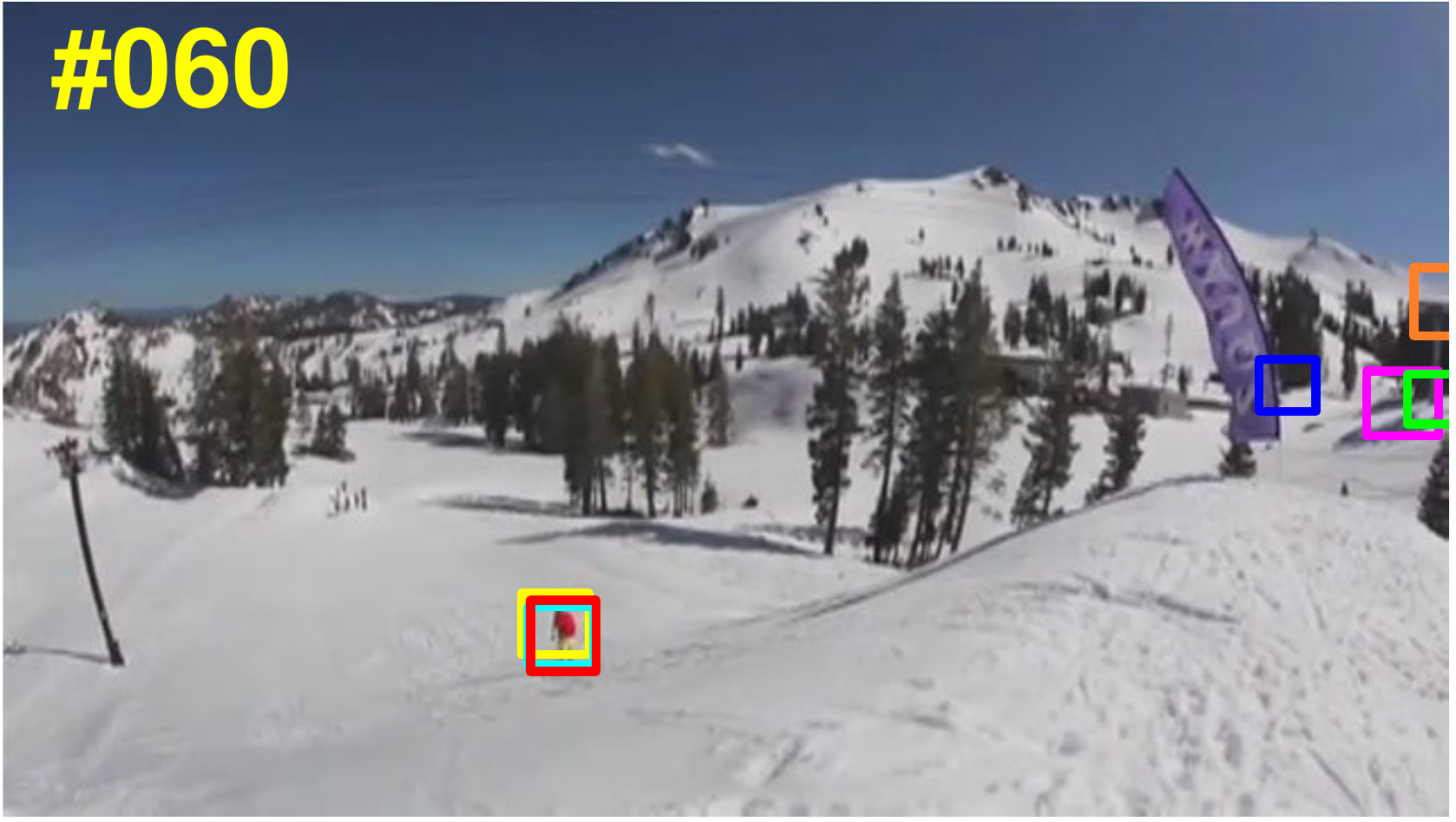} & 
\includegraphics[trim = 2mm 2mm 5mm 1mm, clip,width=.16\textwidth]{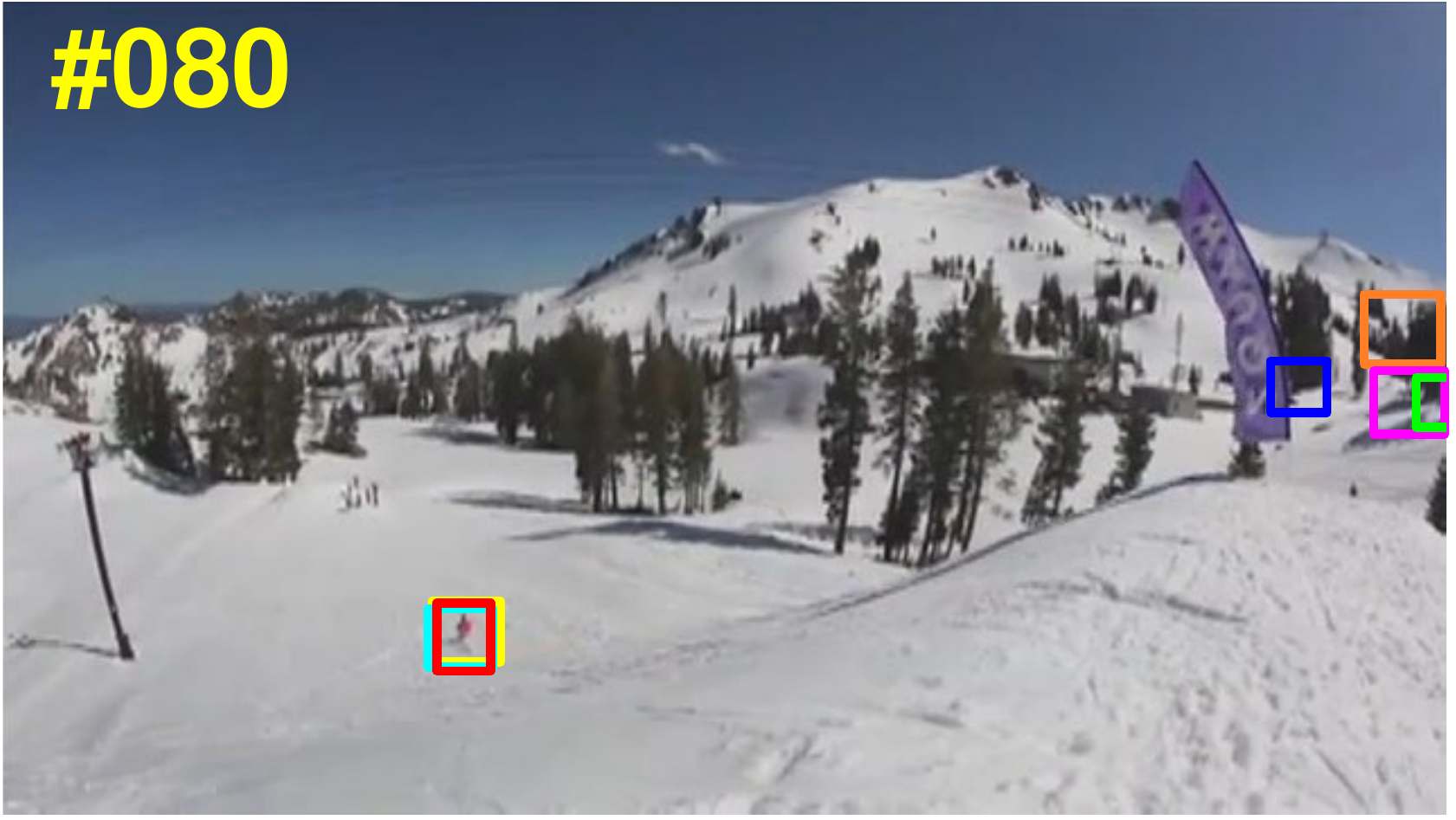} \\
\includegraphics[trim = 2mm 2mm 5mm 1mm, clip,width=.16\textwidth]{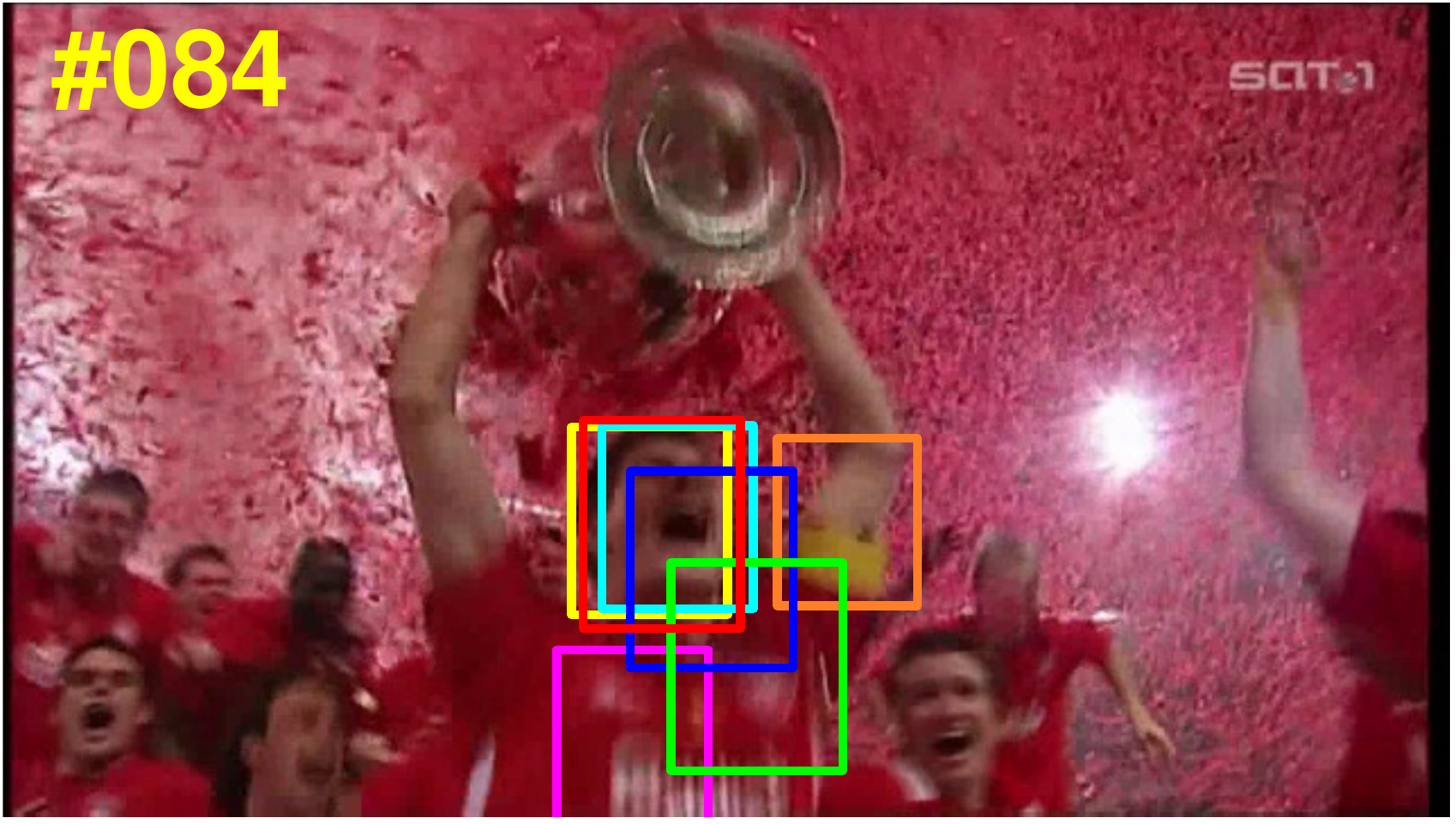}& 
\includegraphics[trim = 2mm 2mm 5mm 1mm, clip,width=.16\textwidth]{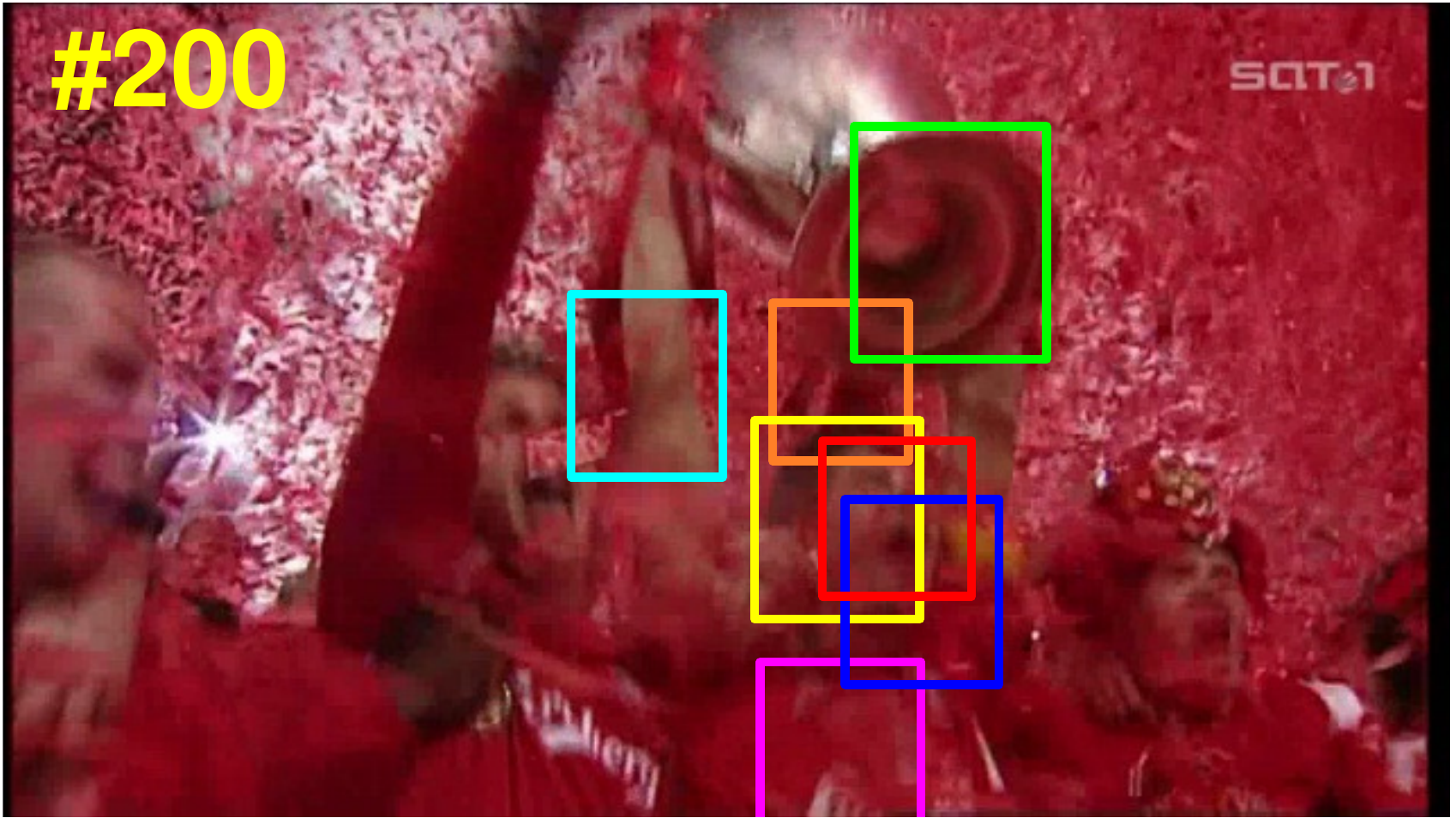}& 
\includegraphics[trim = 2mm 2mm 5mm 1mm, clip,width=.16\textwidth]{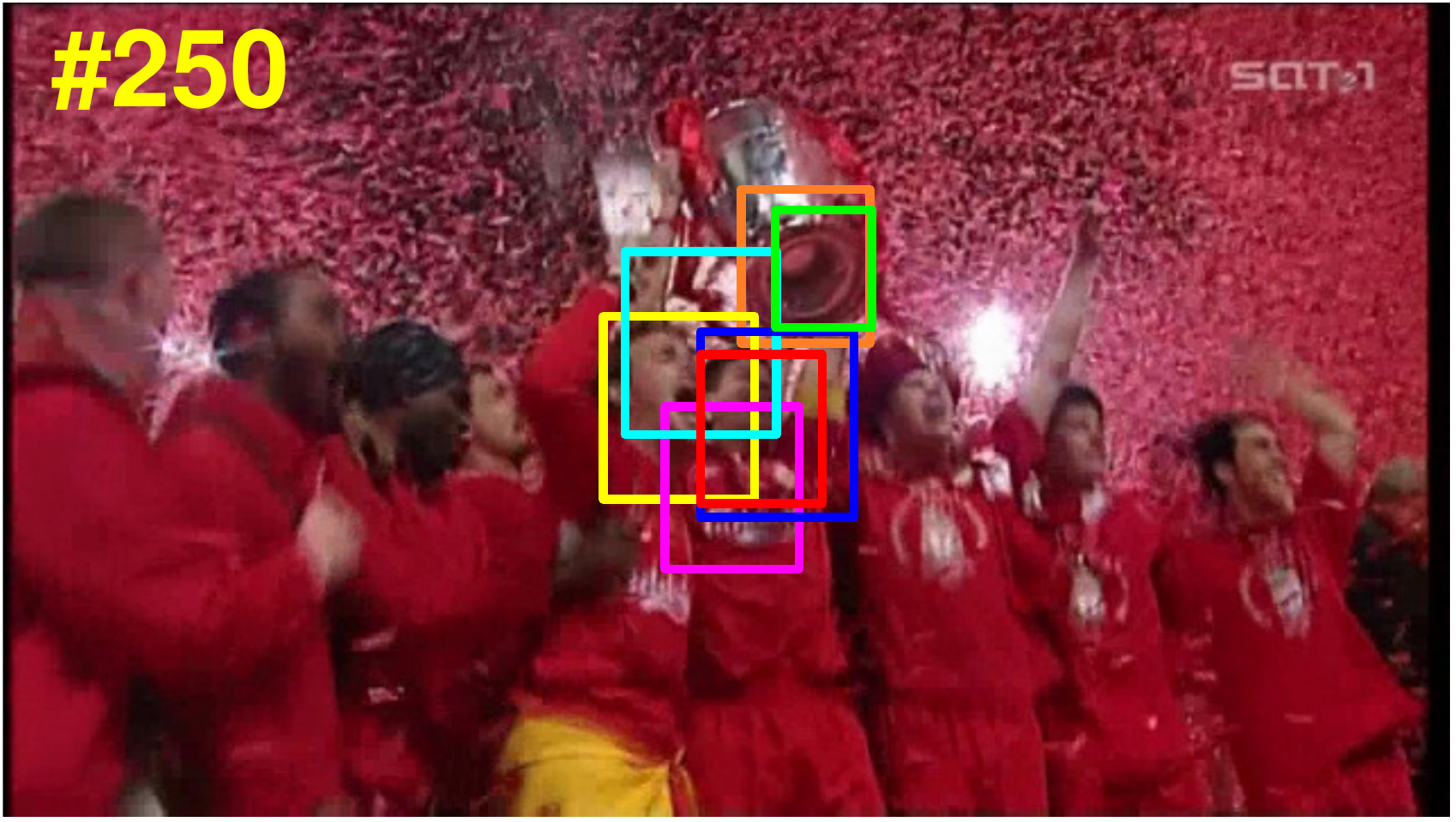} &
\includegraphics[trim = 2mm 2mm 5mm 1mm, clip,width=.16\textwidth]{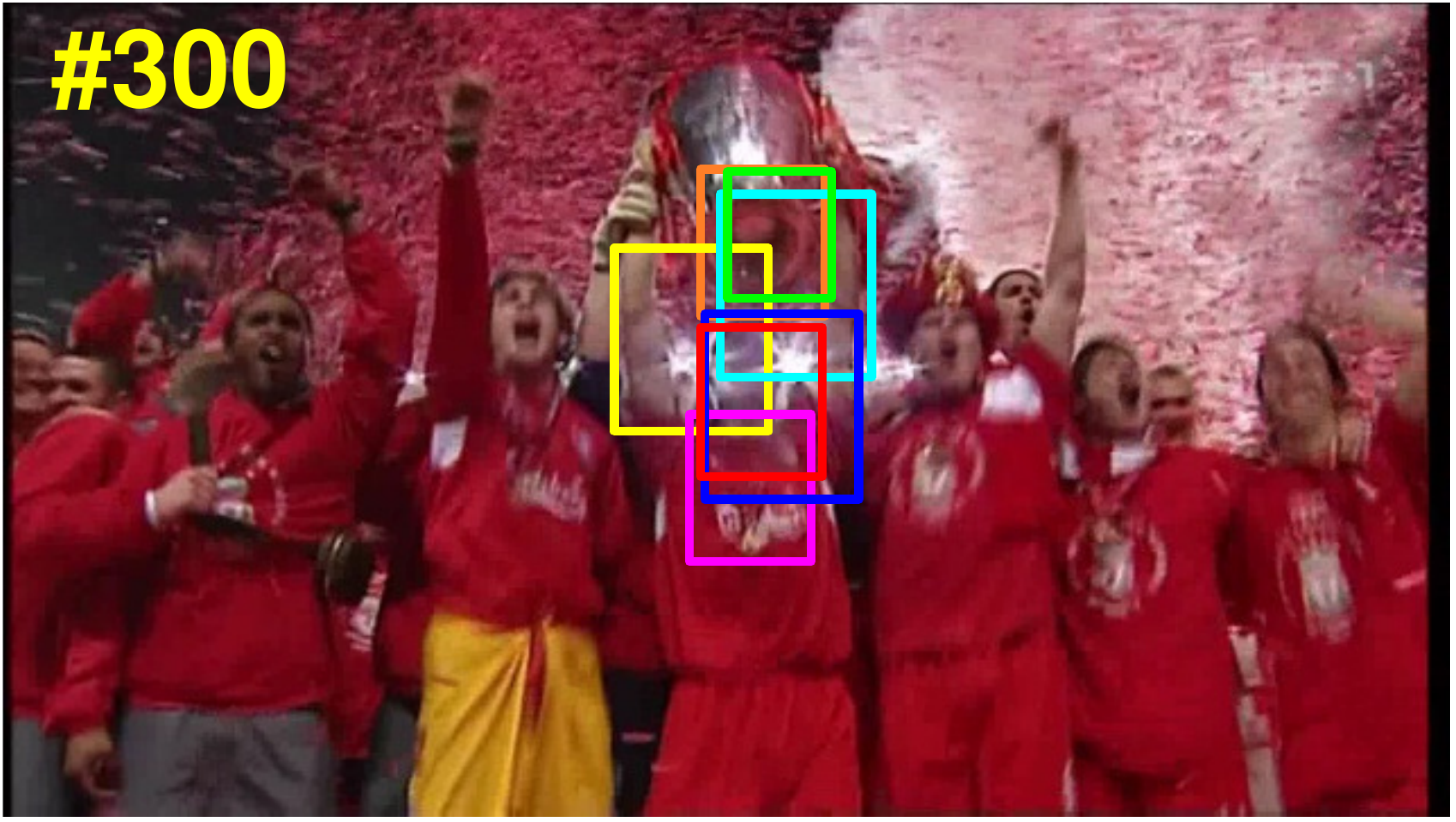}& 
\includegraphics[trim = 2mm 2mm 5mm 1mm, clip,width=.16\textwidth]{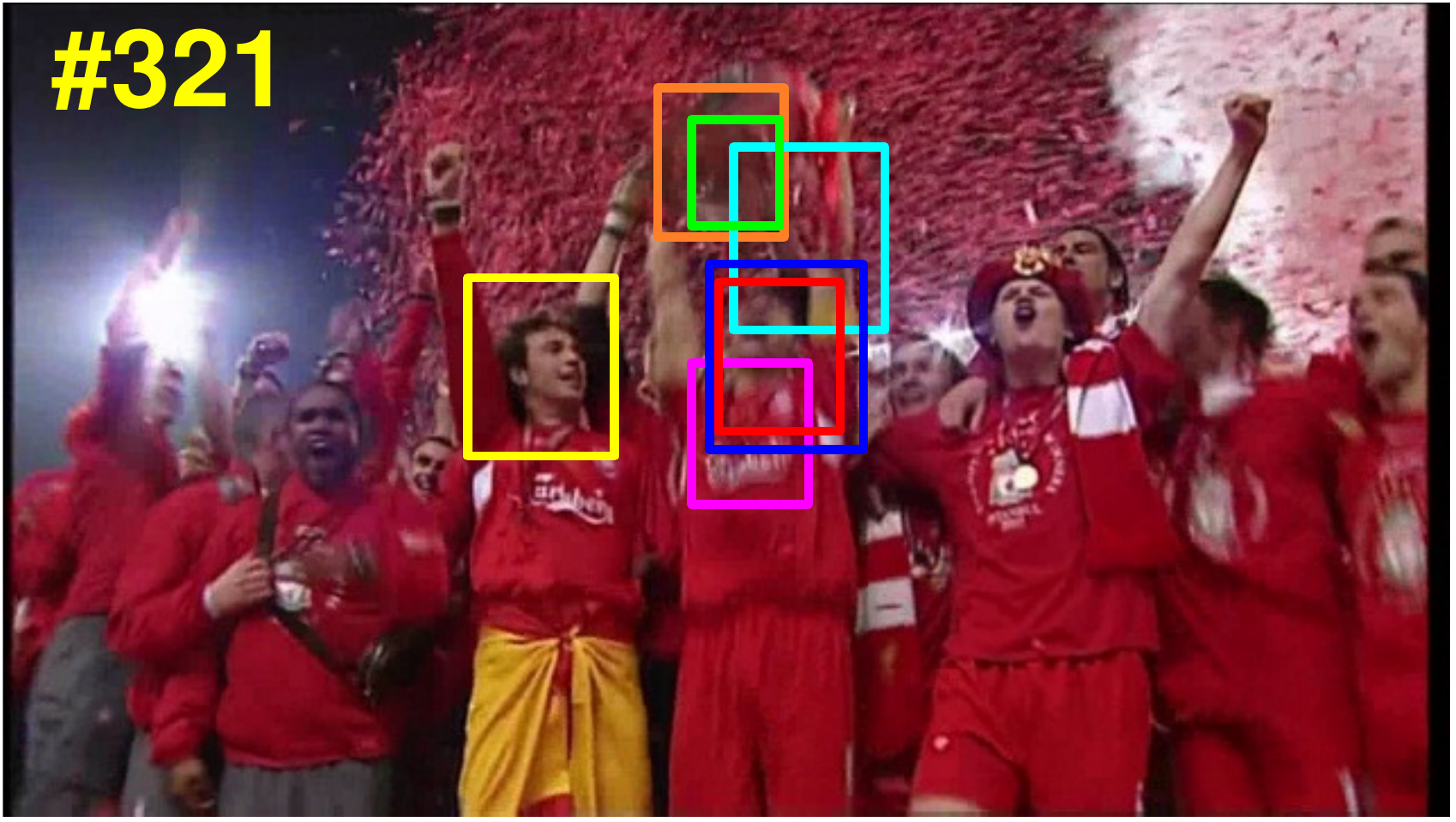} &
\includegraphics[trim = 2mm 2mm 5mm 1mm, clip,width=.16\textwidth]{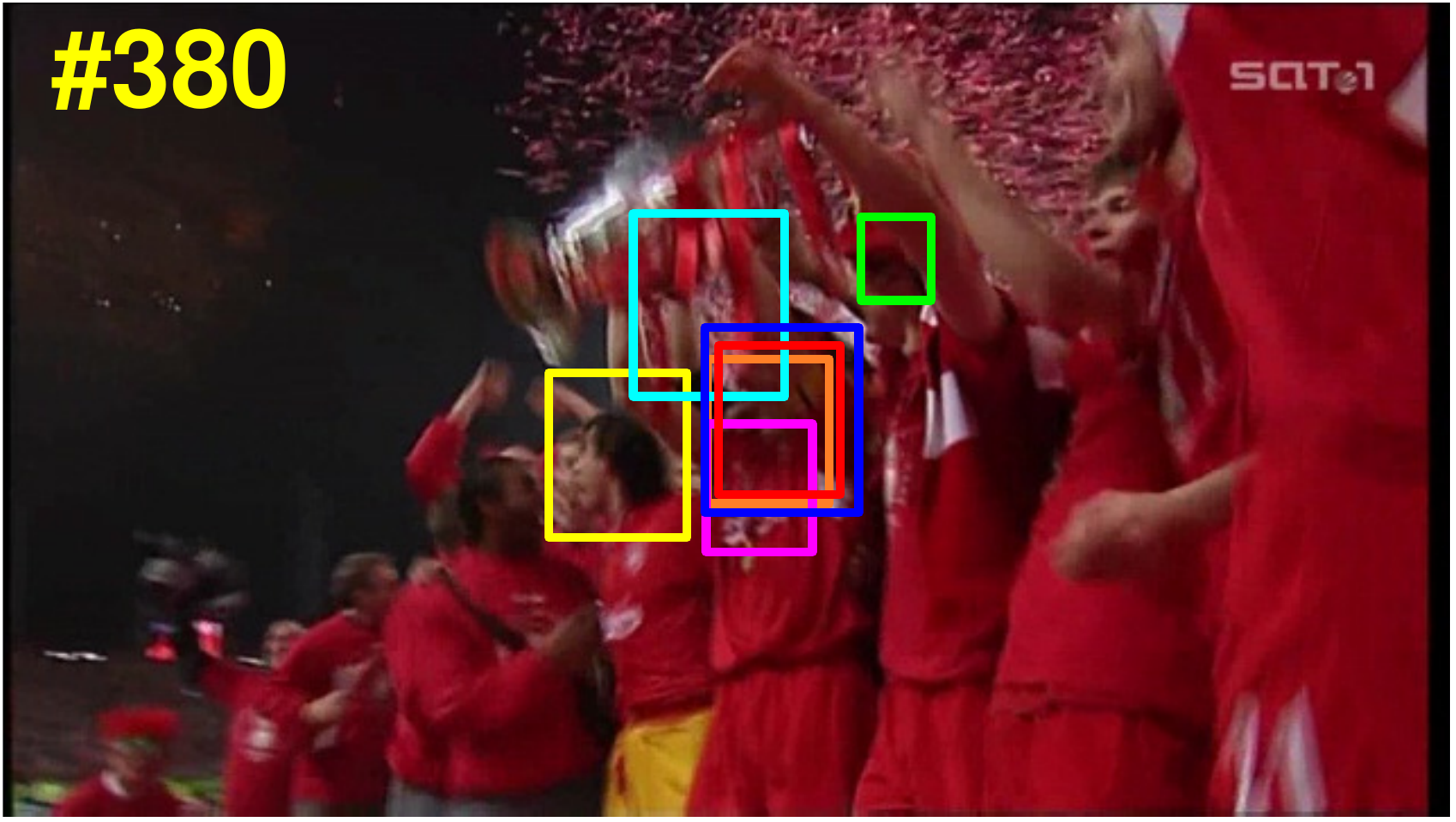} \\
\end{tabular}
\includegraphics[width=.94\textwidth]{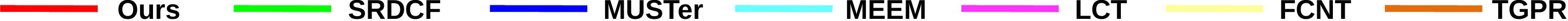} 
\vspace{-2mm}
\caption{\tb{Sample tracking results.} We show some tracking results of the SRDCF~\cite{DBLP:conf/iccv/DanelljanHKF15},
MUSTer~\cite{Hong_2015_CVPR},
MEEM~\cite{DBLP:conf/eccv/ZhangMS14},
LCT~\cite{DBLP:conf/cvpr/MaYZY15}
FCNT~\cite{DBLP:conf/iccv/WangOWL15}
TGPR~\cite{DBLP:conf/eccv/GaoLHX14} methods, 
and the proposed algorithm
on four most challenging sequences 
(From top: \textit{Matrix}, \textit{MotorRolling},
\textit{Skiing}, and \textit{Soccer}). 
%Our tracker performs well against the state-of-the-art methods.
}
\label{fig:result}
\vspace{-2mm}
\end{figure*}

\subsubsection{Classifier Analysis}
The long-term memory correlation filter plays a critical role in scoring the confidence of 
each tracked result.
Taking the \textit{Lemming} sequence as an example, we first analyze the stability of the correlation filter classifier. 
Figure~\ref{fig:lemming} shows two classifiers with different incremental learning schemes to determine the confidence of every tracked result of the HCFT~\cite{DBLP:conf/iccv/MaHYY15} (bottom left). 
Without a re-detection module, the HCFT loses the target after the 360th frame. 
A correlation filter $g(\cdot)$ may easily adapt to 
noisy appearance changes and fail to predict tracking failures after the 360th frame. 
In contrast, the conservatively learned filter $g(\cdot)$ is robust to noisy updates and succeeds in estimating the confidence of every tracked result. 
For the HCFT* with a re-detection module, we analyze the model stability using different thresholds to determine tracking failures.
Overall, the performance of the HCFT* 
is not sensitive when the threshold value is between 0.2 to 0.5.
In this work, we set the threshold $T_0$ to 0.2 for all experiments.

\begin{table}[t]
\setlength{\tabcolsep}{.6em}
\caption{\tb{Average ranks of accuracy and robustness under baseline and region noise experiments on the VOT2014 dataset. }The first, second and 
third best scores are highlighted in red, blue and green.}
\vspace{-4mm}
\label{tb:vot2014}
\scriptsize
\begin{center}
\begin{tabular}{lccc}
\hline
& {baseline} & {region\_noise} & \\
& {Expected overlap} & {Expected overlap} & \ \ { \ \ Overall \ \ } \ \ \\\hline
{HCFT*} & \first{0.3532} & \first{0.2798} & \first{0.3165} \\
{HCFT} & \second{0.3242} & \third{0.2695} & \second{0.2968} \\
% {MCT} & \third{0.3087} & 0.2653 & \third{0.2870} \\
{DSST} & \third{0.2990} & 0.2630 & \third{0.2810} \\
{DGT} & 0.2770 & 0.2624 & 0.2697 \\
{BDF} & 0.2826 & 0.2528 & 0.2677 \\
{HMMTxD} & 0.2613 & \second{0.2702} & 0.2657 \\
{VTDMG} & 0.2687 & 0.2562 & 0.2625 \\
{MatFlow} & 0.2936 & 0.2308 & 0.2622 \\
{ABS} & 0.2615 & 0.2565 & 0.2590 \\
{SAMF} & 0.2741 & 0.2343 & 0.2542 \\
{KCF} & 0.2734 & 0.2336 & 0.2535 \\
{ACAT} & 0.2401 & 0.2243 & 0.2322 \\
{DynMS} & 0.2293 & 0.2143 & 0.2218 \\
{ACT} & 0.2346 & 0.1915 & 0.2131 \\
{PTp} & 0.2161 & 0.2047 & 0.2104 \\
{EDFT} & 0.2208 & 0.1965 & 0.2086 \\
{IPRT} & 0.2078 & 0.1856 & 0.1967 \\
{LT\_FLO} & 0.2035 & 0.1814 & 0.1925 \\
{SIR\_PF} & 0.1943 & 0.1872 & 0.1908 \\
{FoT} & 0.1935 & 0.1622 & 0.1779 \\
{Struck} & 0.1858 & 0.1670 & 0.1764 \\
%{ThunderStruck} & 0.1775 & 0.1748 & 0.1762 \\
{FSDT} & 0.1553 & 0.1544 & 0.1549 \\
{CMT} & 0.1577 & 0.1472 & 0.1524 \\
{FRT} & 0.1590 & 0.1434 & 0.1512 \\
{IVT} & 0.1527 & 0.1452 & 0.1489 \\
{OGT} & 0.1470 & 0.1307 & 0.1389 \\
{MIL} & 0.1579 & 0.1112 & 0.1346 \\
%{IIVTv2} & 0.1380 & 0.1309 & 0.1345 \\
%{Matrioska} & 0.1389 & 0.1252 & 0.1321 \\
{CT} & 0.1262 & 0.1173 & 0.1218 \\
%{IMPNCC} & 0.1097 & 0.1034 & 0.1065 \\
{NCC} & 0.0790 & 0.0749 & 0.0770 \\\hline
\end{tabular}
\end{center}
\vspace{-5mm}
\end{table}

\subsubsection{Qualitative Evaluation}
Figure~\ref{fig:result} shows qualitative comparisons with the top performing tracking methods:
SRDCF~\cite{DBLP:conf/iccv/DanelljanHKF15},
MUSTer~\cite{Hong_2015_CVPR},
MEEM~\cite{DBLP:conf/eccv/ZhangMS14},
LCT~\cite{DBLP:conf/cvpr/MaYZY15},
FCNT~\cite{DBLP:conf/iccv/WangOWL15},
TGPR~\cite{DBLP:conf/eccv/GaoLHX14} and the proposed algorithm on four most challenging sequences from the OTB2015 dataset.
The FCNT also exploits the deep feature hierarchy and performs well in sequences with deformation and rotation (\textit{MotorRolling} and \textit{Skiing}), but fails when the background is cluttered and fast motion occurs (\textit{Matrix} and \textit{Soccer}) as the underlying random sampling scheme to infer target states is not robust to cluttered backgrounds.
The MEEM tracker uses quantized color features and performs well in the \textit{Skiing} sequence. 
The SRDCF, MUSTer and LCT 
all build on adaptive correlation filters with hand-crafted features, which are not robust to large appearance changes. 
These trackers drift quickly at the beginning of the \textit{MotorRolling}, \textit{Matrix} and \textit{Soccer} sequences even if with a re-detection module (MSUTer and LCT). 
The reasons that the proposed algorithm performs well are three-fold: 
First, the visual representation using hierarchical convolutional features learned from a large-scale dataset are more effective than the conventional hand-crafted features.
With CNN features from multiple levels, 
the representation scheme contains both category-level semantics and fine-grained spatial details which account for appearance changes caused by deformations, rotations and background clutters (\textit{Matrix}, \textit{Skiing}, and \textit{Soccer}).
It is worth mentioning that for the most challenging~\textit{MotorRolling} sequence, none of the other methods are able to track targets well whereas the HCFT* achieves the distance precision rate of 94.5\%. 
Second, the linear correlation filters trained on convolutional features are updated properly to account for appearance 
variations. 
Third, two types of designed region proposals help to recover lost target objects as well as estimate scale changes.

%MH: use smaller fonts for legends, e.g., thunderstruck overlaps with VTDMG
%CM: updated
\begin{figure}[t]
\setlength{\tabcolsep}{2pt}
\begin{tabular}{cc}
\includegraphics[ width=.23\textwidth]{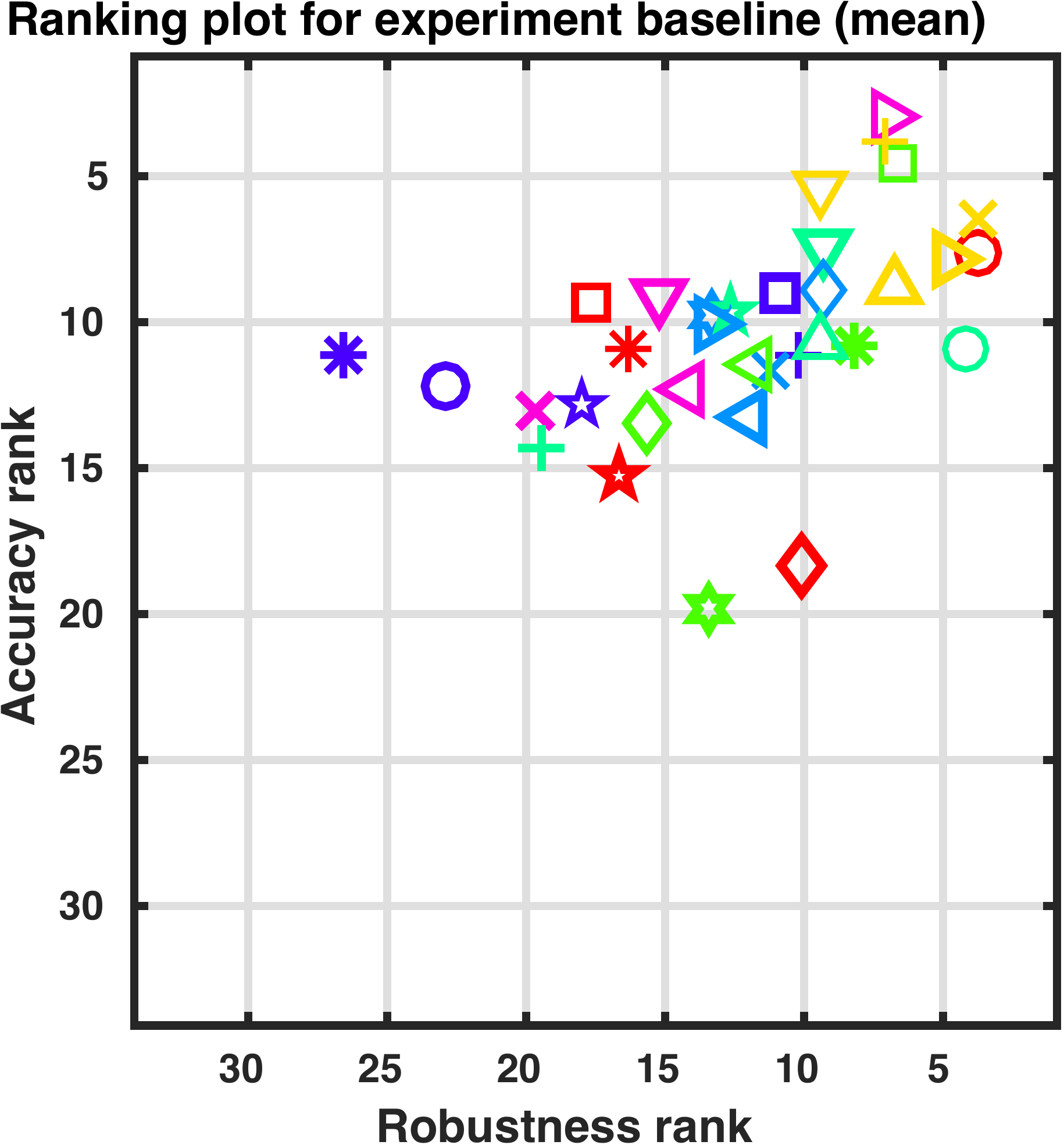} &
\includegraphics[width=.24\textwidth]{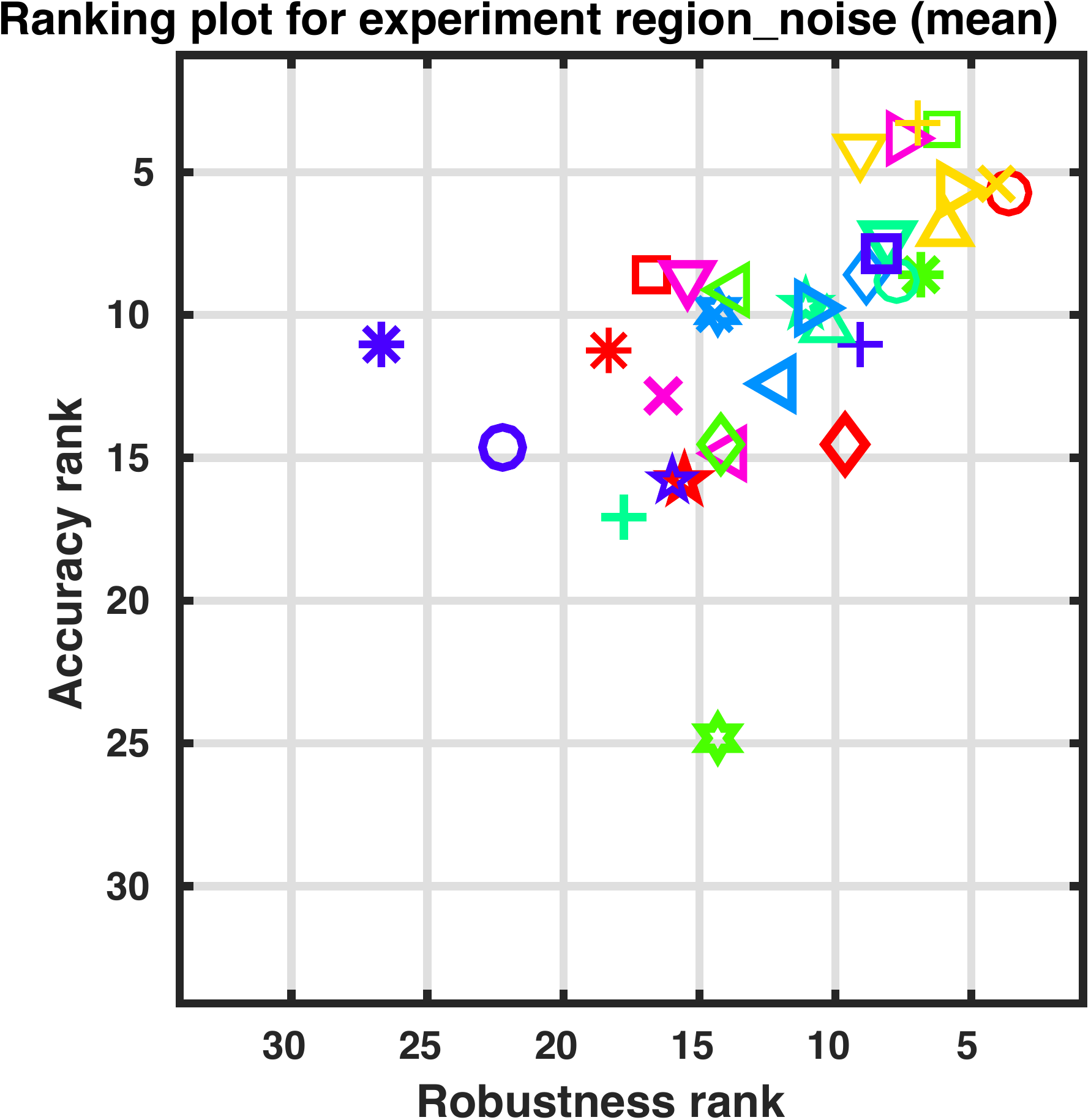} \\ 
\end{tabular}
\includegraphics[width=.48\textwidth]{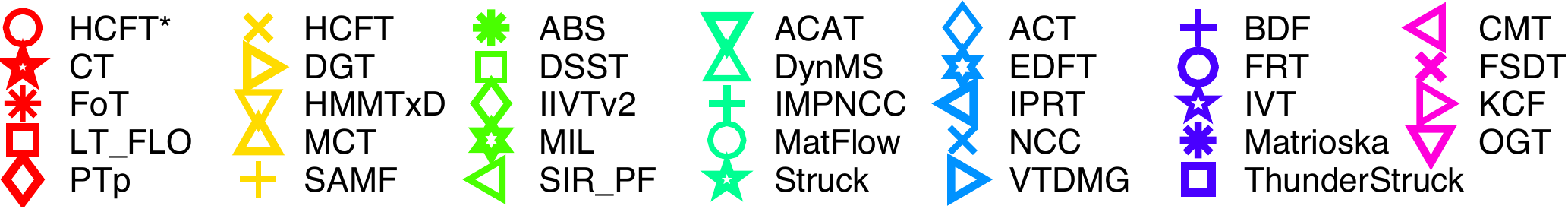} 
\vspace{-2mm}
\caption{\tb{Robustness-accuracy ranking plots under the baseline and region noise experiments on the VOT2014~\cite{Hadfield14d} dataset.} Trackers closer to upper right corner perform better.}
\label{fig:vot2014}
\vspace{-2mm}
\end{figure}

\begin{table}[t]
\centering
\caption{\tb{Results on VOT2015.} Average ranks of accuracy and robustness under baseline experiments on the VOT2015~\cite{DBLP:conf/iccvw/KristanMLFCFVHN15} dataset. The first, second and 
third best scores are highlighted in red, blue and green colors, respectively}
\label{tb:vot2015}
\vspace{-2mm}
\scriptsize
\setlength{\tabcolsep}{.8em}
\begin{tabular}{lccc}
\hline
& ~ ~ Accuracy ~ ~ & Robustness & Expected Overlap \\ \hline
MDNet & \first{0.5939} & \first{0.7656} & \first{0.3783} \\
DeepSRDCF & 0.5621 & \third{1.0000} & \second{0.3181} \\
{\bf HCFT*} & 0.4843 & \second{0.9844} & \third{0.2878} \\
SRDCF & 0.5506 & 1.1833 & 0.2877 \\
LDP & 0.4932 & 1.3000 & 0.2785 \\
%sPST & 0.5397 & 1.4167 & 0.2767 \\
SCEBT & 0.5379 & 1.7189 & 0.2548 \\
% nsamf & 0.5300 & 1.4500 & 0.2536 \\
Struck & 0.4597 & 1.4956 & 0.2458 \\
%Rajssc & \third{0.5674} & 1.7500 & 0.2420 \\
% s3tracker & 0.5244 & 1.6667 & 0.2403 \\
SumShift & 0.5115 & 1.6167 & 0.2341 \\
SODLT & 0.5573 & 1.8067 & 0.2329 \\
DAT & 0.4804 & 1.8833 & 0.2238 \\
MEEM & 0.4994 & 1.7833 & 0.2212 \\
%RobStruck & 0.4896 & 1.5833 & 0.2198 \\
OACF & \second{0.5676} & 1.8833 & 0.2190 \\
MCT & 0.4677 & 1.7378 & 0.2188 \\
%HMMTxD & 0.5261 & 2.2500 & 0.2185 \\
ASMS & 0.5011 & 1.7833 & 0.2117 \\
% mkcf\_plus & 0.5189 & 1.9667 & 0.2095 \\
% tric & 0.4517 & 1.9878 & 0.2088 \\
AOGTracker & 0.5109 & 1.7400 & 0.2080 \\
% sme & 0.5475 & 2.1000 & 0.2068 \\
% Mvcft & 0.5095 & 1.9333 & 0.2059 \\
% srat & 0.4520 & 1.9833 & 0.2031 \\
% dtracker & 0.4929 & 1.9833 & 0.2022 \\
SAMF & 0.5115 & 2.0833 & 0.2021 \\
% ggt & 0.4553 & 1.9356 & 0.1958 \\
MUSTer & 0.5116 & 2.2044 & 0.1950 \\
TGPR & 0.4636 & 2.2433 & 0.1938 \\
% baseline & 0.4766 & 2.2756 & 0.1935 \\
% kcfv2 & 0.4703 & 2.0333 & 0.1927 \\
% cmil & 0.4539 & 2.1333 & 0.1890 \\
ACT & 0.4579 & 2.2167 & 0.1861 \\
% kcf\_mtsa & 0.4792 & 2.4500 & 0.1751 \\
LGT & 0.4003 & 2.0511 & 0.1737 \\
DSST & 0.5283 & 2.7167 & 0.1719 \\
MIL & 0.4161 & 2.6078 & 0.1710 \\
%KCF2 & 0.4741 & 2.4333 & 0.1707 \\
%sKCF & 0.4642 & 2.7500 & 0.1620 \\
% bdf & 0.3802 & 2.8333 & 0.1526 \\
% kcfdp & 0.4801 & 2.5333 & 0.1526 \\
PKLTF & 0.4458 & 2.6167 & 0.1524 \\
HT & 0.4236 & 2.8111 & 0.1515 \\
FCT & 0.4170 & 2.9500 & 0.1512 \\
Matflow & 0.4037 & 2.8167 & 0.1500 \\
SCBT & 0.4059 & 2.6367 & 0.1496 \\
DFT & 0.4527 & 3.8167 & 0.1395 \\
%FoT & 0.4200 & 3.5833 & 0.1385 \\
%LT\_FLO & 0.4171 & 3.9389 & 0.1328 \\
L1APG & 0.4607 & 4.0333 & 0.1270 \\
OAB & 0.4243 & 3.9167 & 0.1259 \\
IVT & 0.4218 & 4.1000 & 0.1220 \\
STC & 0.3801 & 3.7333 & 0.1179 \\
CMT & 0.3795 & 3.9000 & 0.1153 \\
CT & 0.3742 & 3.5589 & 0.1135 \\
FragTrack & 0.3942 & 4.6833 & 0.1081 \\
% zhang & 0.3189 & 3.7333 & 0.0997 \\
%Loft\_lite & 0.3459 & 5.3667 & 0.0807 \\
NCC & 0.4818 & 8.1833 & 0.0795 \\
% amt & 0.1950 & 6.1000 & 0.0349 \\ 
\hline
\end{tabular}
\vspace{-4mm}
\end{table}

\subsection{VOT2014 Dataset}

The VOT2014~\cite{Hadfield14d} dataset contains 25 real-world video sequences. 
We conduct two sets of experiments based on the metrics~\cite{Hadfield14d}: 
(1) the baseline evaluation in which trackers are initialized with ground truth locations; and (2) the noise evaluation in which the initial target locations are perturbed with random noise. 
The VOT challenge provides a re-initialization protocol, where trackers are reset with ground-truths when tracking failures occur. 

We use two metrics to rank all the trackers: 
(1) accuracy measure based on 
the overlap ratio with ground truth bounding box and 
(2) robustness measure based on the number of tracking failures.
We validate the proposed HCFT* in comparison to all the trackers submitted to the VOT2014 challenge~\cite{Hadfield14d}. 
Table~\ref{tb:vot2014} presents the average accuracy and robustness rank of all 
evaluated trackers. 
In both the baseline and region noise validations, the proposed HCFT* 
performs well in terms of accuracy and robustness scores. 
Note that the DSST~\cite{DBLP:conf/bmvc/DanelljanKFW14} learns adaptive correlation filters over HOG features for estimating the translation and scale changes. 
In contrast, the HCFT* learns correlation filter over hierarchical deep CNN features, which provide multi-level semantical abstractions of target objects for robust tracking. 
Figure~\ref{fig:vot2014} shows the accuracy-robustness plots on the VOT2014 dataset. 
Our method performs favorably among all the other trackers in terms of both accuracy and robustness (see the upper right corner).

\subsection{VOT2015 Dataset}
We evaluate the proposed algorithm on the VOT2015~\cite{DBLP:conf/iccvw/KristanMLFCFVHN15} dataset, which contains 60 challenging video sequences with sufficient variations. 
We use accuracy and robustness as the performance metrics.
Table~\ref{tb:vot2015} presents the average accuracy and robustness rank 
of all evaluated trackers including those submitted to the VOT2015. 
%
%We refer readers to~\cite{DBLP:conf/iccvw/KristanMLFCFVHN15} for more details regarding the compared trackers. 
%
The proposed HCFT* ranks second and third in terms of robustness and overall overlap, respectively. 
Note that MDNet~\cite{DBLP:conf/cvpr/NamH16} 
draws multiple binary samples for learning CNNs. 
With the use of a negative-mining scheme to alleviate sampling ambiguity as well as 
the domain transfer learning scheme to learn from more training data offline, 
the MDNet method performs favorably among all trackers. 
We note the proposed HCFT* is computationally more efficient than the MDNet 
method in terms of tracking speed: 6.7 FPS (HCFT*) vs. 1.2 FPS (MDNet).

\begin{figure}[t]
\centering
\setlength{\tabcolsep}{.1em}
\begin{tabular}{ccc}
\includegraphics[trim = 0mm 0mm 0mm 0mm, clip, height=.1\textwidth]{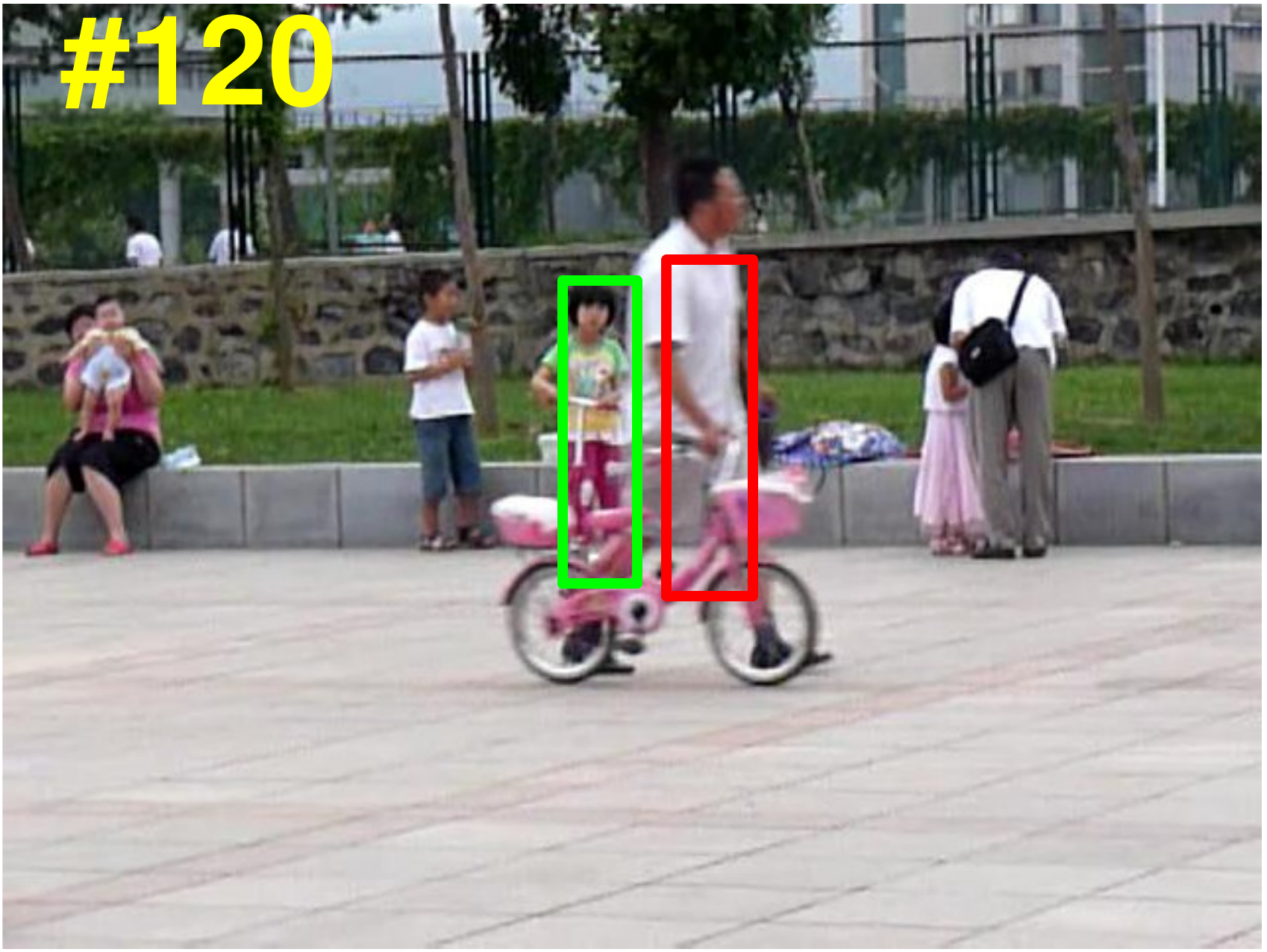} &
\includegraphics[trim = 0mm 0mm 10mm 1mm, clip, height=.1\textwidth]{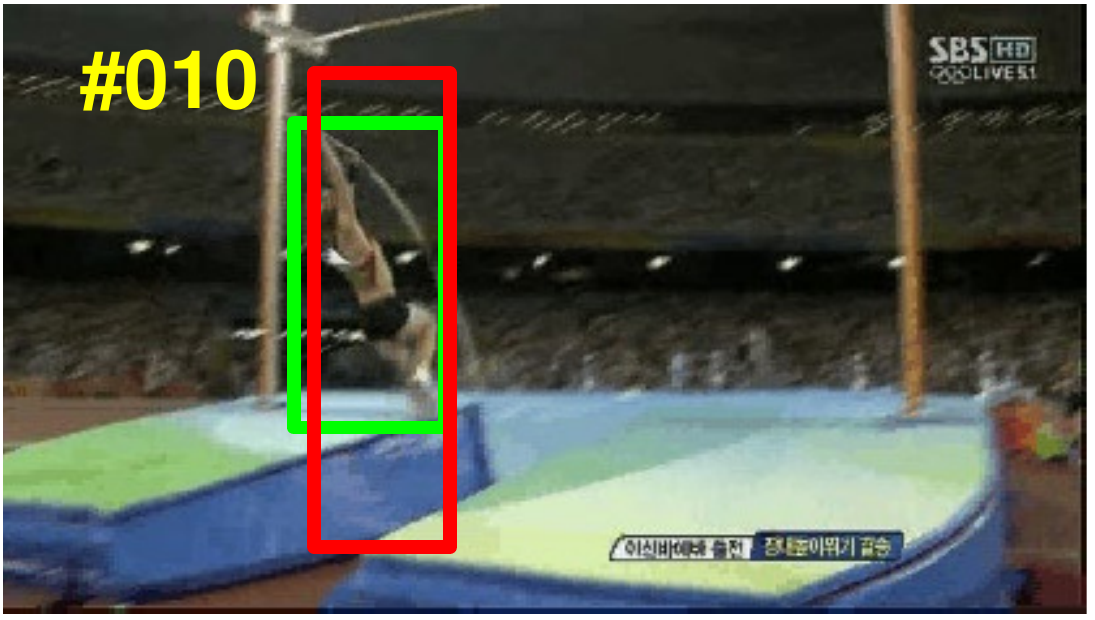} &
\includegraphics[trim = 0mm 0mm 10mm 0mm, clip, height=.1\textwidth]{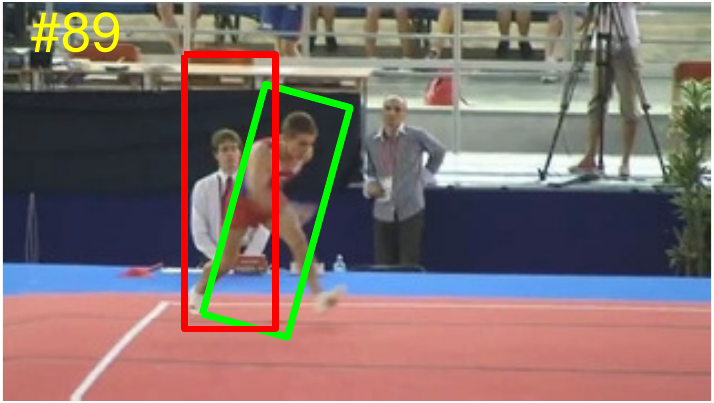} 
\end{tabular}
\caption{\tb{Failure cases on the \textit{Girl2} (OTB2013~\cite{DBLP:conf/cvpr/WuLY13}), \textit{Jump} (OTB2015~\cite{DBLP:journals/pami/WuLY15}) and \textit{Gymnastics} (VOT2015~\cite{DBLP:conf/iccvw/KristanMLFCFVHN15}) sequences.} Red boxes show our results and the green ones are ground truth.
}
\label{fig:failure}
%\vspace{-4mm}
\end{figure}

\subsection{Failure Cases}

We show sample tracking failures by the HCFT* in Figure~\ref{fig:failure}. 
For the \textit{Girl2} sequence, when long-term occlusions occur, 
the proposed re-detection scheme is 
not successfully activated due to the high similarity between the target person and surrounding 
people. 
In the \textit{Jump} sequence, we use the EdgeBox~\cite{DBLP:conf/eccv/ZitnickD14} method 
with a small step size to generate scale proposals tightly around the estimated target position. 
This approach does not perform well in the presence of drastic scale changes. 
For the \textit{Gymnastics} sequence, our approach does not 
generate the rotated bounding boxes as the annotated ground truth labels.
%
%In fact, we use unrotated minimum exterior rectangles as input, which contains a large body of noisy background in the initial step. 
As the proposed HCFT* does not consider in-plane rotation, pixels in the background regions are considered as parts of the foreground object, thereby causing inaccurate model updates 
and drifts. 
%This adversely affects our approach in precisely localizing the target.

\section{Conclusion}
In this paper, we exploit rich feature hierarchies of CNNs for visual tracking.
We use the semantic information within the last convolutional layers for handling significant appearance variations and spatial details within early convolutional layers 
for locating target objects. 
Specifically, we train a linear correlation filter on each convolutional layer and infer the target position from hierarchical correlation maps in a coarse-to-fine manner.
We further address the challenging problems of scale estimation and target re-detection.
We conservatively learn another correlation filter with a long-term memory of target appearance on two type of region proposal generated by the EdgeBox method, and respectively select the proposals with highest confidence scores as estimated scale and candidate re-detection results. 
We extensively validate the proposed algorithm on the OTB2013, OTB2015, VOT2014, and VOT2015 datasets. 
Experimental results show that the proposed algorithm performs favorably against the state-of-the-art methods in terms of accuracy and robustness.

We summarize the potential directions to improve our approach and shed light on our future work. 
First, since correlation filters regress the circularly shifted versions of deep features to soft labels, the performance of our approach is sensitive to the initial center location in the first frame. 
On the VOT2014 and VOT2015 datasets, we use the \emph{unrotated} minimum 
enclosing rectangles as ground truth, which may contain a large body of noisy background. 
For some sequences, the geometric centers of ground truth are far from the actual center positions of target objects. 
Our approach on the VOT2014 and VOT2015 datasets do not perform as well as on the OTB2013 and OTB2015 datasets. 
This suggests our future work using a better way to define 
search windows for learning correlation filters.
Second, we hierarchically infer the target location with empirical weights. 
Although the soft weight scheme used in this paper shows advantages when compared to the hard weight scheme used for our preliminary results~\cite{DBLP:conf/iccv/MaHYY15}, we do not have principled interpretations and explanations. 
Our future work will emphasize a learning based approach to avoid manually setting these empirical weights.
Third, our approach relies on several free parameters such as thresholds to activate the re-detection module as well as to accept/reject the detection results.
As these fixed parameters cannot be applied to all sequences effectively, 
it is of great interest to develop adaptive re-detection modules for recovering target objects 
from tracking failures.

%experimental results show that the proposed algorithm performs favorably against the state-of-the-art methods in terms of accuracy and robustness. 

%% use section* for acknowledgment
%\ifCLASSOPTIONcompsoc
% % The Computer Society usually uses the plural form
% \section*{Acknowledgments}
%\else
% % regular IEEE prefers the singular form
% \section*{Acknowledgment}
%\fi
%
%C. Ma and X. Yang are supported in part by NSFC Grants (61527804, 61025005, 61129001, and 61221001), STCSM Grants (14XD1402100, and 13511504501), 111 Program Grant (B07022). M.-H. Yang is supported in part by NSF CAREER Grant (1149783) and NSF IIS Grant (1152576).
%% Can use something like this to put references on a page
%% by themselves when using endfloat and the captionsoff option.
%\ifCLASSOPTIONcaptionsoff
% \newpage
%\fi

\section*{Acknowledgments}
This work is supported in part by the National Key Research and Development Program of China (2016YFB1001003), NSFC (61527804, 61521062) and the 111 Program (B07022).

\vspace{-2mm}
\bibliographystyle{IEEEtran}
\bibliography{bib_short_title,bib_all}
\vspace{-15mm}
\begin{IEEEbiography}[{\includegraphics[width=1in,height=1.25in,clip,keepaspectratio]{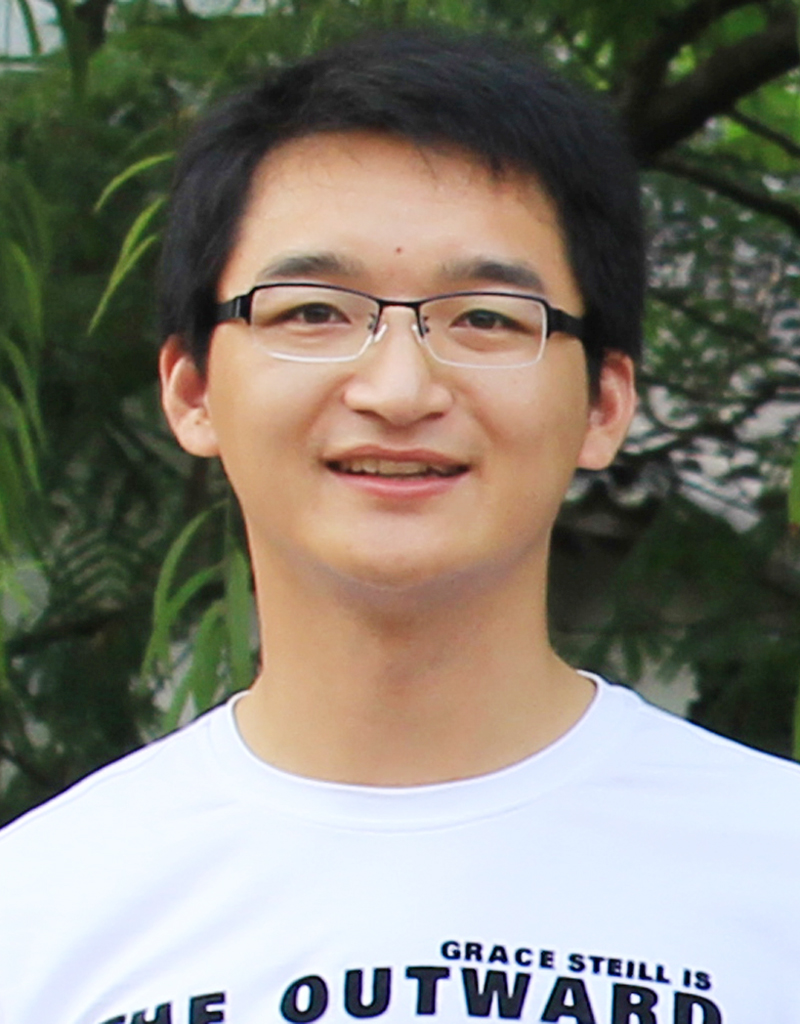}}]{Chao Ma}
is a senior research associate with School of Computer Science at The University of Adelaide. He received the Ph.D. degree from Shanghai Jiao Tong University in 2016. His research interests include computer vision and machine learning. He was sponsored by China Scholarship Council as a visiting Ph.D. student in University of California at Merced from the fall of 2013 to the fall of 2015. %He is a member of the IEEE.
\end{IEEEbiography}
\vspace{-15mm}
\begin{IEEEbiography}[{\includegraphics[width=1in,height=1.25in,clip,keepaspectratio]{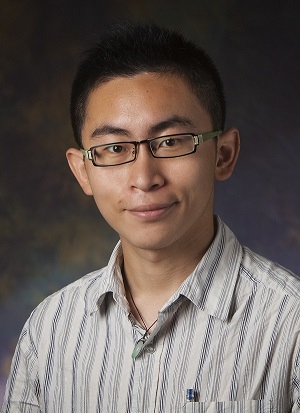}}]{Jia-Bin Huang} is an assistant professor in the Bradley Department of Electrical and Computer Engineering at Virginia Tech. He received the B.S. degree in Electronics Engineering from National Chiao-Tung University, Hsinchu, Taiwan and his Ph.D. degree in the Department of Electrical and Computer Engineering at University of Illinois, Urbana-Champaign in 2016. He is a member of the IEEE.
\end{IEEEbiography}
\vspace{-15mm}
\begin{IEEEbiography}[{\includegraphics[width=1in,height=1.25in,clip,keepaspectratio]{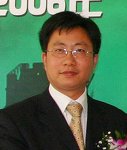}}]{Xiaokang Yang} received the B.S. degree from Xiamen University, Xiamen, China, in 1994; the M.S. degree from Chinese Academy of Sciences, Shanghai, China, in 1997; and the Ph.D. degree from Shanghai Jiao Tong University, Shanghai, in 2000.
He was a Research Fellow with the Center for Signal Processing, Nanyang Technological University, Singapore, from 2000 to 2002 and a Research Scientist with the Institute for Infocomm Research, Singapore, from 2002 to 2004. He is currently a full professor and the deputy director of the AI Institute, Shanghai Jiao Tong University. 
\end{IEEEbiography}
\vspace{-15mm}
\begin{IEEEbiography}[{\includegraphics[width=1in,height=1.25in,clip,keepaspectratio]{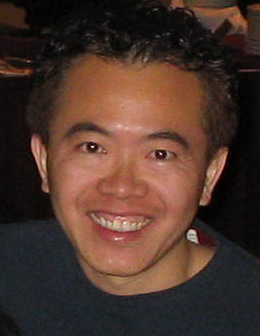}}]{Ming-Hsuan Yang}
is an associate professor in Electrical Engineering and Computer Science at the University of California, Merced. He received the
PhD degree in computer science from the University of Illinois at Urbana-Champaign in 2000. 
%Prior to joining UC Merced in 2008, he was a senior research scientist at the Honda Research Institute working on vision problems related to humanoid robots. He coauthored the book Face Detection and Gesture Recognition for Human- Computer Interaction (Kluwer Academic 2001) and edited special issue on face recognition for Computer Vision and Image Understanding in 2003, and a special issue on real-world face recognition for IEEE Transactions on Pattern Analysis and Machine Intelligence. 
Yang served as an associate editor of the IEEE Transactions on Pattern Analysis and Machine Intelligence from 2007 to 2011 and is an associate editor of the International Journal of Computer Vision, Computer Vision and Image Understanding, Image and Vision Computing and Journal of Artificial Intelligence Research. He received the NSF CAREER award in 2012, and Google Faculty Award in 2009. He is a senior member of the IEEE and the ACM.
\end{IEEEbiography}
\end{document}